\documentclass[preprint,12pt]{elsarticle}

\usepackage{amssymb}
\usepackage{amsmath}
\usepackage{soul}

\usepackage{url}
\usepackage{graphicx}
\usepackage{setspace}
\usepackage{hyperref}
\usepackage[numbers]{natbib}
\usepackage{xcolor}
\usepackage[margin=1in]{geometry}
\usepackage{multirow}
\usepackage{tikz}
\usepackage{csquotes}
\usepackage{algorithm} 
\usepackage{algorithmic}
    \newfloat{algorithm}{t}{lop}
\usepackage{subfig}
\usepackage{longtable}
\usepackage{float}
\usepackage{array}

\usepackage{stix}
\journal{Robotics and Autonomous Systems}

\begin{document}

\begin{frontmatter}



\title{Autonomous search of real-life environments combining dynamical system-based path planning and unsupervised learning}

%\author[label1,label2,label3,label4]{}
\affiliation[label1]{organization={Dynamic Systems and Intelligent Machines Lab}}
\affiliation[label2]{organization={Dynamic Systems and Control Lab}}

\affiliation[label3]{organization={San Diego State University, Mechanical Engineering},%Department and Organization
            addressline={5500 Campanile Drive}, 
            city={San Diego},
            postcode={92182}, 
            state={California},
            country={United States of America}}

\author[label1,label3]{Uyiosa Philip Amadasun}
\author[label1,label3]{Patrick McNamee}
\author[label1,label3]{Zahra Nili Ahmadabadi\thanks{Corresponding author.}}
\author[label2,label3]{Peiman Naseradinmousavi}
%% Abstract
\begin{abstract}
In recent years, advancements have been made towards the goal of using chaotic coverage path planners for autonomous search and traversal of spaces with limited environmental cues. However, the state of this field is still in its infancy as there has been little experimental work done. The existing experimental works have not developed robust methods to satisfactorily address the immediate set of problems a chaotic coverage path planner needs to overcome in order to scan realistic environments within reasonable coverage times. These immediate problems are as follows: (1) an obstacle avoidance technique that reduces halts or disruptions in continuous chaotic trajectories, 
 (2) a means to spread chaotic trajectories across the environment (especially crucial for large and/or complex-shaped environments) that need to be covered, and (3) a real-time coverage calculation technique that is accurate and independent of cell size. This study addresses these problems by developing a novel applied framework for real-world applications of chaotic coverage path planners while providing techniques for effective obstacle avoidance, chaotic trajectory dispersal, and accurate real-time coverage calculation.
These algorithms were created within the ROS framework and make up a newly developed chaotic path planning application. The performance of this application was comparable to that of a conventional optimal path planner. The performance tests were carried out in environments of various sizes, shapes, and obstacle densities, both in real-life and Gazebo simulations. The source code is available at: \url{https://gitlab.com/dsim-lab/paper-codes/Autonomous_search_of_real-life_environments}
\end{abstract}


\begin{highlights}
\item A novel accurate real-time scanning approach for chaotic coverage path planning (CCPP).
\item  Enabling efficient, accurate and real-time scanning of complex-shaped realistic environments.
\item A retroactive collision avoidance method, reducing deviations from continuous paths. \color{black}
\item Comparable performance of CCPP and boustrophedon path planner.
\item Practical experimental real-time implementation of CCPP.
\end{highlights}

%% Keywords

\end{frontmatter}



\newpage
\section*{NOMENCLATURE}
\begin{center}
\begin{longtable}{p{2.1cm} p{5cm} p{3.5cm} p{4.5cm}}
$A$ & Arnold system's parameter & ${P}(\text{o})_{(X_i,Y_i)}$ & Occupancy probability value at a coordinate \\
$arnpnt$ & Tuple of Arnold system coordinates and accompanied trajectory points & $n_{TP_{DS-R}}$ & Vector containing the last row of $TP_{DS-R}$ \\
$B$ & Arnold system's parameter & $O_{X,Y}$ & Origin of an occupancy-grid map \\
$C$ & Arnold system's parameter & $PDA$ & Occupancy probability data array \\
$Cell_{F}$ & Representation of a cell with an occupancy probability value of 0 & $R_{SZ}$ & Ratio of maximum sensing area to average zone size \\
$Cell_{O}$ & Representation of a cell with an occupancy probability value of 100 & $SC_a$, $SC_{r_t}$ & Scan angle and scan range at time $t$ \\
$Cell_{U}$ & Representation of a cell with an occupancy probability value of -1 & $SR$ & Sensing range \\
$Cell_{X,Y}$ & Coordinates of a cell in an occupancy-grid map & $TC_{DS-R}$ & Temporary matrix of Arnold dynamical system and robot coordinates \\
$Cost$ & Cost of travel to a trajectory point & $TF_{MS_t}$ & Transformation matrix for map frame to sensing frame at a specific time \\
$c_z$ & Zone's coverage rate & $Th_{1}$, $Th_{2}$ & Cost thresholds \\
$CT$ & Time for desired coverage rate & $tp$ & A potential or current trajectory point for the robot to follow \\
$d$ & Distance between the robot and a zone centroid & $tp_{n-1}$ & $tp$ at the last iteration \\
$dc$ & Desired coverage rate  & $tp_{new}$ & $tp$ used for replacement \\
$dist$ & Distance between the robot and a cell & $Tp$ & Set of trajectory points \\
$FOV$ & Field of view of the sensor & $v$ & Robot's velocity \\
$f,g$ & Cost function parameters & $x(t),\ y(t),\ z(t)$ & Coordinates of the Arnold system \\
$H,W$ & Height and width of the occupancy grid & $(X,Y)_{M}$, $(X,Y)_{S}$ & Point coordinates in the map frame and sensor frame \\
$ind$ & Index number of cell in the probability array & $(X,Y)_{RM}$, $(X,Y)_{SM}$ & Robot and sensor pose in the map frame \\
$i,j$ & Incremental indices & $(X,Y)_{RM_t}$, $(X,Y)_{SM_t}$ & $(X,Y)_{RM}$ and $(X,Y)_{SM}$ at a specific time \\
$M_{C}$ & Matrix storing coverage status and the zone identification of cells & $(X,Y)_{zone}$ & Centroid of a zone \\
$M_{Z}$ & Matrix storing coverage information and the centroid of every zone & $zid$ & Zone identification index \\
$n_{iter}$ & Number of iterations & $\alpha$ & Orientation of a cell from the sensor field of view \\
$ns$ & Set of iterations & $\bar{P}(\text{o})_{Cell_{(X,Y)}}$ & Average occupancy probability value representing cost function parameter\\
$\theta$ & Mapping variable & $\chi$ & Characteristic function \\
\end{longtable}
\end{center}
\setcounter{table}{0}
\newpage
\section{Introduction}
Coverage path planning (CPP) algorithms are primarily developed for the purpose of creating trajectories  that enable a robot to visit an area while avoiding any obstacles. Such algorithms have several applications, from use in varied environments such as households \cite{choi2017b}, farm fields \cite{hameed2014intelligent}, and surveillance and search tasks \cite{choi2020energy,di2016coverage,faigl2011sensor,grotli2012path,hsu2014complete,kapoutsis2017darp,li2014k,zhu2019complete}. One of the relatively recent methods of CPP algorithm development is found in the use of chaotic motion, thus the moniker, chaotic coverage path planning (CCPP). Chaotic motion as a basis for coverage path planning opens some potential advantages over other more established CPP algorithms in the area of surveillance missions. 
It allows the robot to scan uncertain environments while avoiding obstacles as well as adversarial agents, i.e., unpredictability of the motion enables avoiding attacks. Although the generated trajectories appear random to adversaries, the deterministic nature of these planners provide the possibility of being controlled by the designer to adjust the level of unpredictability and quality of coverage.  \\
Studies have used nonlinear dynamical systems (DS) to generate chaotic trajectories \cite{nakamura2001chaotic,pimentel2017chaotic,li2013improved,sridharan2020multi,sridharan2022online}\color{black}. Nakamura and Sekiguchi \cite{nakamura2001chaotic} showed that while algorithms like random walk generate unpredictable trajectories, the density of said trajectories are less uniform than that of chaotic model. This is because the random walk algorithm resists continuity in the robots motion where the chaotic model does not. To build off this earlier work, other studies introduced the chaos manipulation techniques including random number generators \cite{pimentel2017chaotic}, as well as arccosine and arcsine transformations \cite{li2013improved} to improve dispersal of trajectory points across an environment map. While these methods might have indirectly helped to decrease the coverage time to some extent, they mostly focused on improving the coverage rate rather than the coverage time. Our previous works \cite{sridharan2020multi,sridharan2022online} focused on increasing the efficiency of the chaotic path planning approaches by specifically controlling the chaotic trajectories to directly reduce the coverage time. These works proposed various chaos control techniques to make the planner adaptable and scalable to cover environments varying in size and property within a finite time. These techniques were successful in significantly reducing the coverage time and bringing it closer to optimal time.
While the previous studies have contributed to the field of CCPP, the challenges pertaining to practical implementation of these planners to real world environments \color{black} remain largely unexplored and unaddressed. 

The problem this study addresses is the lack of rigorous experimental validation and implementation of the CCPP approaches in existing studies. The few experimental works in this field \cite{volos2013experimental,majeed2020mobile,tlelo2014application,sooraska2010no} do not contain the components essential for implementation of the methods in realistic cluttered environments and with reasonable coverage times. These methods specifically lack: 

\textbf{(1)} an adaptable obstacle avoidance technique which can cause challenges and inefficiencies in cluttered environments containing complex-shaped barriers. In contrast to many other CPP methods, such as cellular decomposition methods and random walk approaches that produce discrete trajectories \cite{ahuraka2023chaotic}, CCPP algorithms generate continuous trajectories (when based on continuous dynamical systems). The existing CCPP methods employ obstacle avoidance techniques that require the robot to frequently break from these continuous paths, halt its motion (to take necessary actions) \cite{volos2013experimental}, or create paths with sharp angles that violate its kinematic constraints \cite{sooraska2010no}.  

\textbf{(2)} an effective dispersal technique to thoroughly spread chaotic trajectories, which particularly limits the applications in large and/or cluttered environments. The existing implementations of CCPP overlook the coverage time criterion and do not employ any particular technique to enforce more even coverage; a strategy which would result in infinitely long coverage times in such environments. 

\textbf{(3)} an accurate coverage calculation method, leading to either an overestimation or underestimation of the coverage. Existing CCPP studies do not incorporate sensor data and instead calculate coverage based on the number of cells the trajectory passed through. This approach often counts partially covered cells as fully covered and fails to account for multiple cells being covered simultaneously.

The objective of this study is to address the aforementioned shortcomings and establish a novel applied framework for real-world applications of CCPP by making the following contributions to the \ul{existing CCPP} methods:

\textbf{(1)} A retroactive obstacle avoidance technique for CCPP applications, which significantly reduces halts or disruptions in the planned continuous trajectories of the chaotic planner and prevents violation of kinematic constraints. This technique uses descritized map data stored in a quadtree \cite{samet1988overview} and a cost function which we developed.

\textbf{(2)} An adaptive dispersal technique for CCPP applications, featuring an unsupervised machine learning-based map-zoning approach to spread chaotic trajectories across the map. This method enables efficient scanning, reducing coverage time particularly in large or complex-shaped environments.

\textbf{(3)} A fast accurate coverage calculation technique for CCPP applications, featuring a real-time computation technique that utilizes the aforementioned quadtree, as well as sensor data, and storage matrices. To our knowledge, using quadtree for fast real-time coverage rate computation (with sensor data) is a novel contribution to the field of coverage path planning. Other papers \cite{huang2020multi,jan2019complete} which discussed the use of the quadtree data structure for CPP application, employed it for the purpose of efficient partitioning of the coverage area into smaller regions (and subregions), thus enabling easier identification of the areas in need of coverage.

We use two metrics to evaluate the effectiveness of this novel applied framework for CCPP: coverage time and thoroughness of coverage. The performance was evaluated using these metrics in varied realistic irregular-shaped environments (in both a real-life environment and Gazebo simulations) and was compared against a widely used optimal path planner, the Boustrophedon method. 

\color{black}

These contributions were made through the development of algorithms which communicate with each other via the Robot Operating Software (ROS) framework. The previous studies did not carry out the experiments/simulations in specialized robotics simulation software, such as
ROS, and hardly went beyond MATLAB. This adds further challenge in accurately assessing
the effectiveness of the techniques in real-world scenarios. The advancements made in this study can significantly broaden the range of applications for CCPP, including search and rescue, cleaning, surveillance, and exploration in irregular-shaped environments.  \color{black}

The rest of the paper is arranged
as follows. Section 2 provides an overview of system architecture for our CCPP methodology. \color{black} Section 3 explains the integration of chaotic systems to develop robot trajectory.
Section 4 discusses the path planning and chaos control techniques. Section 5 will discuss results of
performance in both simulated Gazebo environments and live tests with varying environments, as
well as performance comparison with the well-established boustrophedon path planning algorithm.
Section 6 concludes the paper and discusses future work.
 %All algorithms developed in this work are written in the Python language.

\section{Overview of System Architecture}
\label{sec:Overview of System Architecture}
A more in-depth discussion of the algorithms and software packages that make up our system architecture will be provided in section \ref{sec:Path planning strategy and chaos control technique}. This section will serve as a primer. 

% In your preamble:
%\usepackage{xcolor}
\definecolor{darkpurple}{RGB}{50,0,255}
\definecolor{lightblue}{RGB}{0,100,200}
\definecolor{green}{RGB}{0,195,0}
% In your document:
\begin{figure}[h]
    \centering
     \rotatebox{0}{\includegraphics[width=17cm]{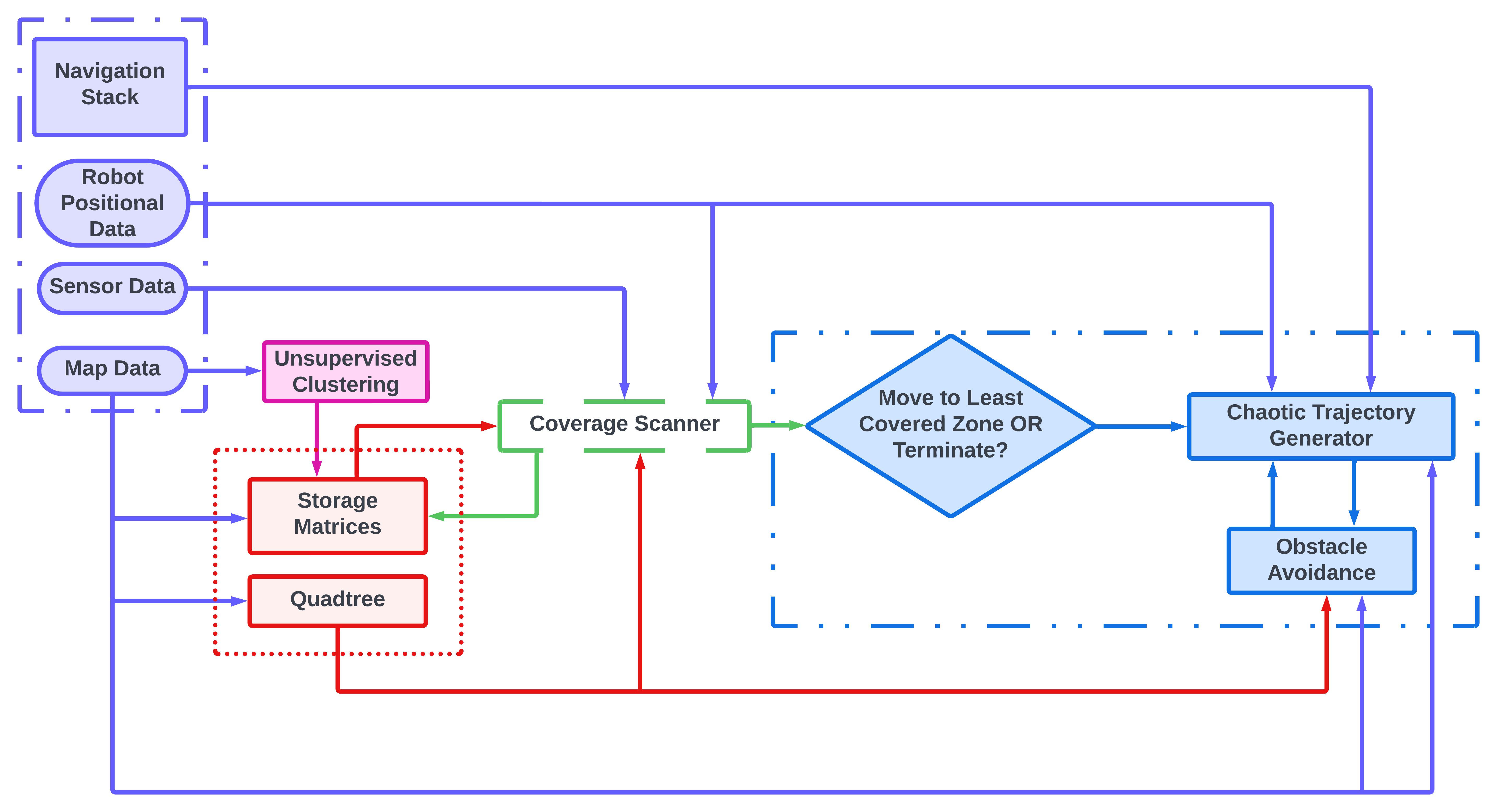}}
\caption{System architecture.
     {\color{darkpurple}\textbf{\_ \hspace{0.1em} \raisebox{-0.38ex}{.} \hspace{0.1em} \_}}: ROS Software Stack, 
     {\color{red}\textbf{.......}}: Data Structures, 
     {\color{green}\textbf{\_\_\_  \hspace{0.1em} \_\_\_ \hspace{0.1em} \_\_\_}}: Coverage Scanner, 
     {\color{lightblue}\textbf{\_  \hspace{0.1em} \raisebox{-0.38ex}{.} \hspace{0.1em} \raisebox{-0.38ex}{.}  \hspace{0.1em} \_}}: Path Planning Controller\color{black}}

 \label{fig:system_architecture}
     
\end{figure}

Figure \ref{fig:system_architecture} depicts our system architecture. Map data of free space in the environment is arranged in a quadtree structure. This same data is also divided into zones using unsupervised clustering (see section \ref{sec:Map-zoning Technique For Effective Dispersal of Chaotic Trajectories}). Under the Coverage Scanner implementation, the total coverage and coverage of each zone are updated by the coverage calculator as the robot moves around the environment via the Path Planning Controller. All coverage information is recorded in storage matrices. The storage matrices are created using map data and zone information from unsupervised clustering. The coverage calculator performs real-time coverage scanning based on matching the map data retrieved from the quadtree with current sensor data. This matching represents scannable space within the robots sensing range (see section \ref{subsec:Real-time computation technique for coverage calculation}). The Path Planning Controller generates chaotic trajectories across the map and away from static obstacles (see section \ref{sec:Trajectory Point Generation and Obstacle Avoidance Technique}). Using map data (retrieved via quadtree) and a cost function, the Path Planning Controller manages an obstacle avoidance method which tries to considerably reduce deviation from the original chaotic path. The cost function uses map data to retrieve information about the closeness of free space to obstacle and/or unknown space (see section \ref{sec:Custom cost functions}). This obstacle avoidance method also ensures that the robot does not need to halt its movement during obstacle avoidance. A procedural decision making process controls the robot's movement to least covered zones so as to distribute chaotic trajectories across the entire map. The decision making process also dictates when to terminate the CPP once desired coverage ($dc$) is achieved (see section \ref{sec:Decision Making Process for Chaotic trajectory Proliferation}). This decision making works based on communication between the Path Planning Controller and the Coverage Scanner. It is important to note that the use of the quadtree data structure is essential to our system architecture as it enables quick search of map data which is crucial to our real-time use case. The underlying model used for chaotic trajectory generation is discussed in the next section.\color{black}

\section{Integration of chaotic dynamical systems for mobile robot}
\label{sec:Integration of chaotic dynamical systems for mobile robot}
The chaotic dynamical systems can be in form of continuous ~\cite{volos2012implementation,nakamura2001chaotic,agiza2001synchronization,li2016bounded,lu2004new} or discrete systems ~\cite{petavratzis2022experimental,moysis2021chaotic,petavratzis20212d,petavratzis2019coverage,li2013improved,arrowsmith1993bogdanov,curiac20142d,li2017chaotic,li2015chaotic}. This work will use the continuous Arnold system simply because our previous study ~\cite{sridharan2022online} showed the potential efficiency and flexibility to adapt and scale to different coverage tasks. The Arnold system ~\cite{nakamura2001chaotic} is described by the following equations:

\begin{equation}
\label{eqn: 1}
\begin{cases}
\mathrm{d}\mathit{x(t)}/\mathrm{dt}=A\sin \mathit{z(t)} + C\cos \mathit{y(t)}\\
\mathrm{d}\mathit{y(t)}/\mathrm{dt}=B\sin \mathit{x(t)} + C\cos \mathit{z(t)}\\
\mathrm{d}\mathit{z(t)}/\mathrm{dt}=C\sin \mathit{y(t)} + B\cos \mathit{x(t)}\
\end{cases}
\end{equation}

where variables $x(t)$, $y(t)$, and $z(t)$ are the DS coordinates; and $A$, $B$, and $C$ are the Arnold system parameters. A nonlinear dynamical system becomes chaotic upon gaining these properties: (1) sensitivity to initial conditions (ICs), and (2) topological transitivity. Per our previous work ~\cite{sridharan2022online}, we use the tuple $(x_0,y_0,z_0)=(0,1,0)$ as initial conditions (ICs) for the Arnold system. The Arnold system parameters are chosen to be $A=0.5$, $B=0.25$, and $C=0.25$. The analysis of the parameter space conducted in our previous study ~\cite{sridharan2022online} showed that these values satisfy the properties (1) and (2) for a chaotic system and result in reasonably uniform coverage of the environment when the Arnold system is integrated into the robot's controller. 

\begin{figure} [h]
    \centering
     \includegraphics[width=5cm]{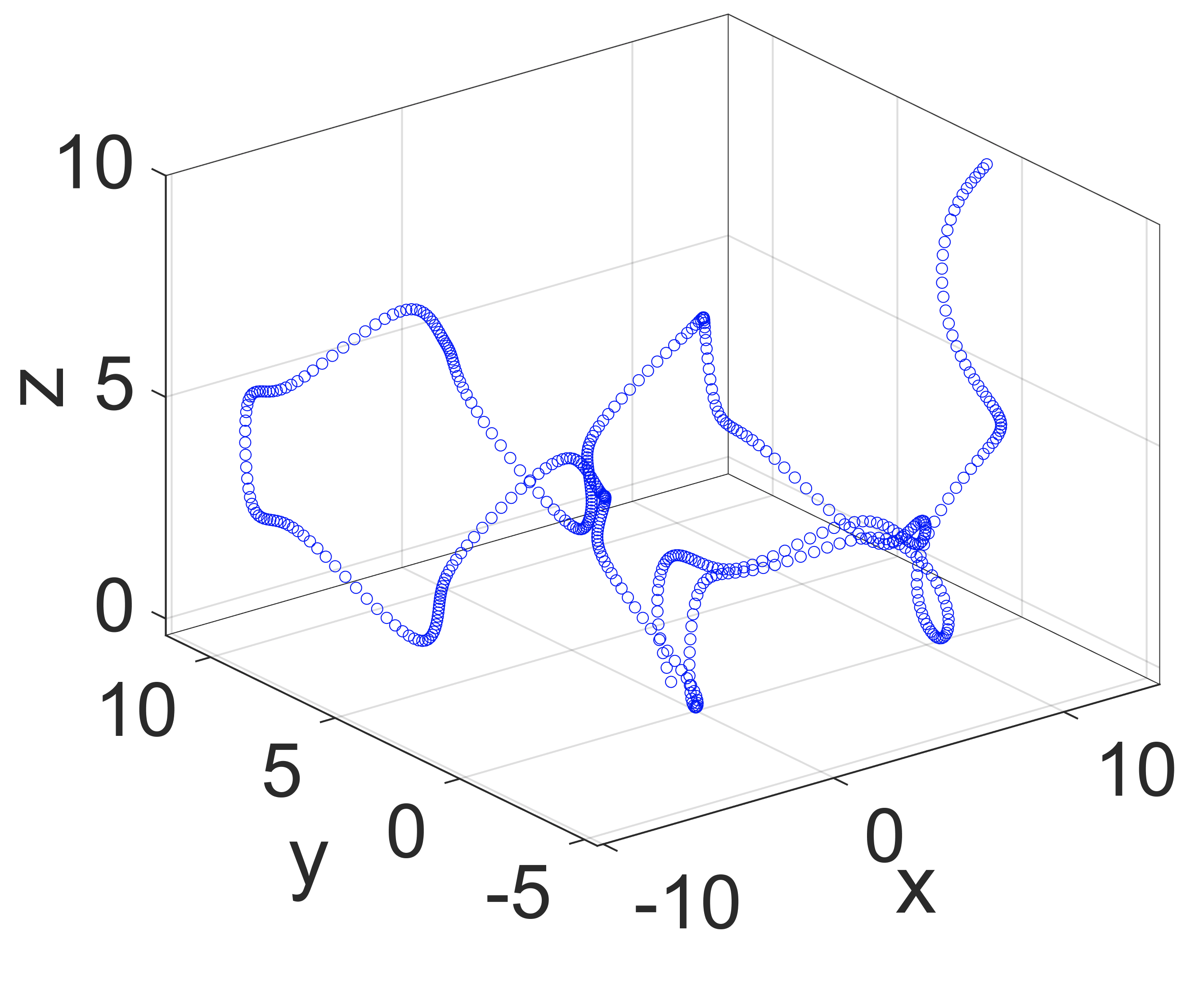} 
 \caption{3-D chaos attractor for the Arnold system, first published in \cite{sridharan2022online} by Springer Nature.}
    \label{fig:Arnoldattractor}
\end{figure}

\begin{figure} [h]
    \centering
     \includegraphics[width=5cm]{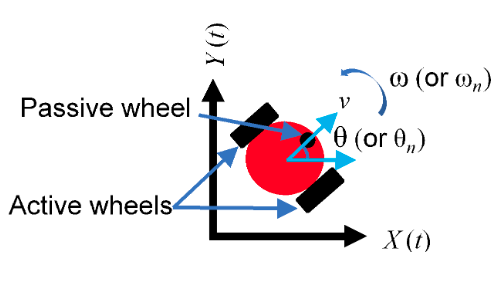} 
 \caption{Mobile robot’s motion on a plane, first published in \cite{sridharan2020multi} by IEEE.}
    \label{fig:robot_schematic}
\end{figure}

Fig. \ref{fig:Arnoldattractor} shows the 3-D chaos attractor of the Arnold system created using these parameters. The CCPP algorithm maps the DS coordinates into the robot's kinematic equation (Eq. (\ref{eqn:2})). New states obtained from mapping process are set as trajectory points for the robot to travel to via ROS navigation stack. 

\begin{equation}
\label{eqn:2}
\begin{cases}
\mathrm{d}\mathit{X(t)}/\mathrm{d}t=v\cos \mathit{x(t)}\\

\mathrm{d}\mathit{Y(t)}/\mathrm{d}t=v\sin \mathit{x(t)}\\

\mathit{\omega(t)}=\mathrm{d}\theta(t)/\mathrm{d}t=\mathrm{d}\mathit{x(t)}/\mathrm{d}t\\
\end{cases}
\end{equation}

where $X(t)$ and $Y(t)$ are the robot's coordinates, $v$ is the robot's velocity, d$t$ is the time step, and $x(t)$ is one of the Arnold system's coordinates mapped into the robot's kinematic relations. The other coordinates of the Arnold system can replace $x(t)$ to perform the mapping, however, the trajectories produced by each coordinate will be different from the others. We will later use this feature in section 3.1.1 to improve the coverage. 
Fig.  \ref{fig:robot_schematic} illustrates the general schematics of a two wheel differential drive mobile robot used in this study. The robot comprises of two active fixed wheels with one passive caster, subject to a non-holonomic constraint.

In our developed algorithms, the generated $X(t)$ and $Y(t)$ coordinates are set as goal coordinates in ROS. The ROS navigation stack specifically sets the goal coordinates on the environment map and uses the global planner to create a trajectory to aforementioned coordinates using Djikstras algorithm. The robot then follows this trajectory while making course corrections via the local planner. For our use case, we examined two popular local planner packages already developed within ROS, the DWA \cite{dwa2020} and TEB \cite{{teb2020}} local planners. Both local planner favor curved trajectories which are generally favorable for chaotic motion. However, the DWA local planner was ultimately chosen for reasons explained in section \ref{sec:Path planning strategy and chaos control technique}. Section 4 provides a detailed explanation of our system. \color{black}

\section{Path planning strategy and chaos control technique}
\label{sec:Path planning strategy and chaos control technique}
The objective of this work is to tackle the challenges associated with experimental real-time implementation of CCPP in order to make this method  as competitive as the other commonly used coverage path planning methods. We adapt some of the theoretical chaos control techniques proposed in our previous study \cite{sridharan2022online} and make them applicable to real-life environments while providing the following contributions. 
(1) A retroactive obstacle avoidance technique which utilizes a cost function and a quadtree data structure created from discretized map data to set trajectories away from obstacles while significantly reducing both deviations from the chaotic path as well as frequency of disruptions that break continuous motion. \color{black}
(2) A map-zoning technique based on an unsupervised machine-learning clustering algorithm which continuously directs the robot to less-visited distant areas of the environment. And
(3) a computation technique  which is independent of cell size, but instead based on discretized map data (via quadtree) and sensor data to provide accurate real-time coverage calculation. 

\begin{equation}
\label{eqn:3}
ind = (Cell_X \times W) + Cell_Y
\end{equation}
\begin{equation}
\label{eqn:4}
\begin{cases}
X_M=(Cell_X \times res) + O_X\\
Y_M=(Cell_Y \times res) + O_Y
\end{cases}
\end{equation}
\begin{equation}
\label{eqn:5}
\begin{cases}
Cell_X = ind  \bmod   W\\
Cell_Y = \frac{(ind - Cell_X)}{W}
\end{cases}
\end{equation}
\begin{equation}
\label{eqn:6}
\begin{cases}
Cell_X = \frac{X_M - O_X}{res}\\
Cell_Y = \frac{Y_M - O_Y}{res}
\end{cases}
\end{equation}

In what follows: (1) the algorithms interact with map data information of a 2D occupancy-grid map. (2) Cells refer to the smallest possible segments of the map, while zones represent larger segments. (3) The relevant map data information includes: occupancy probability data array ($PDA$), map resolution ($res$), map height ($H$), map width ($W$), and origin of map ($O_{(X,Y)}$). (4) The $PDA$ contains the occupancy probability values of every cell in the occupancy-grid map and the algorithms interact with this information in various ways. Eqs. (\ref{eqn:3})-(\ref{eqn:6}) are used in several algorithms to convert relevant data field information to useful formats to perform various tasks. Eq. (\ref{eqn:5}) derives the coordinate of a cell in the occupancy-grid map ($Cell_{(X,Y)}$) from its corresponding $PDA$ index value ($ind$). Eq. (\ref{eqn:6}) derives the coordinates of a cell ($Cell_{(X,Y)}$) using the $O_{(X,Y)}$ and its map frame coordinate ($(X,Y)_M$). (5) The map of the environment is broken into individual zones before all other algorithms start to run. (6) The quadtree data structure solely contains the $Cell_{(X,Y)}$ positions of every cell which represents free space. The quadtree and the storage matrices ($M_Z$ and $M_C$) are created (directly or indirectly) using relevant map data information (before the start of any algorithms excluding the unsupervised machine-learning clustering algorithm) for quick queries and zone coverage updates. The quadtree algorithms developed in \cite{Quadtrees} have been adapted for this work. (7) The transformation matrix ($TF_{MS}$) (created via ROS) transforms point coordinates from the map frame to the sensor frame and is used along with sensor data for coverage calculation. (8) The communications between any algorithms that do not directly invoke each other is performed via the ROS framework. (9) As the robot moves across the map, trajectory points are set as goals by the ROS navigation stack. The navigation stack provides a global path planner based on the Dijkstra algorithm. The DWA and TEB local planners were tested to see which would be the best. In the final analysis, a DWA local planner is chosen with full awareness of the sub-optimal trajectories it often generates. However, the DWA local planner was chosen for the following reasons. (i) It is less sensitive to the choice of parameters, making it easier to tune and use in different environments. (ii) It can handle real-world scenarios of dynamic obstacles more effectively, as it takes into account the motion of the robot and the motion of other objects in the environment when planning a path. And (iii) it is computationally less expensive, making it more suitable for use on robots with limited computational resources. (10) It must be noted that all algorithms are written in the Python language and therefore any indices of any vectors, arrays or matrices discussed starts at 0.

\begin{figure} [h]
    \centering
     \rotatebox{0}{\includegraphics[width=17cm]{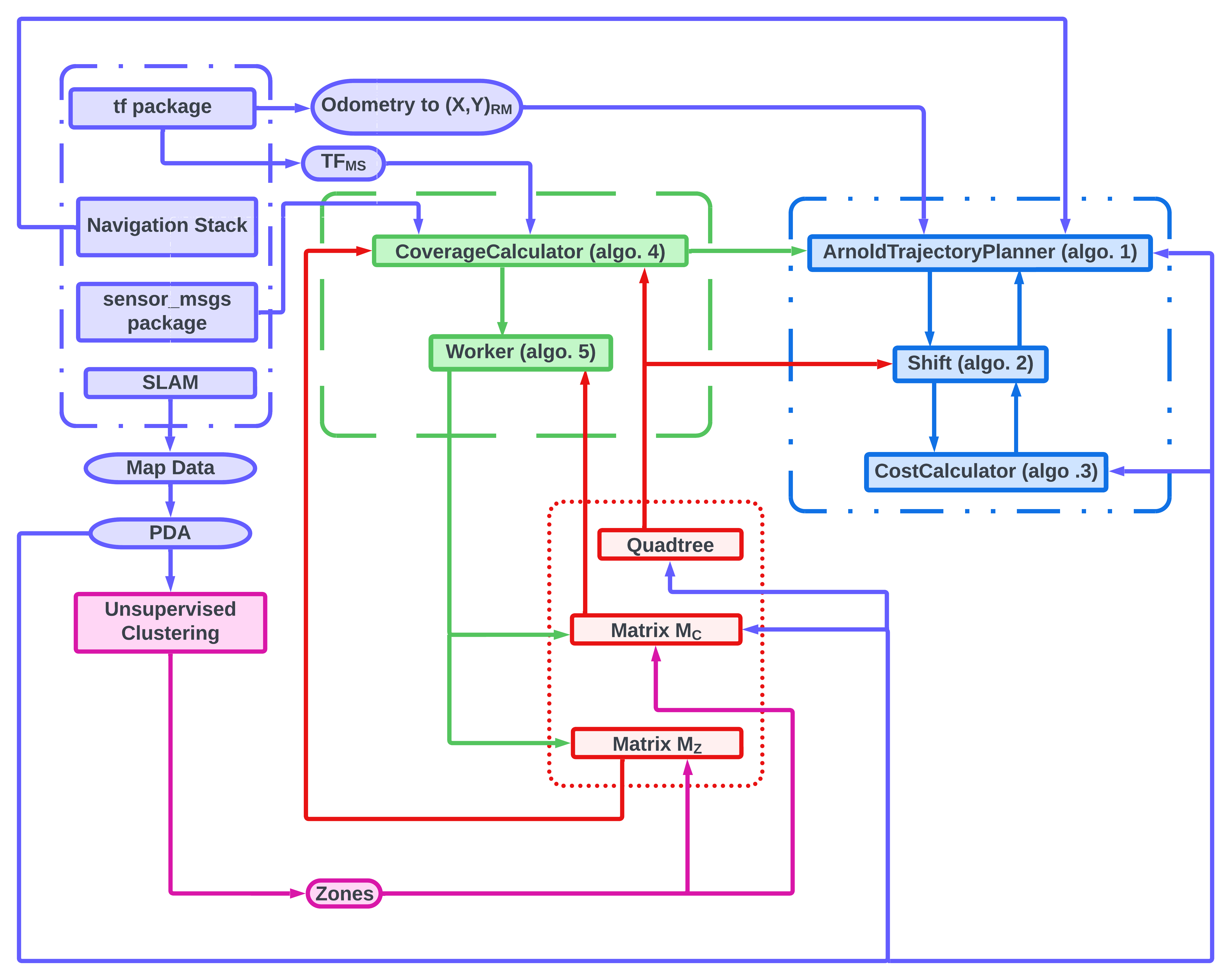}}
\caption{Flowchart showing algorithm flow in the ROS1 environment.
     {\color{darkpurple}\textbf{\_ \hspace{0.1em} \raisebox{-0.38ex}{.} \hspace{0.1em} \_}}: ROS Software Stack, 
     {\color{red}\textbf{.......}}: Data Structures, 
     {\color{green}\textbf{\_\_\_  \hspace{0.1em} \_\_\_ \hspace{0.1em} \_\_\_}}: Coverage Scanner, 
     {\color{lightblue}\textbf{\_  \hspace{0.1em} \raisebox{-0.38ex}{.} \hspace{0.1em} \raisebox{-0.38ex}{.}  \hspace{0.1em} \_}}: Path Planning Controller\color{black}}
    \label{fig:Flowchart}
\end{figure}

The rest of this section will provide an overview of the CCPP process in ROS using the flowchart shown in Fig. \ref{fig:Flowchart}. As seen, the first step is to create the occupancy-grid map via simultaneous localization and mapping (SLAM). The resulting occupancy grid map is then published as map data. As discussed earlier, part of the map data is the $PDA$. The $(X,Y)_M$ values of all cells that represent free space, are derived from the $ind$ values of the aforementioned cells in the $PDA$ (using Eqs. (\ref{eqn:5}) and then (\ref{eqn:4})). These $(X,Y)_M$ values are stored in an array which is fed as part of the input to the clustering algorithm. The clustering algorithm uses this information to divide the environment into zones. The matrix $M_Z$ stores zones information. The storage matrix $M_C$ is created using all the index information from the $PDA$ and zones information. $M_C$ keeps track of cell coverage to memory.

Then, the chaotic coverage path planning process is carried out. We developed 5 algorithms which make up the Coverage Scanner and Path Planning Controller introduced in section \ref{sec:Overview of System Architecture}. Algorithms 1, 2 and 3 make up the Path Planning Controller, while Algorithms 4 to 5 form the Coverage Scanner. \color{black}The ArnoldTrajectoryPlanner ($ATP$) function, as outlined in Algorithm 1, employs the Arnold system to continuously generate chaotic trajectories within the map frame until the $dc$ is achieved.  Algorithm 1 utilizes the information provided by the $PDA$ to facilitate obstacle avoidance decision making. In response to the information provided by the $PDA$, Algorithm 1 interacts with Algorithms 2 and 3, directing chaotic trajectories away from obstacles. Algorithm 2 (called by Algorithm 1) leverages the quadtree and a two-parameter cost function. Algorithms 2 and 3 exchange information to calculate these two parameters, which they subsequently relay to Algorithm 1. 
Meanwhile, Algorithm 5 plays a role in publishing data regarding the least-covered zones. Algorithm 1 utilizes this information to strategically distribute chaotic trajectories across the map. 
Algorithm 4 receives published data from the ROS sensor\_msgs and ROS tf package, which it combines with information from the quadtree to calculate total coverage ($tc$) with the help of Algorithm 5. Algorithm 5 updates the information in the storage matrices $M_C$ and $M_Z$ to keep track of the coverage of every zone. $M_C$ keeps track of covered cells and $M_Z$ tracks zone coverage. Algorithm 4 then uses the information in $M_Z$ to calculate $tc$. Each iteration of Algorithm 4 uses up-to-date sensor data and $TF_{MS}$ to accurately calculate $tc$ until $dc$ is reached. Once $dc$ is reached, Algorithms 4 and 1 are prompted to stop. The initial position of the robot in the map frame ($(X,Y)_{RM}$) is used as a starting point to generate trajectory points successively. $(X,Y)_{RM}$ is retrieved from the transformation of odometry data using the tf package. The next section starts with discussing the obstacle avoidance technique and explains the interaction between the Algorithms 1, 2, and 3.
%\newpage
%\newpage
\subsection{Trajectory Point Generation and Obstacle Avoidance Technique}
\label{sec:Trajectory Point Generation and Obstacle Avoidance Technique}
The proposed dynamic of trajectory point generation and obstacle avoidance uses a quadtree and cost function to maintain continuous motion (with reduced disruptions/ deviations) while avoiding \color{black} obstacles and provide quick chaotic path planning capabilities in environments varying in size, shape, and obstacle density. Our work creates this quadtree from descritized map data to quickly access occupancy-grid map coordinates representing free space. Conceptually,  the dynamic works as follows: (1) The Runge-Kutta fourth-order (RK4) method produces a trajectory point ($tp$). (2) The corresponding $PDA$ index value ($ind$) of the $tp$ is calculated and evaluated in the $PDA$ to check its probability value. Probability values $\ne$ 0 are deemed non-viable. (3) Cells representing free space within some radius ($r$) of a non-viable $tp$ are accessed from the quadtree (via querying \cite{Quadtrees}). It should be noted that the $r$ value is chosen based on the resolution ($res$) of the occupancy-grid map. The $res$ for all simulations and tests is 0.05 meters per pixel. (4) Each coordinate is evaluated using a cost function to determine a favorable replacement coordinate. We used a maximum $r$ value of 19 in all tests. This choice did not noticeably impact the overall computation speed of the program. (5) A replacement coordinate is set as the new $tp$ which is then used in the generation of the next $tp$.

In scenarios where there are no free cells in the immediate vicinity of a non-viable $tp$ for replacement, additional processes are carried out to generate a viable $tp$. This case, as well as other special scenarios, will be further discussed in section \ref{sec:Algorithm workflow}. This whole process attempts to replace most occurrences of non-viable $tp$ such that the evolution of the chaotic system is not hindered unless absolutely necessary. In other words, the robot does not have to stop moving, or break from its ongoing general direction unless absolutely necessary. As a positive side effect, we discovered that the chaotic trajectory across the local area around the robot's position is spread more efficiently. Therefore, increasing the likelihood of covering new cells at a faster rate and ultimately reducing the coverage time ($CT$). Quick access to cell information is critical for fast computation, allowing these steps to take place within a sufficiently short time frame, which is essential for achieving obstacle avoidance without the need to stop motion. It should be noted that the ROS navigation stack sets the next successive goal once the robot is within some threshold of the current one. This threshold is set to a certain length based on the ROS path messages \cite{navmsgs}, which provide a list of coordinate points that the robot follows on the path to a goal. 
%\newpage
\begin{algorithm}[htbp]
  \caption*{\normalfont\textbf{Algorithm 1 (main):} $ArnoldTrajectoryPlanner(A,B,C,v)$}
   %\algsetup{linenosize=\tiny}
  %\scriptsize
  \begin{algorithmic}[1]
    \STATE $|TP_{DS-R}|_{first \ row} \leftarrow$ [$x_0,y_0,z_0$,$(X,Y)_{RM}$]
    \WHILE{$tc \ < \ dc$}
        \IF{$\sum_{i=1}^{n} ns==n_{iter}$ $\texttt{OR}$ \text{successive bad runs}}
           \STATE $(X,Y)_{\text{zone}} \leftarrow $ $\min\limits_{zid_1,\dots zid_T}\leftarrow CoverageCalculator()$ 
            \STATE  $(Cost,tp_{new}) \leftarrow Shift((X,Y)_{zone},[],0)$
            \STATE robot navigates to $tp_{new}$
            \STATE $TP_{DS-R}\leftarrow$ [$x_0,y_0,z_0,tp_{new}$]
        \ENDIF
         \IF{$\sum_{i=1}^{n} ns< n_{iter}$}
            \IF{set $ns$ ends with the generation of a viable $tp$}
               \STATE $ TP_{DS-R} \leftarrow n_{TP_{DS-R}}$
            \ENDIF
            \IF{the generation of previous set $ns$ is interrupted by a $tp$ set outside boundaries}
                \STATE $(Cost,tp_{new}) \leftarrow Shift((X,Y)_{RM},[],0)$
                \STATE $TP_{DS-R} \leftarrow$[$x_0,y_0,z_0,tp_{new}$]
            \ENDIF
        \ENDIF
        \STATE $Tp = [ \ ]$
        \STATE  $n_{TP_{DS-R}} \leftarrow TP_{DS-R}$
        \STATE   sub algorithm 1 -- First stage
        \STATE   sub algorithm 2 -- Second stage
    \ENDWHILE
\end{algorithmic}
\end{algorithm}

\subsubsection{Algorithm workflow}
\label{sec:Algorithm workflow}
Algorithm 1 propagates Eqs. (\ref{eqn: 1}) and (\ref{eqn:2}) to calculate the approximate states of the Arnold system as time progresses. As the evolution of one of the Arnold system DS coordinates ($DS_{index}$) is mapped onto the robot's kinematics, a chaotic trajectory is created. It is important to note that the robot does not initially try to move to every trajectory point ($tp$) as it is created. This means that Algorithm 1 proceeds through multiple iterations ($ns$) at once, generating a set of trajectory points ($Tp$) that describe a chaotic trajectory. The algorithm may then retroactively modify these trajectory points to avoid obstacles as the robot attempts to move sequentially to each $tp \in Tp$.
By using $ns > 1$ instead of $ns = 1$, this process generally produces smoother trajectories as well. It also results in a more favorable local distribution of trajectories within the robot's vicinity, particularly in areas with high obstacle density.
This retroactive approach contrasts with an immediate proactive strategy that attempts to finalize the placement of each generated $tp$ in every iteration. Such a proactive method, which makes corrections for each individual $tp$ as it is generated, did not produce a favorable distribution of trajectories within the robot's vicinity. By generating multiple points before modification, the retroactive method allows for a more holistic and efficient adaptation to the environment. \color{black}This is the reason that a large bulk of Algorithm 1 is developed into two stages, as presented in sub-algorithms 1 and 2. Each stage is a for-loop. In the \textbf{first stage}, trajectory points are successively generated via RK4 method to make up the set $Tp$. At this stage, a trajectory point ($tp$) might be replaced (for obstacle avoidance) before it is added to the set. Therefore every trajectory point ($tp$) $\in Tp$ affects the next $tp$ as each $tp$ was successively calculated from the last. Set $Tp$ establishes some chaotic path. In the \textbf{second stage}, the robot tries to follow this chaotic path. Replacement of any $tp \in Tp$ at this stage does not affect the next $tp$ as the general chaotic path was already established in the first stage. This two-stage process for replacing the trajectory points provides a higher likeliness of continuity of motion. \color{black}This becomes clear in the next paragraphs which provide full explanations of the sub-algorithms processes. 

Each stage of Algorithm 1 will make use of a custom cost function which sets trajectories away from obstacles. The \textbf{first stage} makes use of a threshold ($Th_1$) to judge the cost of any replacement $tp$ to improve the odds involved in the search for highly favorable replacement coordinates. The \textbf{second stage} makes use of threshold ($Th_2$) which is of a smaller value than $Th_1$. This threshold is used as a contingency to check for any viable $tp$ set too close to an obstacle. In this way, said $tp$ can be immediately replaced before it is set as a goal (via ROS) in stage 2.

\begin{algorithm}[htbp]
    \caption*{\normalfont\textbf{Algorithm 1 (sub-algorithm 1):} $ArnoldTrajectoryPlanner(A,B,C,v)$}
   %\algsetup{linenosize=\tiny}
  %\scriptsize
  \begin{algorithmic}[1]
        \FOR{$i<ns$}
            \STATE $arnpnt \leftarrow$ Eqs. (\ref{eqn: 1})  and  (\ref{eqn:2}) with $n_{TP_{DS-R}},A,B,C,v,dt$ and $DS_{index}$ \STATE(The tp and DS are paired and returned one variable) 
            \STATE $tp \leftarrow arnpnt$ 
            \IF{$tp$ is set at a coordinate which does not exist within the occupancy-grid map boundaries}
              \STATE drop this $tp$
               \STATE Break
            \ENDIF
            \IF{$tp$ is viable}
               \STATE $Tp,n_{TP_{DS-R}},TP_{DS-R}\leftarrow$ store and update relevant information to memory
            \ENDIF
            \IF{$tp$ is not viable}
                    \STATE $(Cost,tp_{new})\leftarrow Shift(tp,tp_{n-1},1)$
                \IF{$Cost \geq Th_{1}$} 
                    \STATE $arnpnt \leftarrow$ Eqs. (\ref{eqn: 1}) and  (\ref{eqn:2}) with $n_{TP_{DS-R}},A,B,C,v,dt$ and a different $DS_{index}$ 
                    \STATE If creation of viable $tp$ is unsuccessful $\texttt{OR }$$Cost \geq Th_{1}$(after use of $Shift$ function), continue $DS_{index}$ switching. 
                    \STATE Upon exhaustion of $DS_{index}$ switching, choose the $tp_{new}$ associated with the lowest $Cost$. The $DS$ pair of said $tp$ is stored to memory.
                \ENDIF
            \ENDIF
        \ENDFOR

  \end{algorithmic}
\end{algorithm}

The important decision making processes within the \textbf{first stage} are as follows: 

(1) If the $Cost$ of the $tp_{new}$ found via Algorithm 2 is $\ge Th_1$, this stage utilizes deterministic $DS_{index}$ switching to generate a new $tp$ (via RK4 method). In other words, a different Arnold system coordinate ($DS_{index}$) is mapped onto the robot's kinematics. This $tp$ goes through the same process as the $tp$ created from the previous mapping using the original $DS_{index}$. If $DS_{index}$ switching is exhausted, the $tp_{new}$ associated with the least cost is stored in memory along with the paired DS coordinates. In the event that this process still results in the selection of an irreplaceable non-viable $tp$, the algorithm proceeds with this $tp$. It is added to set $Tp$ and will be replaced at stage 2. It must be noted that the next iteration uses the original $DS$ coordinate before $DS_{index}$ switching occurred.

(2) In the scenario where there is no single $Cell_F$ to be queried within $r$ cell lengths of a non-viable $tp$, this $tp$ itself is assigned as a $tp_{new}$ of the highest possible cost ($Cost_{max}$). This is represented in lines 3-4 of Algorithm 2 (Shift function). This cost value is greater than any possible cost which could be assigned. 

(3) Any $tp$ generated outside the boundaries of the occupancy-grid map is stored in memory (added to set $Tp$). However, the first stage terminates immediately upon this addition. 

\begin{algorithm}[htbp]
   \caption*{\normalfont\textbf{Algorithm 1 (sub-algorithm 2):} $ArnoldTrajectoryPlanner(A,B,C,v)$}
   %\algsetup{linenosize=\tiny}
  %\scriptsize
  \begin{algorithmic}[1]
        \FOR{$tp \in Tp$} 
            \IF{$Cost \ge Th_{2}$ OR non-viable $tp$}
                \STATE $(Cost,tp_{new}) \leftarrow Shift((X,Y)_{RM},[],0)$
            \ENDIF
            \STATE robot navigates to $tp$
        \ENDFOR
 \end{algorithmic}
\end{algorithm}

The \textbf{second stage} (sub-algorithm 2) provides an additional measurement to replace any $tp \in Tp$ $\ge Th_2$, as well as any non-viable $tp \in Tp$ that could not be replaced in the first stage. This stage successively scrutinizes each $tp \in Tp$ within Algorithm 2 immediately before setting it as a goal. During replacement, Algorithm 2 chooses a favorable coordinate by scrutinizing all $Cell_F$ within $r$ of robot's current position, $(X,Y)_{RM}$. Any $tp$, with cost greater than $Th_2$, goes through the same process as a non-viable $tp$. The replacement $tp$ is paired with the $DS$ coordinates corresponding to the $tp$ that was replaced. The entire process of Algorithm 1 strives to establish a continuation between the successive sets of $ns$. It maintains the progression of the chaotic system which favors more efficient coverage. This is achieved by setting the last pair of the $DS$ coordinates and $tp$ stored in $n_{TP_{DS-R}}$ as the IC for the next set of iterations. 

\begin{algorithm}[htbp]
   \caption*{\normalfont\textbf{Algorithm 2:} $Shift(tp,tp_{n-1},\lambda)$}\label{Alg:shift}
  \begin{algorithmic}[1]
    \STATE Query every $Cell_F$ within $r$ of $tp$  
    \IF{Query = 0}
        \STATE $tp_{new}$ = $tp$; the non-viable $tp$ could not be replaced
        \RETURN $|Cost_{max},tp_{new}|$
  
    \ELSE
    \STATE  $tp_{new} \leftarrow argmin_{Cell_F \in Query}\ (CostCalculator(Cell_F) \ + \ \lambda \|(X,Y)_{M} - tp_{n-1}\|_{2}$)
    \RETURN $(Cost_{tp_{new}},tp_{new})$
    \ENDIF
 \end{algorithmic}
\end{algorithm}

\begin{algorithm}[htbp]
  \caption*{\normalfont\textbf{Algorithm 3:} $Cost Calculator(Cell_{(X,Y)})$}\label{Alg:Cost Calculator}
  \begin{algorithmic}[1]
    \STATE $\bar{P}(\text{o})_{Cell_{(X,Y)}} = \frac{1}{n} \sum_{i=1}^{n} P(\text{o})_{(X_i, Y_i)} \cdot \chi(\sqrt{(X_i - X_{Cell_{(X,Y)}})^2 + (Y_i - Y_{Cell_{(X,Y)}})^2} \leq l)$.
    \RETURN $ g \leftarrow \bar{P}(\text{o})_{Cell_{(X,Y)}}$
  \end{algorithmic}
\end{algorithm}

\subsubsection{Custom Cost Functions}
\label{sec:Custom cost functions}
Algorithms 2 and 3 use a cost function, shown by Eq. (\ref{eqn:7}), to help Algorithm 1 generate trajectory points away from obstacles. In what follows, cells representing free space, obstacle space, and unknown space are designated as $Cell_F$, $Cell_O$, and $Cell_U$, respectively. Algorithms 2 and 3 calculate cost function parameters $f$ and $g$, respectively. In Eq. (\ref{eqn:7}), the parameter $f$ is the distance between the generated $tp$ in the last iteration ($tp_{n-1}$) and the point coordinate ($(X,Y)_{M}$) of the $Cell_F$ being evaluated. Parameter $f$ indicates how best a point coordinate maintains the adherence to the chaotic path\color{black}. The $Shift$ function (Algorithm 2) calculates $f$. Parameter $g$ expresses the closeness of a $Cell_F$ to obstacle and/or unknown space. The parameter $g$ is the average of  assigned values of the coordinates within range $l$ (6 cell lengths) of the $Cell_{F}$ being evaluated. The assigned values are as follows. (1) Any coordinate position which matches a cell position is assigned the absolute value of the occupancy probability of that cell. The probability values of $Cell_O$, $Cell_U$ and $Cell_F$ are 100, -1, and 0 respectively. (2) Any coordinate position that does not match a cell location, i.e., is not within the occupancy-grid boundaries, is assigned a value of 500. Such a position is assigned the highest cost so as to set trajectories away from the occupancy-grid map boundaries. Additionally, this will reduce the likelihood of generating trajectory points outside the boundaries. The parameter $\chi$ in Algorithm 3 is a characteristic function which evaluates to 1 if the $(X,Y)_{i}$ coordinate is within range $l$ of the position of the evaluated $Cell_F$. The $tp_{new}$ is derived from the cell associated with the lowest $Cost$. Parameter $f$ is omitted from the cost function (i.e., $\lambda=0$ in Algorithm 2) in cases where the choice of $tp_{new}$ does not cause appreciable deviation from the chaotic trajectory \color{black} and the primary focus is to maintain a safe distance from obstacles. 

\begin{equation}
\label{eqn:7}
Cost = g + f
\end{equation}

\begin{figure} [htbp]
    \centering
     \includegraphics[height=3.5cm,width=3.5cm]{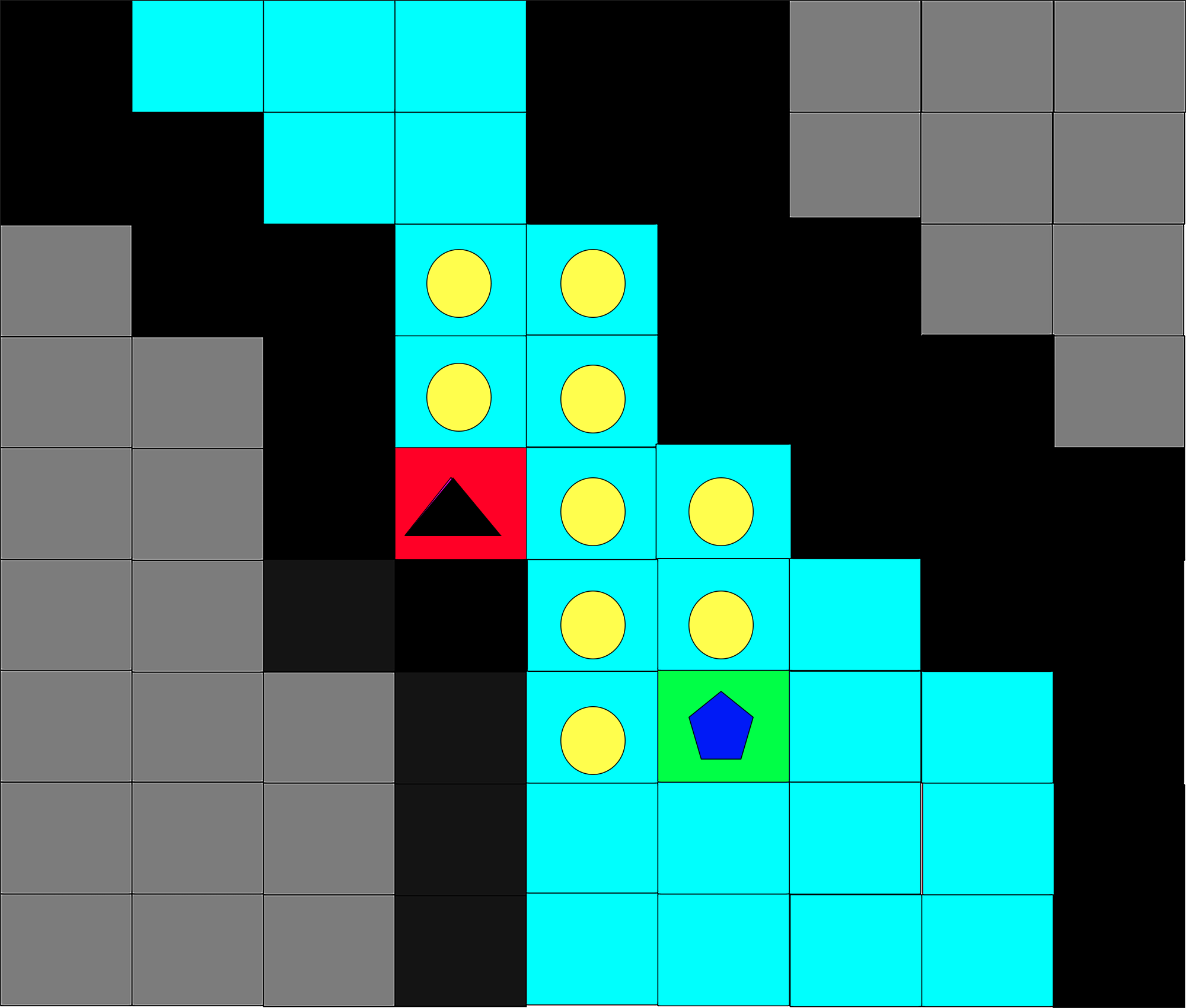} %
    \caption{Process for replacement of a non-viable $tp$ for $r=2$ and $l=1$. Triangle: non-viable $tp$; Circle: $Cell_{F} \in Query$; and Pentagon: $tp_{new}$.}
    \label{fig:obstacleavoidance}
\end{figure}

Fig. \ref{fig:obstacleavoidance} shows an example of this process. Algorithm 2 successively evaluates each $Cell_{F} \in Query$ (cells with circles and pentagon markers), all within $r$ cell lengths of non-viable $tp$ (red cell with triangle marker) with the aid of Algorithm 3 and selects the one with the lowest cost (cell with pentagon marker). This is the only $Cell_{F} \in Query$ completely surrounded by other $Cell_{F}$ within $l$ cell lengths of itself. If multiple $Cell_{F} \in Query$ share the lowest cost value calculated, the first $Cell_{F} \in Query$ to be evaluated at that cost value is chosen for $tp_{new}$.  It is important to view $f$ and $g$ as hyper-parameters that try to balance the performance task of obstacle avoidance and adherence to the chaotic path planner\color{black}. Developing a more standardized objective fitness function to judge the performance of these algorithms can be the subject of a future study.

\begin{figure} [h]
\centering
     \subfloat[\centering ]{{\includegraphics[height=5cm]{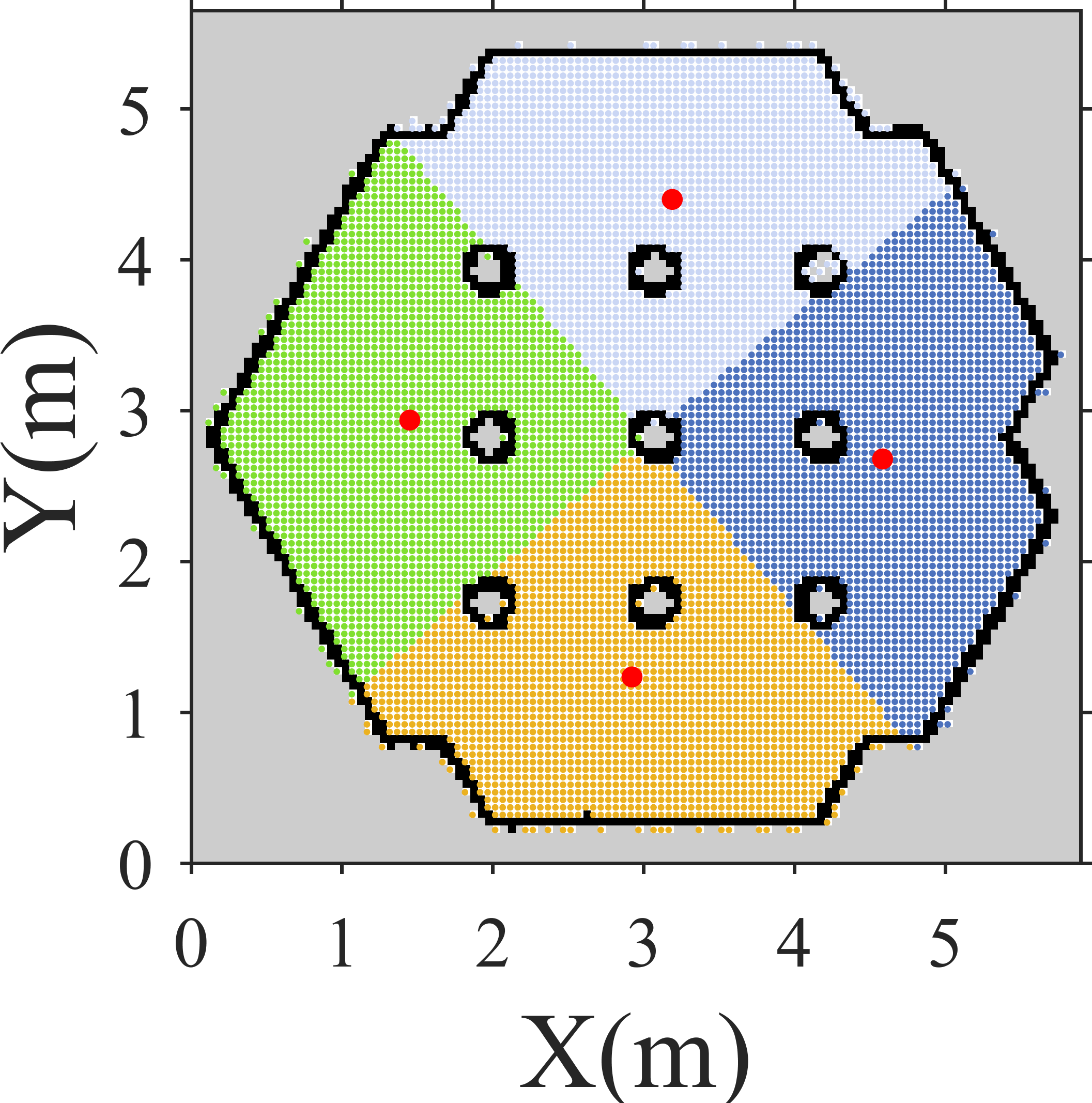} }}%
        \hspace{0.05\textwidth}
    \subfloat[\centering ]{{\includegraphics[height=5cm]{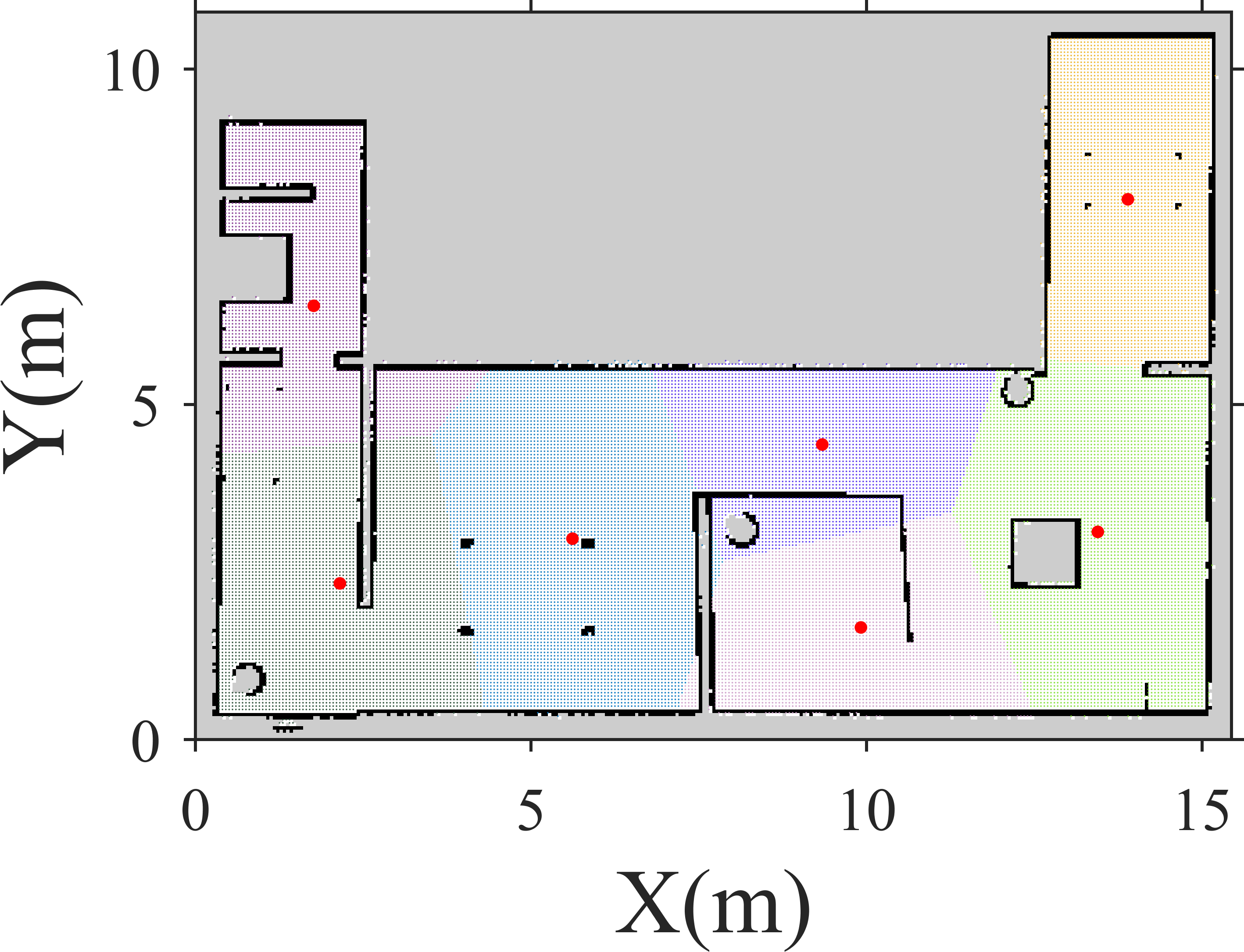} }}%
     \hspace{0.05\textwidth}
    \subfloat[\centering ]{{\includegraphics[height=6cm]{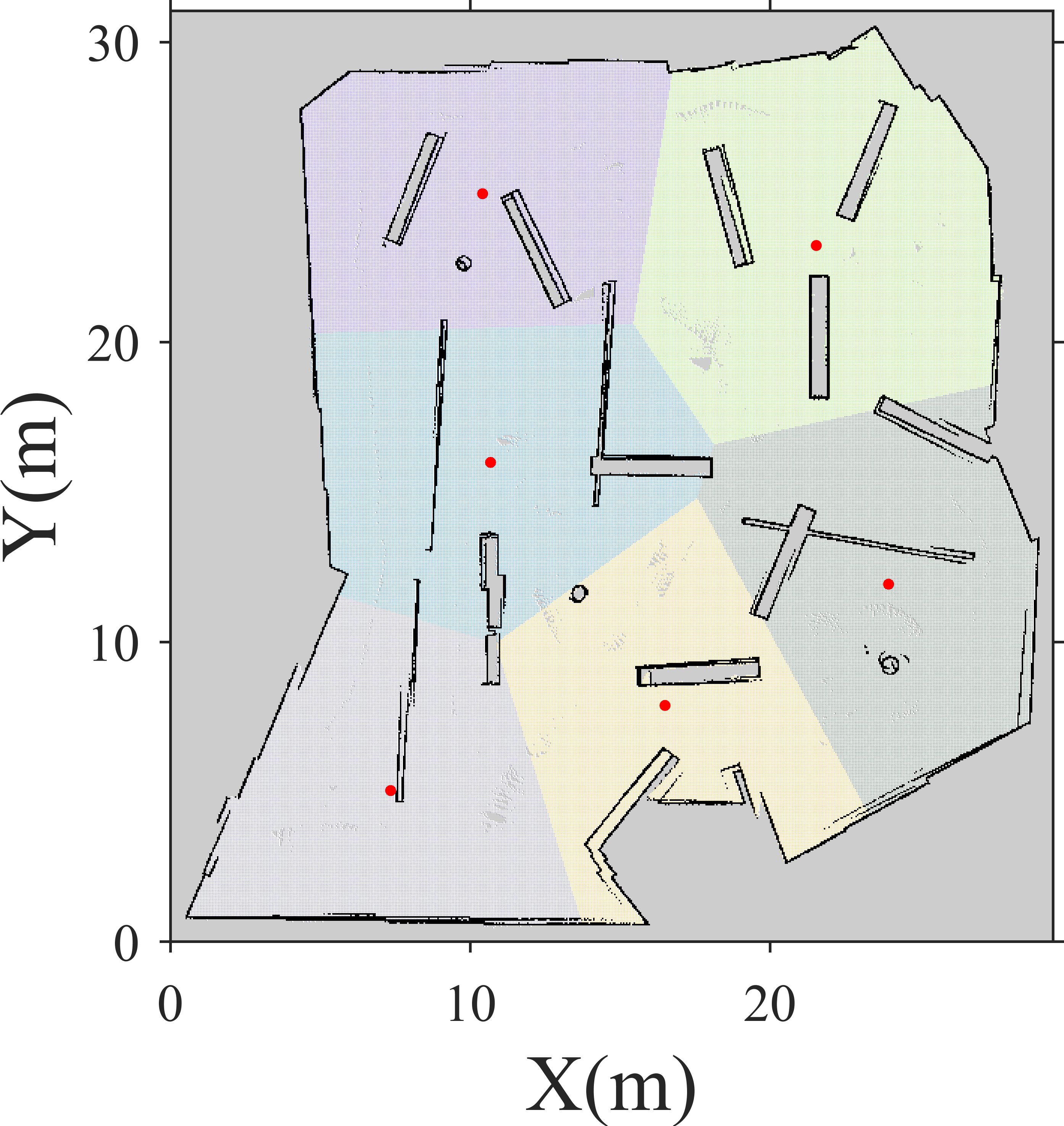} }}%
    
    \caption{Map-zoning technique showcased on maps of environments of various sizes shapes and obstacle configurations. Filled circle: centroid of each zone; same-colored small dots: a specific zone.}
    \label{fig:mapzoning}
\end{figure}

\subsection{Map-zoning Technique For Effective Dispersal of Chaotic Trajectories}
\label{sec:Map-zoning Technique For Effective Dispersal of Chaotic Trajectories}
This technique uses an unsupervised machine-learning clustering algorithm to break the map into individual zones, dispersing chaotic trajectories across the map for more effective scanning and reducing coverage time in large and/or complex environments. The clustering process takes place before the start of the coverage; it groups the $(X,Y)_M$ coordinates of all $Cell_F$ into different zones. The clustering algorithm was obtained from the scikit-learn Python library. The library offers agglomerative and K-means methods for clustering. We chose the k-means method for the following reasons:

(1)  K-means is more biased towards isotropic blob shaped clusters which often leads to more consistent production of clusters of the same relative size given a uniform distribution of data points. Although agglomerative clustering methods might potentially be more ideal for certain shapes of environments, this even distribution of cluster size generally favors our use case for efficient coverage of a wider range of environment types.

(2) K-means has a relatively simpler time and computational complexity. Agglomerative methods like Ward could have provided the same results, however for the bigger maps the computational complexity becomes overwhelming and sometimes leads to the termination of the program. While a more powerful computer could handle this method more effectively, K-means suited our purposes in the final analysis and specially in the experimentation. 
It is important to note that in environments like the one represented in Fig. \ref{fig:mapzoning} (b), evenly sized clusters may not always be favorable. In Fig. \ref{fig:mapzoning} (b), one of the clusters is split into two parts by a wall. It is worth a mention that the various methods devised in agglomerative clustering would still run into the same problem. For the most optimal performance, all clusters should be compacted into one concentrated area. Developing a custom supervised clustering method, tailored for our specific purposes, can be a focus of a future study. We briefly discuss this line of thought in Section \ref{sec:Conclusions}.

\subsubsection{Decision Making Process for Chaotic trajectory Proliferation\color{black}}
\label{sec:Decision Making Process for Chaotic trajectory Proliferation}
The parameters of the k-means algorithm were set to enable sensible centroid assignment and cluster formation (see Figs. \ref{fig:mapzoning}(a)-(c)) within a reasonable computation time. The centroid of a cluster represents the midpoint ($(X,Y)_{zone}$) of the zone. Each zone's midpoint and coverage are stored in memory so that the coverage is continuously updated as the robot traverses the map, and the $(X,Y)_{zone}$ of the least covered zone can be published by Algorithm 4. If there are multiple zones with the least coverage, the $(X,Y)_{zone}$ of the closest distance $d$ to the robot's current position is published. Matrix $M_Z$ stores information about the individual zones, with columns containing the following data from left to right: $X_{zone}$, $Y_{zone}$, the number of $Cell_F$ in each zone, the number of $Cell_F$ covered in a zone, and the zone's coverage rate ($c_z$). Algorithm 4 utilizes $M_Z$ to analyze which zones are least covered. The coverage of every zone is updated by the $Worker$ function (Algorithm 5), which is invoked by Algorithm 4. Algorithm 1 receives $(X,Y)_{zone}$ messages to spread the chaotic trajectories. To do so, Algorithm 1 sets the goal as $(X,Y)_{zone}$. 

Upon the robot reaching the immediate vicinity of the goal, a new chaotic trajectory is generated with ($x_o$, $y_o$, $z_o$) and $(X,Y)_{zone}$ as the IC. Algorithm 1 carries out this procedure after every $n_{iter}$ or if successive sets of $ns$ contain a high number of non-viable $tp$, which could not be replaced even at the second stage of Algorithm 1 (see section \ref{sec:Algorithm workflow})\color{black}. Examples in Figs. \ref{fig:mapzoning}(a)-(c) show the map-zoning technique used on maps of varying shapes, sizes, and obstacle densities, with each zone represented by the same-colored small dots and centroids shown with filled circles. In environments with medium to high obstacle densities, a centroid may sometimes be set close to obstacles. Because of such cases, Algorithm 1 calls Algorithm 2 to evaluate the cost (see section \ref{sec:Custom cost functions})\color{black} of the centroid  along with costs of all $Cell_{F}$ within $r$ cell lengths of it. Parameter $f$ is excluded from the cost calculations because the primary concern is the proximity of the centroid to an obstacle. As a result, the least possible cost that the Algorithm 2 can calculate is 0. This occurs when the evaluated cell is only surrounded by cells that represent free space. If the centroid shares said cost with any other $Cell_{F} \in Query$, the $Shift$ function returns the centroid as part of its output. If the centroid is not associated with this cost, the first $Cell_{F} \in Query$ to be evaluated at this cost is chosen at part of output. If the cost of no cells is 0, the $(X,Y)_{M}$ associated with the least cost is chosen. The associated $(X,Y)_{M}$ of the chosen cell is set as a goal and used for IC. The clustering method ensures that each zone is approximately the same size. Section \ref{subsec:Real-time computation technique for coverage calculation} will further discuss the role of Algorithms 4 and 5 after zones are created.

\begin{figure} [h]
   \centering
     \subfloat[\centering ]{{\includegraphics[height=4.5cm]{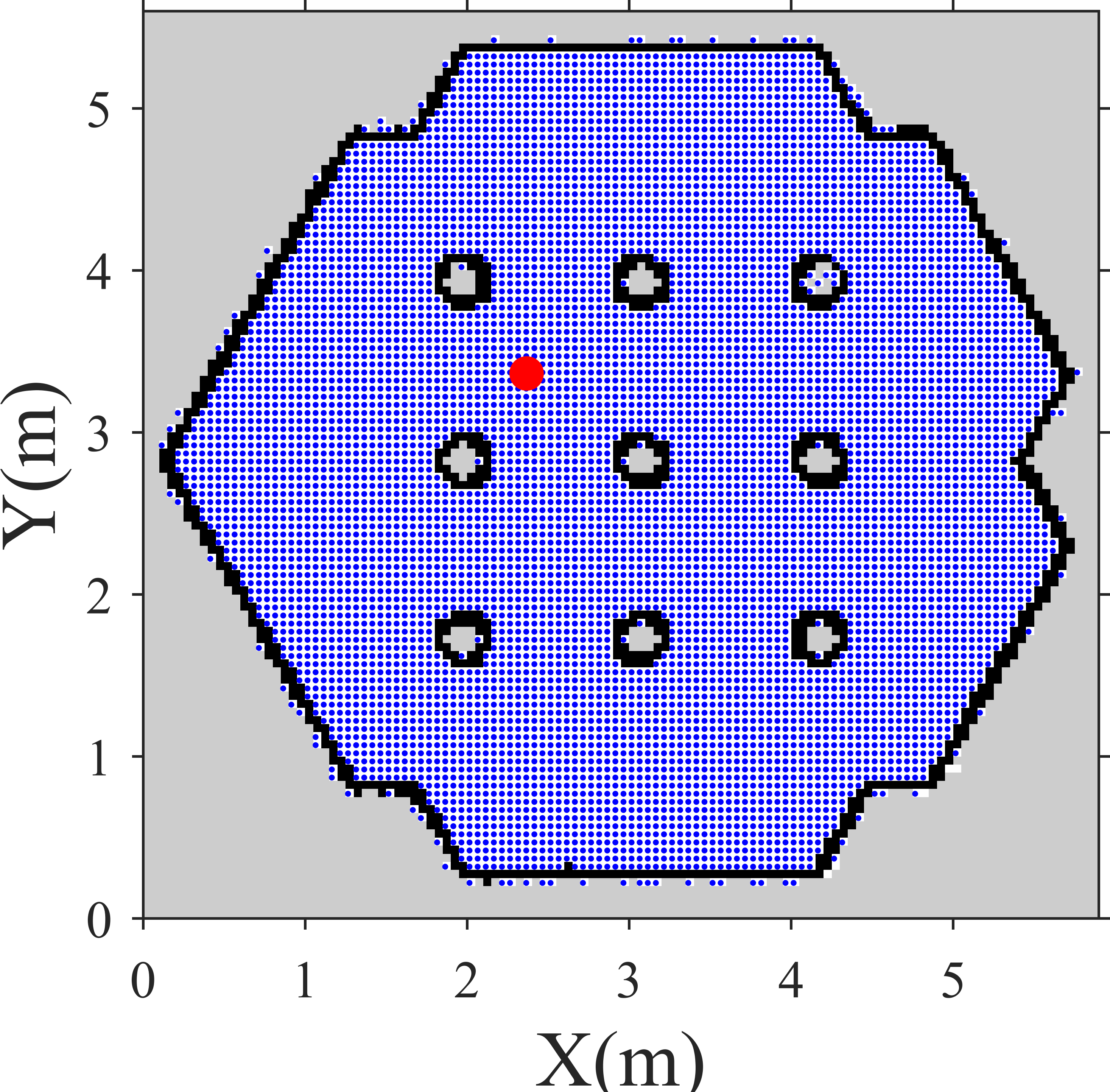} }}%
    \hspace{0.05\textwidth}
    \subfloat[\centering ]{{\includegraphics[height=4.5cm]{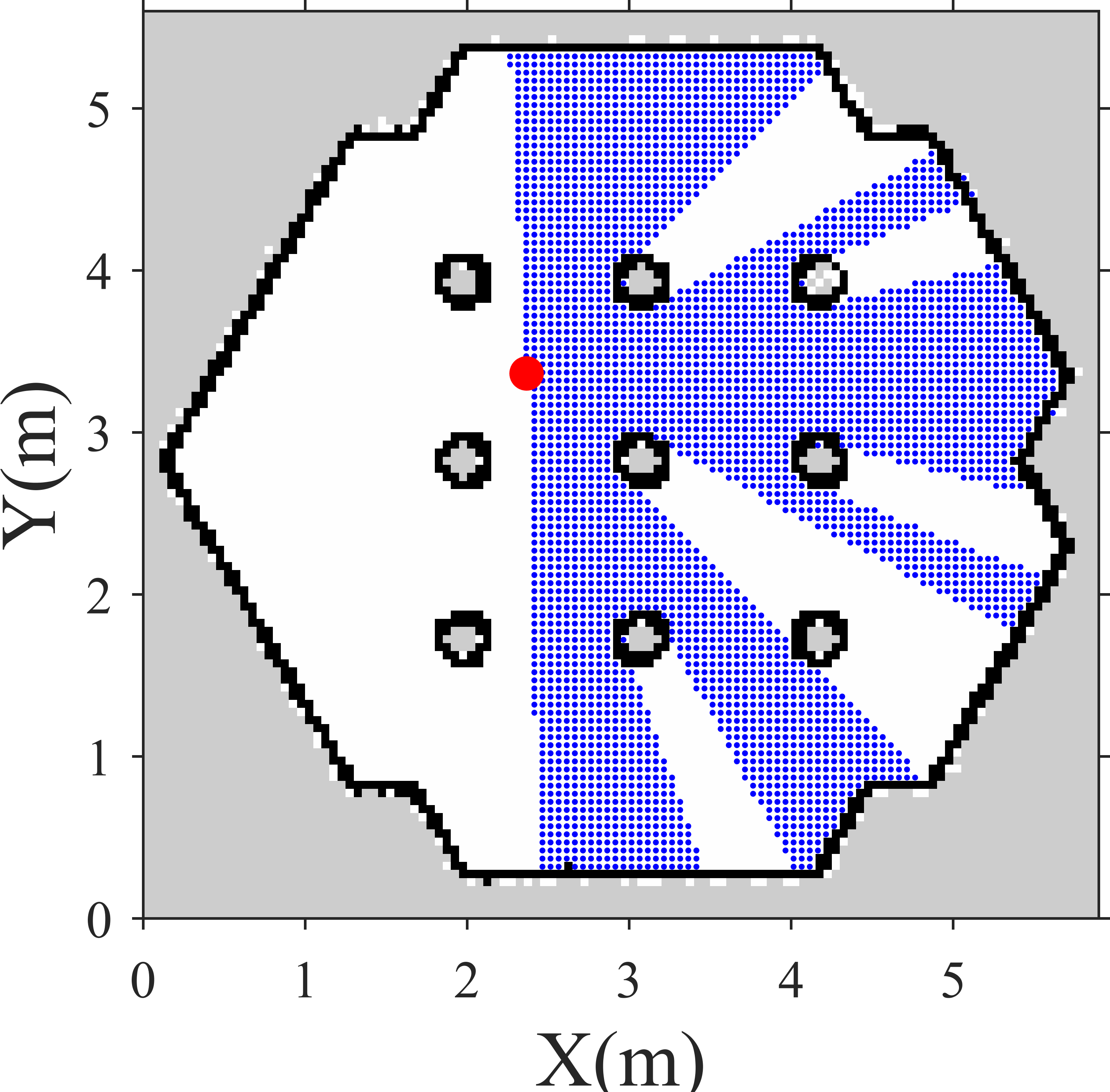} }}%
    \\
    \subfloat[\centering ]{{\includegraphics[height=4.5cm]{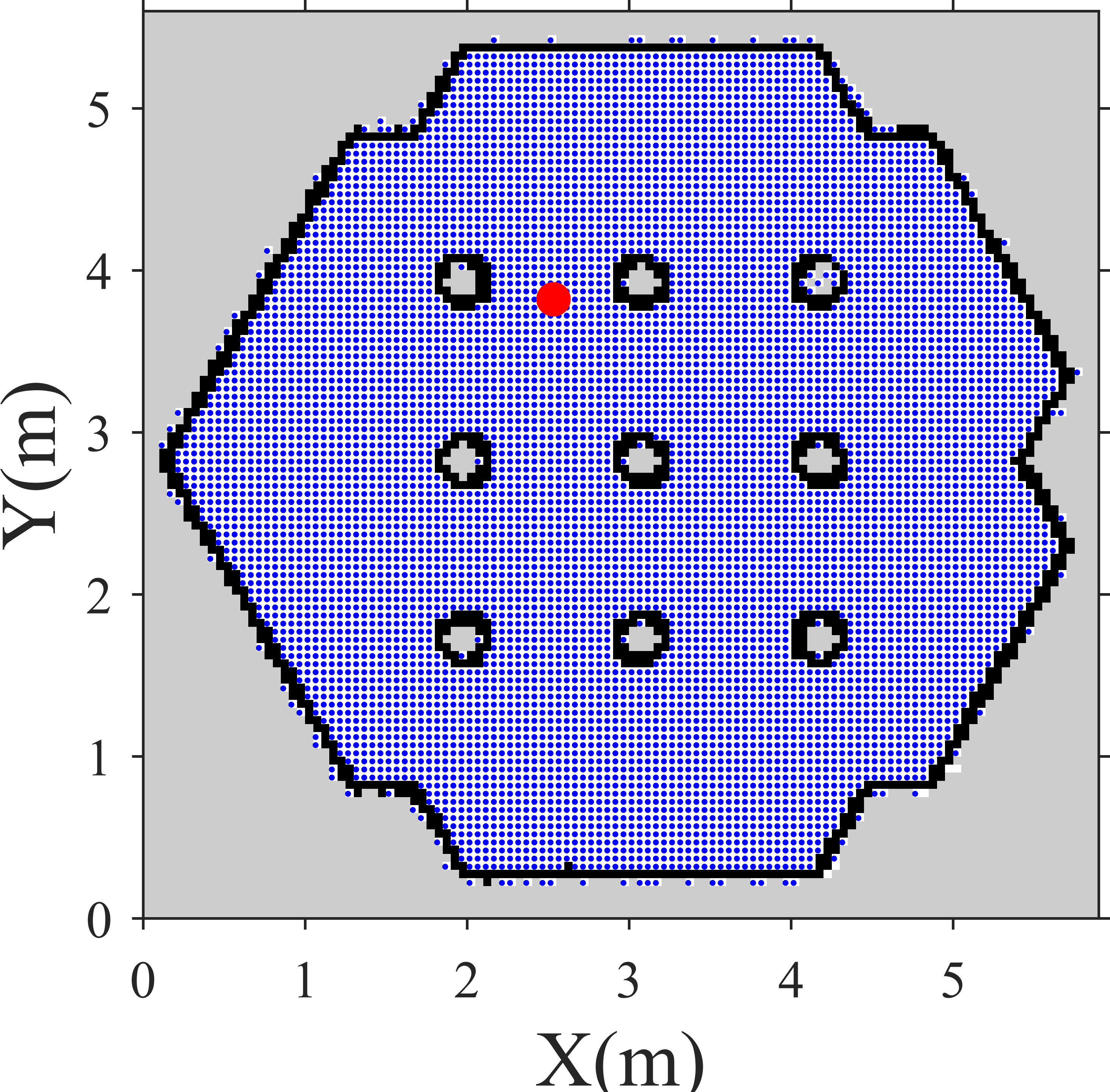} }}%
    \hspace{0.05\textwidth}
    \subfloat[\centering ]{{\includegraphics[height=4.5cm]{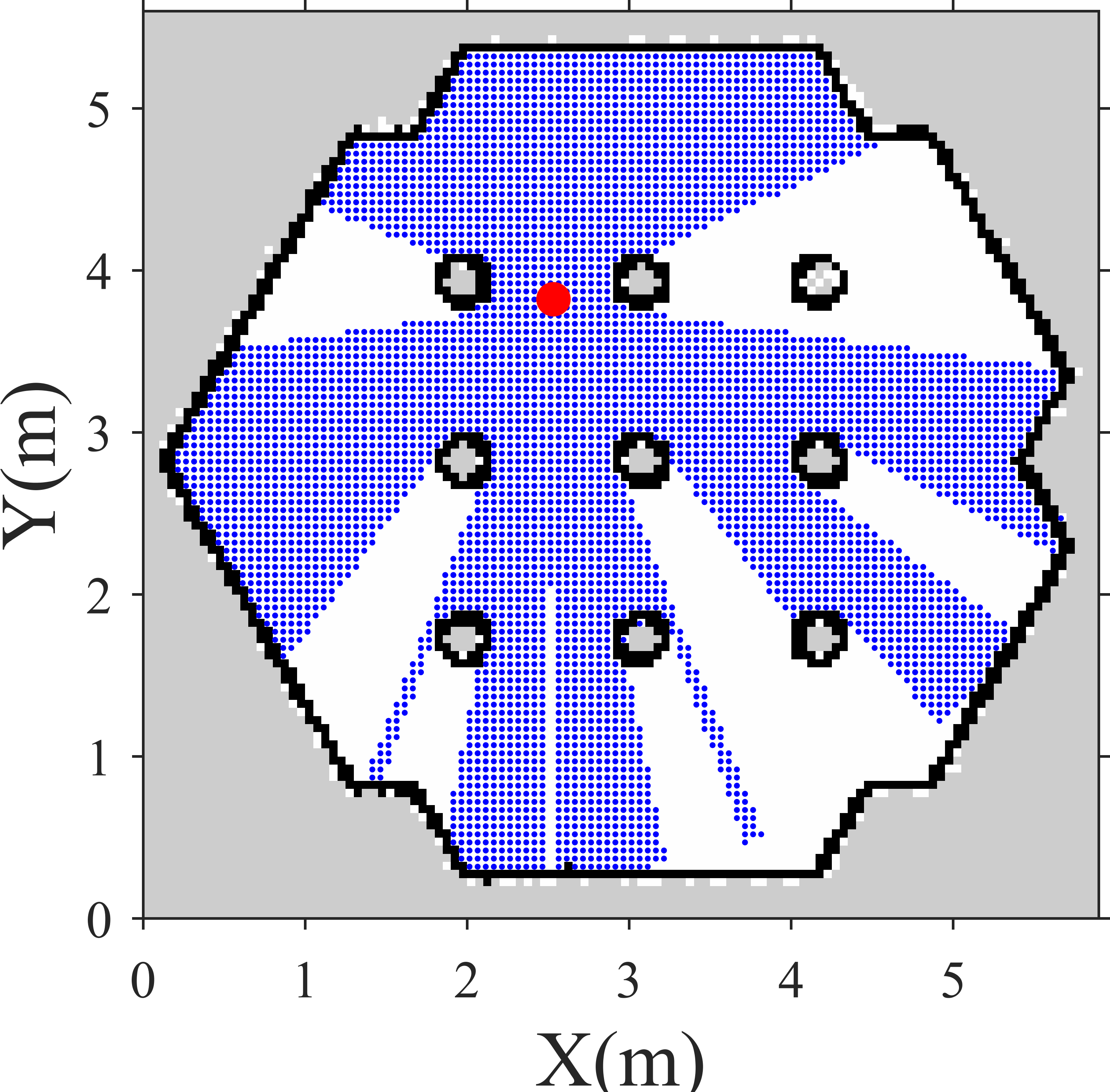} }}%
    \caption{Mapping queried cells onto sensor data: (a) and (c) show all $Cell_F \in$ Query; (b) and (d) show all $Cell_F\in$ Query that underwent successful sensor mapping with 2 different sensors. Filled circle: sensor's position in the map frame ($(X, Y)_{SM}$). Small dots: all the $Cell_F\in$ Query before and after sensor mapping.}
    \label{fig:FOVs}
\end{figure}

\subsection{Real-time Computation Technique For Coverage Calculation}
\label{subsec:Real-time computation technique for coverage calculation}
This study has developed a real-time computation technique that utilizes descritized map data (via quadtree), sensor data, and matrix data structures for fast and accurate coverage rate calculation. The quadtree provides quick access (via querying) to all $Cell_F$ within the sensing range ($SR$) of the robot's sensor at a time $t$. All $Cell_F \in$ Query are then mapped onto sensor data received at $t$ to match which $Cell_F \in$ Query is within the real sensor Field Of View ($FOV$) at time $t$. The coverage status of all $Cell_F\in$ Query which meet the matching are quickly stored to memory via matrices, and coverage is updated. All $Cell_F\in$ Query which do not meet this matching are dropped. Figs. \ref{fig:FOVs}(a)-(d) illustrate the process. Figs. \ref{fig:FOVs}(a) and (c) show all $Cell_F \in$ Query (with small dots). $SR$ is 3.5m, and is converted to cell length by multiplication with map resolution ($res$), so as to obtain the value of the querying radius as input for the quadtree algorithm. Figs. \ref{fig:FOVs}(a) and (b) correspond to a sensor with $FOV$ ranging from -1.57 to 1.57 radians and Figs . \ref{fig:FOVs}(c) and (d) represent the results for a sensor  with $FOV$ ranging from 0 to 6.28 radians. 

To perform the sensor data mapping, every $(X,Y)_{M}$ derived from all $Cell_F \in$ Query must be transformed to coordinate points in the sensor frame using a $TF_{MS}$. This is so to obtain an accurate coverage from the $FOV$ provided by the sensor. The $Coverage Calculator$ function (Algorithm 4) continuously calculates $tc$ and publishes the $(X,Y)_{zone}$ of least covered zones until $dc$ is reached. The reasons for this loop structure are as follows. (1) The relationship between the map frame and sensing frame (i.e., $TF_{MS}$) changes constantly as the robot moves around. Therefore, Algorithm 4 must continuously receive new $TF_{MS}$ information. (2) All the information provided to Algorithm 4 at each iteration must be time synchronized to ensure accurate coverage. This means that sensor data and the sensor position in the mapframe ($(X,Y)_{SM}$) at the time $t$, at which sensor data was received, are fed to Algorithm 4. $TF_{MS}$ is time synchronized with this data as well. This way, Algorithms 4 and 5 are able to accurately recreate the $FOV$ at a specific $t$. In subsection \ref{sec:Algorithm flow2}, all time synchronized data will be designated a subscript of $t$. The time interval between each iteration of Algorithm 4 is infinitesimal, providing continuous coverage calculation even at robot speeds of 5 m/s at an $SR$ of 3.5 m (to be discussed in section \ref{sec:Experiments and results}).

%\newpage
\subsubsection{Algorithm flow}
\label{sec:Algorithm flow2}
The $CoverageCalculator$ function (Algorithm 4) receives $TF_{MS_t}$ (published by the $tf$ package) and sensor data (published by the $sensor\_msgs$ packages). The sensor data is published as a data field \cite{sensormsgs}. 
\begin{equation}
\label{eqn:8}
\alpha = \texttt{int}(\tan{\frac{Y_{S}}{X_{S}} \times \frac{180}{\pi}}) \mod 360
\end{equation}

This data field contains the following information: (1) the array ($SC_a$) which contains the values of the scan angles of the sensor, and (2) the array ($SC_{r_t}$) which represents the scan range associated with each scan angle. It is important to note that the scan angles are converted from radians to degree values which are then rounded down to the nearest integer (similar to the method seen in Eq. (\ref{eqn:8}). This enables matching of the orientations of cells to the scan angles. The $Coverage Calculator$ function (Algorithm 4) uses the sensor's position in the mapframe ($(X,Y)_{SM_t}$), $TF_{MS_t}$, $SC_{r_t}$ and $SC_a$ as inputs. Algorithm 4 provides all $Cell_F \in$ Query, $TF_{MS_t}$, and $SC_{r_t}$ as inputs to the $Worker$ function (Algorithm 5). The $Worker$ function utilizes $TF_{MS_t}$, $SC_a$, and $SC_{r_t}$ to map each queried cell. 

A cell is successfully mapped by meeting two criteria. First, the orientation ($\alpha$; calculated via Eq. (\ref{eqn:8})) is within the scan angles of the robot’s sensor. $\alpha$ is the orientation of the point coordinate of the cell ($(X, Y)_S$) in the sensor frame. 
On passing the first criteria, the second criteria involves the distance ($dist$) of the $(X, Y)_S$ of cell from the sensor. This distance should be less than the scan range at the corresponding scan angle. All cells which successfully pass the two criteria are accounted to memory using the storage matrix $M_C$. The $M_C$ matrix stores the coverage status and zone identification of every $Cell_F$ in the occupancy-grid map in order to ensure that covered cells aren't counted twice and the coverage of each cell counts towards the coverage of the zone assigned to it.
\begin{equation}
\label{eqn:9}
M_{Z}(zid,4) = \frac{M_{Z}(zid,3)}{M_{Z}(zid,2)}\times(100)
\end{equation}

\begin{equation}
\label{eqn:10}
\mathrm{Total\ Coverage\ rate}\ (tc) = \frac{\sum_{n=0}^{n_z-1}M_Z(n,3)}{\text{Total number of } {Cell}_{F} (\in \text{\small Occupancy-grid map})} \times 100
\end{equation}

\begin{algorithm}[htbp]
   \caption*{\normalfont\textbf{Algorithm 4: }$CoverageCalculator()$}
  \begin{algorithmic}[1]
    \WHILE{$Total \ Coverage \ rate < dc$}   
        \STATE   $|TF_{MS},(X,Y)_{SM}, \ sensor \ data|_{at  \ time \ t} \leftarrow$ ROS tf and sensor$\_$msgs package
        \STATE Query every $Cell_F$ within $SR$ of $(X,Y)_{SM_t}$ $\leftarrow$ Requires use of Eq. (\ref{eqn:6}) and converting $SR$ from meters to cell length
        \STATE $Worker(TF_{MS_t},all\ Cell_F \in Query,sc_a,sc_{r_t})$
        \STATE Find zones with the minimum $c_z \leftarrow M_Z(:,4)$
        \STATE  If there is more than one such zone, choose the zone of smallest $d$
        \STATE Share chosen zone's centroid information to $ArnoldTrajectoryPlanner$
        \STATE $tc \leftarrow$ Eq. (\ref{eqn:10})
        \IF{$tc \ \geq \  dc$}
            \STATE $Stop$
        \ENDIF   
    \ENDWHILE
 \end{algorithmic}
\end{algorithm}

 \begin{algorithm}[htbp]
  \caption*{\normalfont\textbf{Algorithm 5: }$Worker(TF_{MS_t}, all\ Cell_F \in Query,sc_a,sc_{r_t})$}
  \begin{algorithmic}[1]
    \FOR{$Cell_F \in Query$}
        \STATE  $ind \leftarrow $  Eq. (\ref{eqn:3})
         \IF{$ M_{C}(ind,2) == \ 0$}
             \STATE $ (X,Y)_{M}\ of\ Cell_F \leftarrow$   Eq. (\ref{eqn:4})
             \STATE $ (X,Y)_{S}\ of\ Cell_F   \leftarrow  dot product(TF_{MS},(X,Y)_{M})$
             \STATE $\alpha \leftarrow$  Eq. (\ref{eqn:8})
             \IF{$\alpha \in SC_a$}
                \STATE $dist \leftarrow$ distance between sensor and $(X,Y)_{S}$
                \IF{$dist\ < sc_{r_t}\ at \ \alpha$}
                    \STATE$M_{C}(ind,2)=1$ 
                    \STATE  $zid \leftarrow M_{C}(ind,1)$ 
                    \STATE  $M_{Z}(zid,3)+=1$
                    \STATE $M_{Z}(zid,4) \leftarrow$  Eq. (\ref{eqn:9})
                \ENDIF
             \ENDIF
        \ENDIF
    \ENDFOR
 \end{algorithmic}
\end{algorithm}

Each row of $M_C$ stores information about an individual $Cell$. The columns of $M_C$ contain the following information from left to right: index ($ind$) values of all cells in the occupancy-grid map, zone identifier ($zid$), and coverage status. It should be noted that:  (1) all $Cell_O$ and $Cell_U$ are not assigned any zone identifiers, (2) every $zid$ represents a row of $M_Z$, and (3) every row in the coverage status column records and updates the individual coverage status of a $Cell_F$. Each cell has a column value of 0 when unaccounted, and a value of 1 when accounted. Once a cell's coverage status is updated in $M_C$, zone’s coverage rate ($c_z$) is also updated in storage matrix $M_Z$ (using Eq. (\ref{eqn:9})). Algorithm 4 then uses the fourth column of $M_Z$ to calculate the total coverage rate ($tc$) (using Eq. (\ref{eqn:10}). Algorithm 4 is continuously invoked in a loop until $dc$ is reached, at which point it will shut down and communicates as well with Algorithm 1 to shut down. 
%\newpage

\section{Experiments and results}
\label{sec:Experiments and results}
To validate the proposed methods described in this study, we conducted three experiments.
These experiments evaluate two key metrics—coverage time and thoroughness of coverage—to assess our framework's effectiveness, while potentially yielding additional insights beyond these primary measures. They are as follows:\color{black}

(1) We compared the total coverage time ($CT$) of developed CCPP application to that of the more established CPP method, boustrophedon coverage path planning (BCPP). These tests were carried out in three simulated environments of varying size, shape, and obstacle density. The robot used in these tests was a TurtleBot3 with a 2D LiDAR ($FOV$ ranging from 0 to 6.28 radians). For each environment we set a desired coverage approximating complete coverage (100\% coverage) for both CPP methods. This made it possible to compare the thoroughness of coverage across the different environments. At the same time we recorded the coverage times of both CPP methods to see if they were in any way comparable (see section \ref{sec:Comparing coverage times}). \color{black}
(2) We performed real-life testing of the algorithms in a classroom environment (see section \ref{sec:Potential advantages of CCPP over BCPP})\color{black}. The robot used in this test was a TurtleBot2 with a camera sensor (with a $FOV$ ranging from -1.047 to 1.047 radians). A video of the trial is available \footnote{\href{https://youtu.be/PYzDKsAqzLk}{Link to video}}. In this experiment, we evaluated both of the key metrics. Due to limitations in the sensor hardware, we set the desired coverage to 90\%, as explained in detail in the relevant section. We compared the coverage times of both methods, but the analysis of coverage thoroughness proved to be the more intriguing aspect. This was due to the computational drawbacks of cellular decomposition, which revealed a potential advantage of our CCPP method over BCPP. \color{black} And (3) we evaluated the computational speed of coverage calculation method. This experiment was carried out in a simulated environment. The experiment aimed to examine whether the presented coverage calculation technique could still provide real time scanning for speeds up to 5 m/s (see section \ref{sec:Test of computational speed of coverage calculation technique}). The computational speed is crucial as it affects both thoroughness of coverage and coverage time.\color{black}

An additional experiment is also showcased at the end of this section to discuss the influence of parameters (specifically $n_{iter}$ and the number of zones) on coverage time (see section \ref{Parameters influence and Future Directions}). The source code of our CCPP application is available \footnote{\url{https://gitlab.com/dsim-lab/paper-codes/Autonomous_search_of_real-life_environments}}. All simulated and real-world environments contain only \textbf{static} obstacles. In both simulated and real-life experiments, the maps of environments was generated via the SLAM (using the gmapping ROS package ~\cite{gmapping}).

\subsection{Comparing Performance of CCPP and BCPP Applications}
\label{sec:Comparing performance of CCPP and BCPP applications}
BCPP \cite{choset2000coverage,choset1998coverage} is a fairly common method for CPP used in various applications, such as cleaning \cite{ntawumenyikizaba2012online}, agriculture \cite{coombes2019flight}, demining \cite{bahnemann2021revisiting}, and multi-robot coverage \cite{rekleitis2008efficient}. The primary reasons for using BCPP for comparison are as follows. (1) Using a widely researched method in the field of CPP will provide a good benchmark for our method's performance. (2) There are open-source BCPP applications available for this comparison \cite{ethz-asl,rjj,gonzalez2005bsa,bormann2018indoor,gomez2017optimal,green,Ipiano}, compared to other methods used in CPP, which are not publicly available to our knowledge. Boustrophedon planning is a method for generating a path for a robot that alternates back and forth across an area. This planning is usually paired with cell decomposition, a technique used to divide a 2D space into simple polygons (cells). Boustrophedon planning and cell decomposition are aptly named boustrophedon cellular decomposition. Each cell represents free space. The boustrophedon paths generated to cover each cell are guided by the shape of the obstacles and boundaries surrounding said cell.
The boustrophedon cellular decomposition  method, unlike our on-line CCPP method, is therefore an off-line coverage path planning technique that requires full prior knowledge of the size and shape of the environment and all obstacles within. 

\subsubsection{Comparing CCPP against BCPP in different large simulated environments of varied complexity}
\begin{figure} [htbp]
\centering
     \subfloat[\centering ]{{\includegraphics[height=4cm]{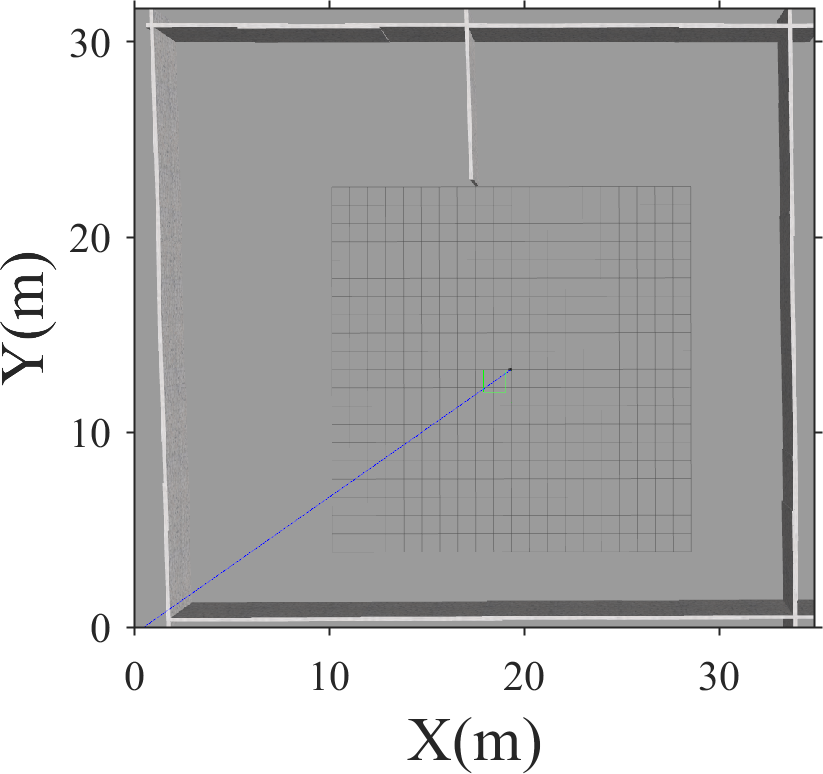} }}%
      \hspace{0.05\textwidth}
    \subfloat[\centering ]{{\includegraphics[height=4cm]{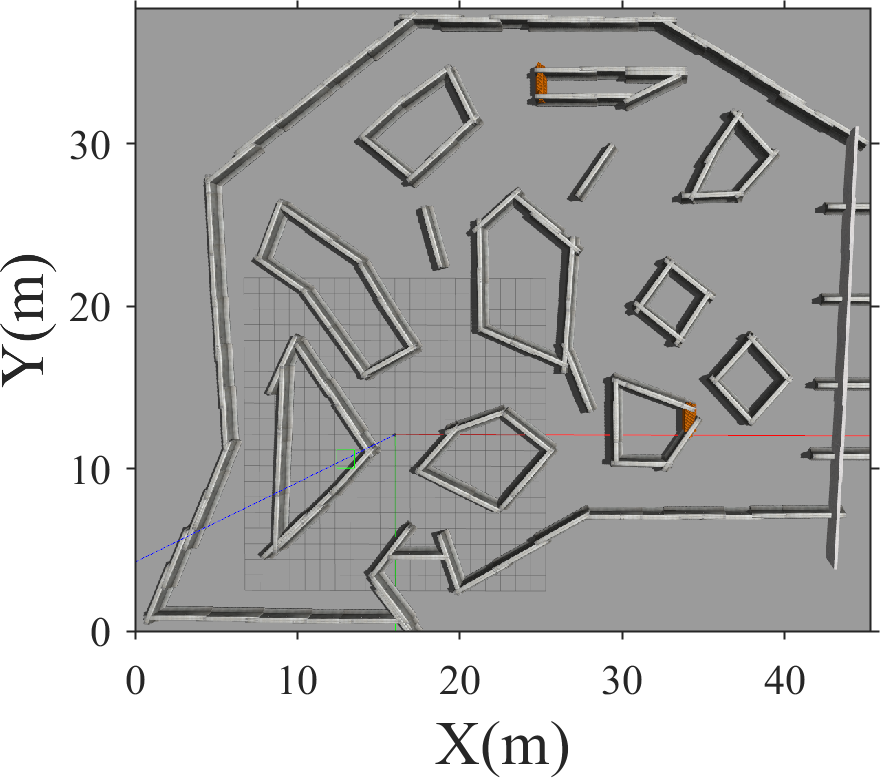} }}%
     \hspace{0.05\textwidth}
    \subfloat[\centering ]{{\includegraphics[height=4cm]{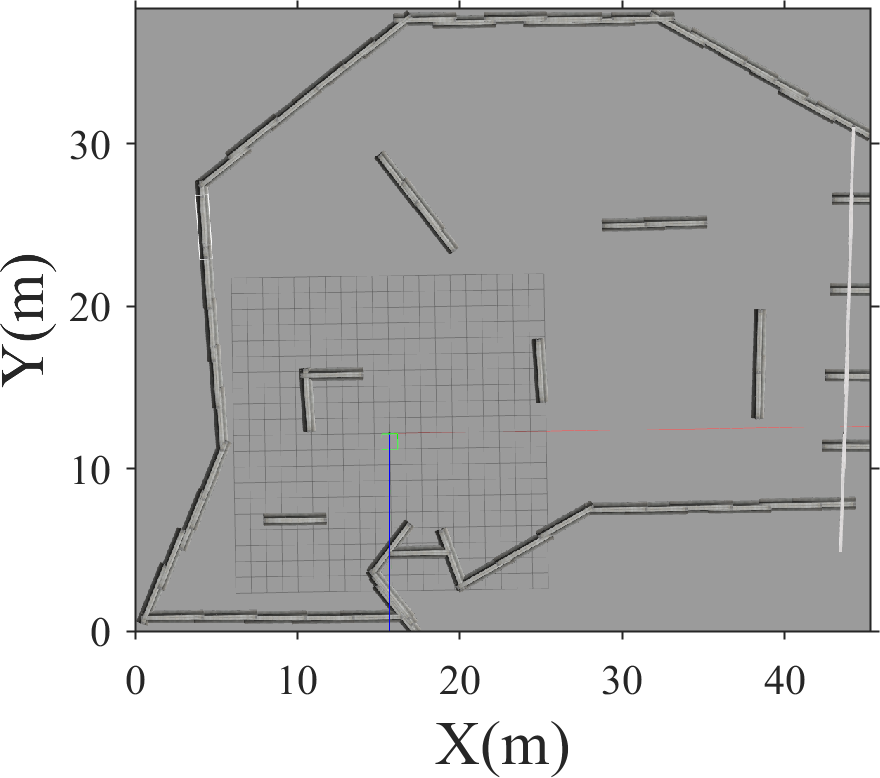} }}%
    \caption{Simulated environments: (a) Esquare, (b) Ldense, and (c) Hdense.}
    \label{fig:simulatedworlds}
\end{figure}

\color{black}
\label{sec:Comparing coverage times}
To compare the performance of our CCPP method with that of BCPP, this study used the BCPP application developed in ROS and provided in \cite{ethz-asl}; the accompanying paper is \cite{bahnemann2021revisiting}). It was difficult to operate the other open-source applications, and technical support from the contributors to those applications was lacking. It must be noted that the BCPP application used in this paper provides coverage times based on more theoretical calculations, as the coverage path plan created was not tested in a Gazebo simulation, similar to our CCPP application. %As of the time of our research, the means to recover the generated coverage time by the coverage path planner developed by \cite{ethz-asl} was not known. 
The BCPP application therefore only provides a visual representation of the generated path and an estimated coverage time. This coverage time is calculated (by the application) based on the generated path and the parameters. The procedure to provide this planned path to the simulated robot (in gazebo) to enact was not provided by BCPP algorithms. We consider three different environments shown in Fig. \ref{fig:simulatedworlds}. 

The environments are significantly larger than the robot's size, to provide large areas to cover. These environments vary in size, shape, and obstacle density to  provide a varied range of scenarios for analysis. Fig. \ref{fig:simulatedworlds}(a) shows a simple-shaped environment with one wall obstacle, while Figs. \ref{fig:simulatedworlds}(b)-(c) show complex-shaped environments with sparse and high obstacle density, respectively. The names of these environments are: Esquare (fig. \ref{fig:simulatedworlds}(a)), Ldense (fig. \ref{fig:simulatedworlds}(b)), and Hdense (fig. \ref{fig:simulatedworlds}(c)). To perform an accurate comparison, we have set the parameters for both applications to values that ensure: (1) both applications achieve an estimate of 100\% coverage rate ($tc$) for each environment. This aspect of the experiment is further discussed later. (2) Both applications adhere to the same physical constraints placed on the TurtleBot3. And (3) $tc$ is achieved within a reasonable time frame. Due to occasional inaccuracies in map generation during the SLAM, some of the obstacles are represented as containing free space. In other occasions, such spurious free space could be created outside the boundaries of the environment map. These inaccuracies occur as a result of erroneous sensor data. Any $Cell_F$ within these spurious free space could be covered during the coverage process. The compensation method for this issue will be discussed later in this section. 

\begin{figure} [htbp]
\centering
   \subfloat[\centering ]{{\includegraphics[height=4.2cm]{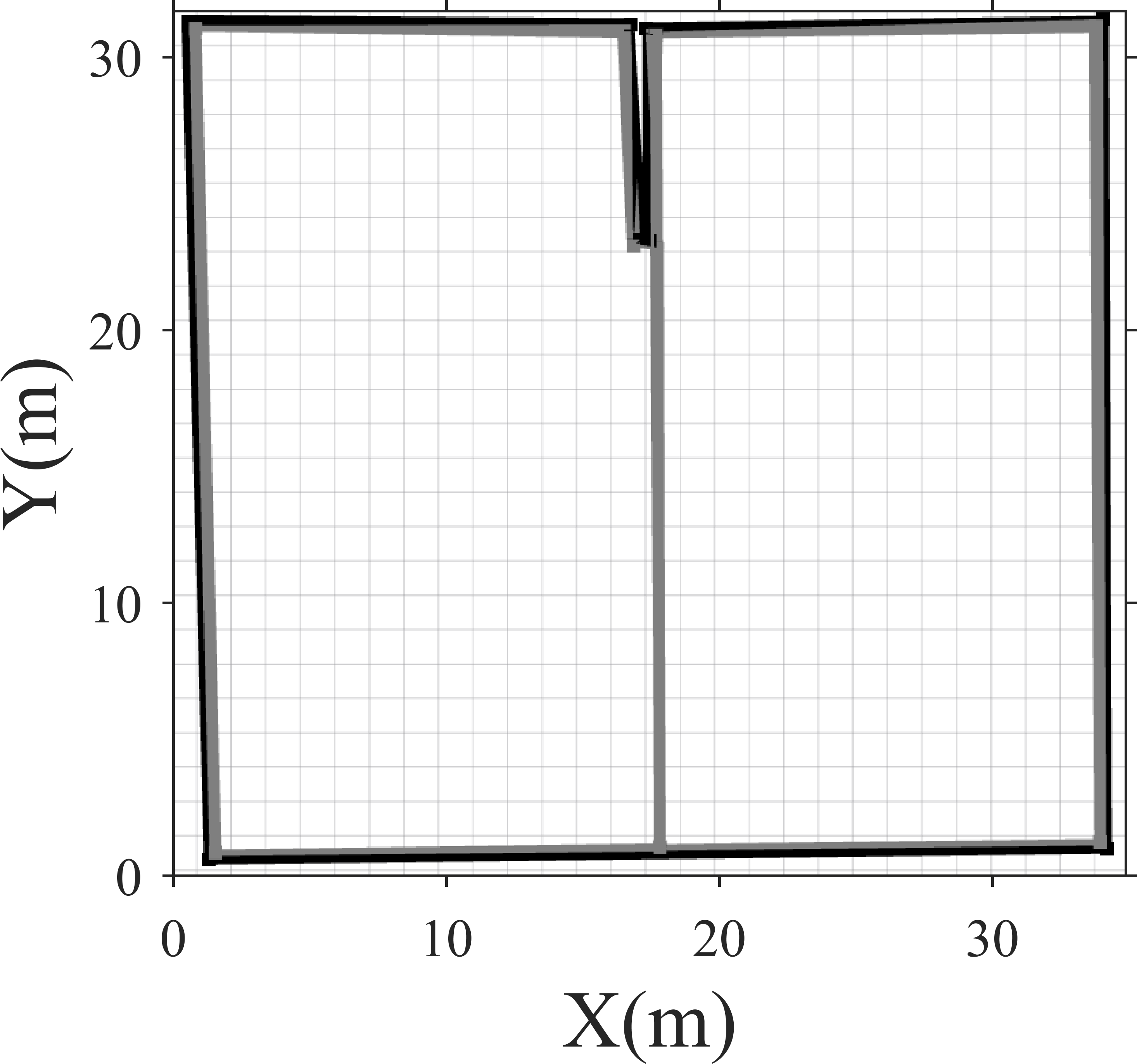} }}%  
     \hspace{0.05\textwidth}
   \subfloat[\centering ]{{\includegraphics[height=4.2cm] {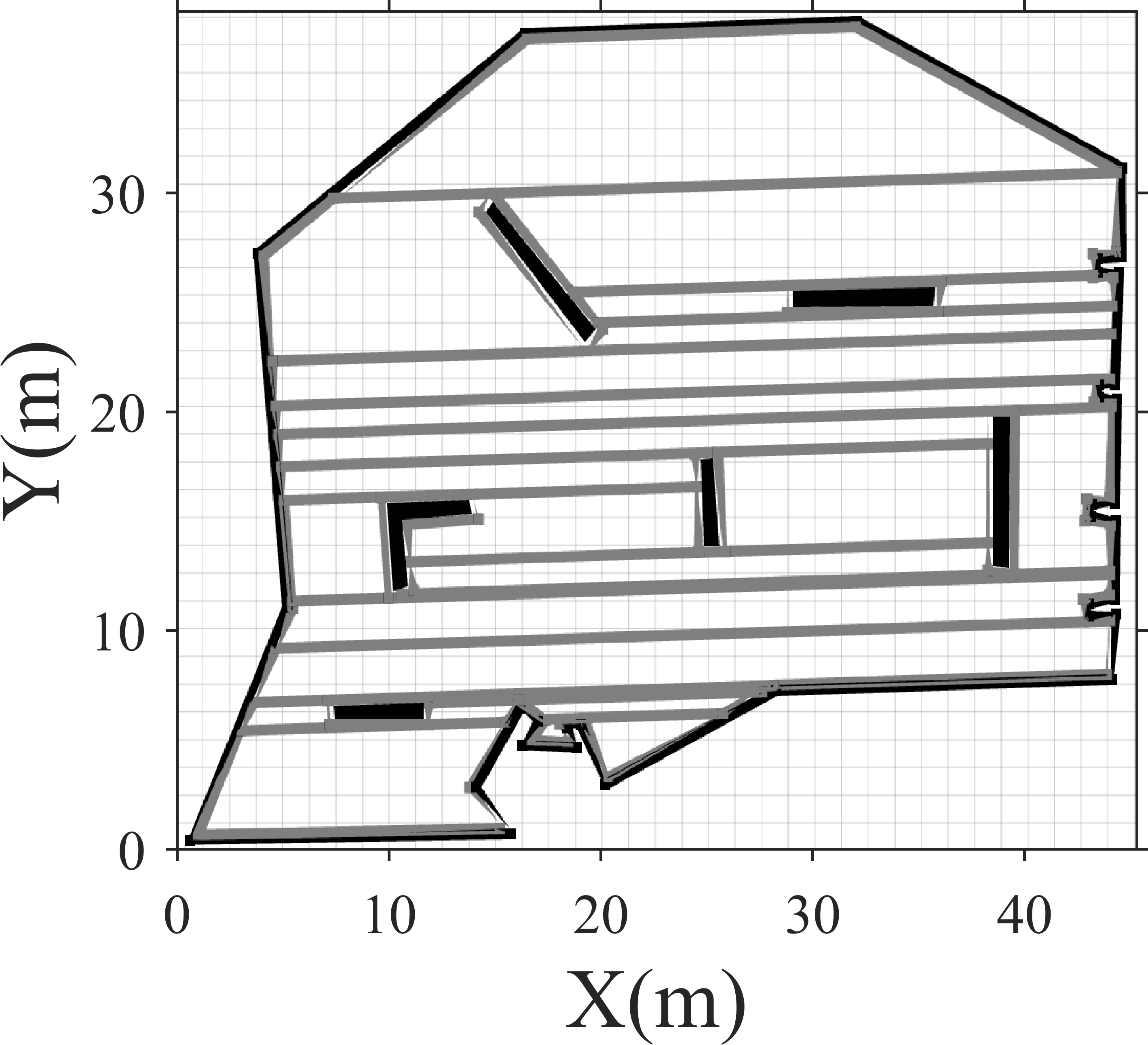} }}%
   \hspace{0.05\textwidth}
   \subfloat[\centering ]{{\includegraphics[height=4.2cm] {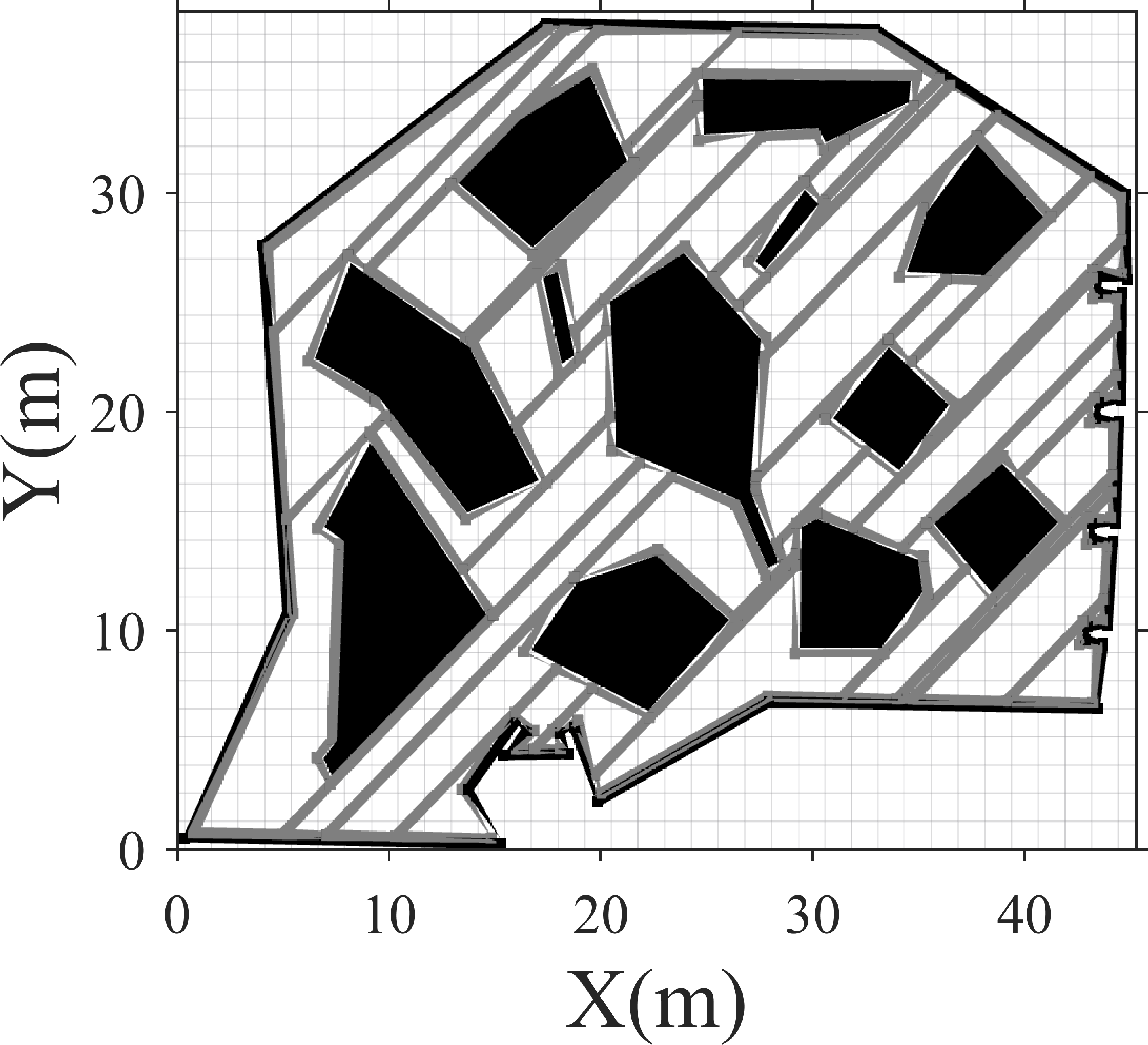} }}%  
      \\
   \subfloat[\centering ]{{\includegraphics[height=4.2cm]  {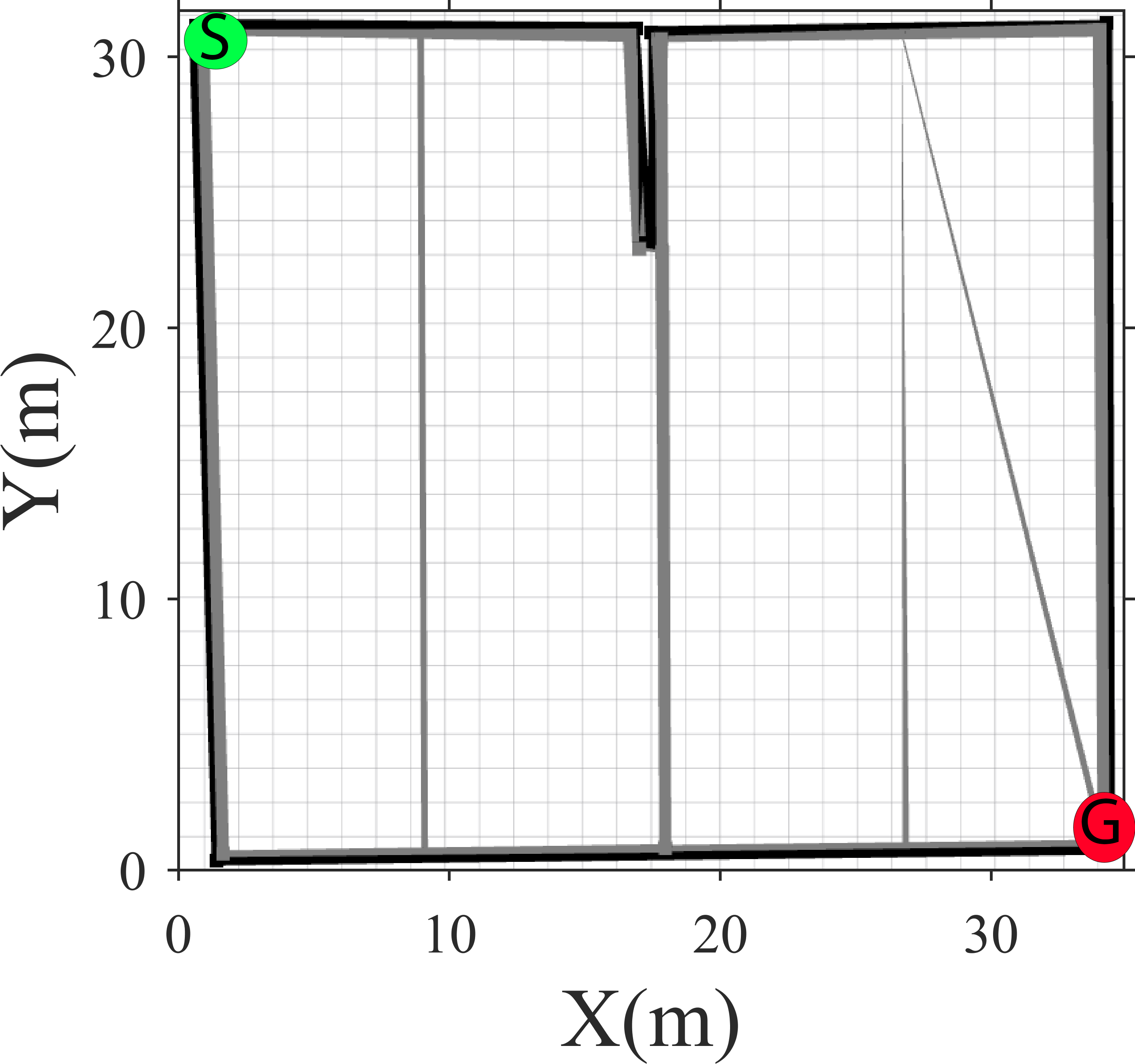} }}%
     \hspace{0.05\textwidth}
    \subfloat[\centering ]{{\includegraphics[height=4.2cm] {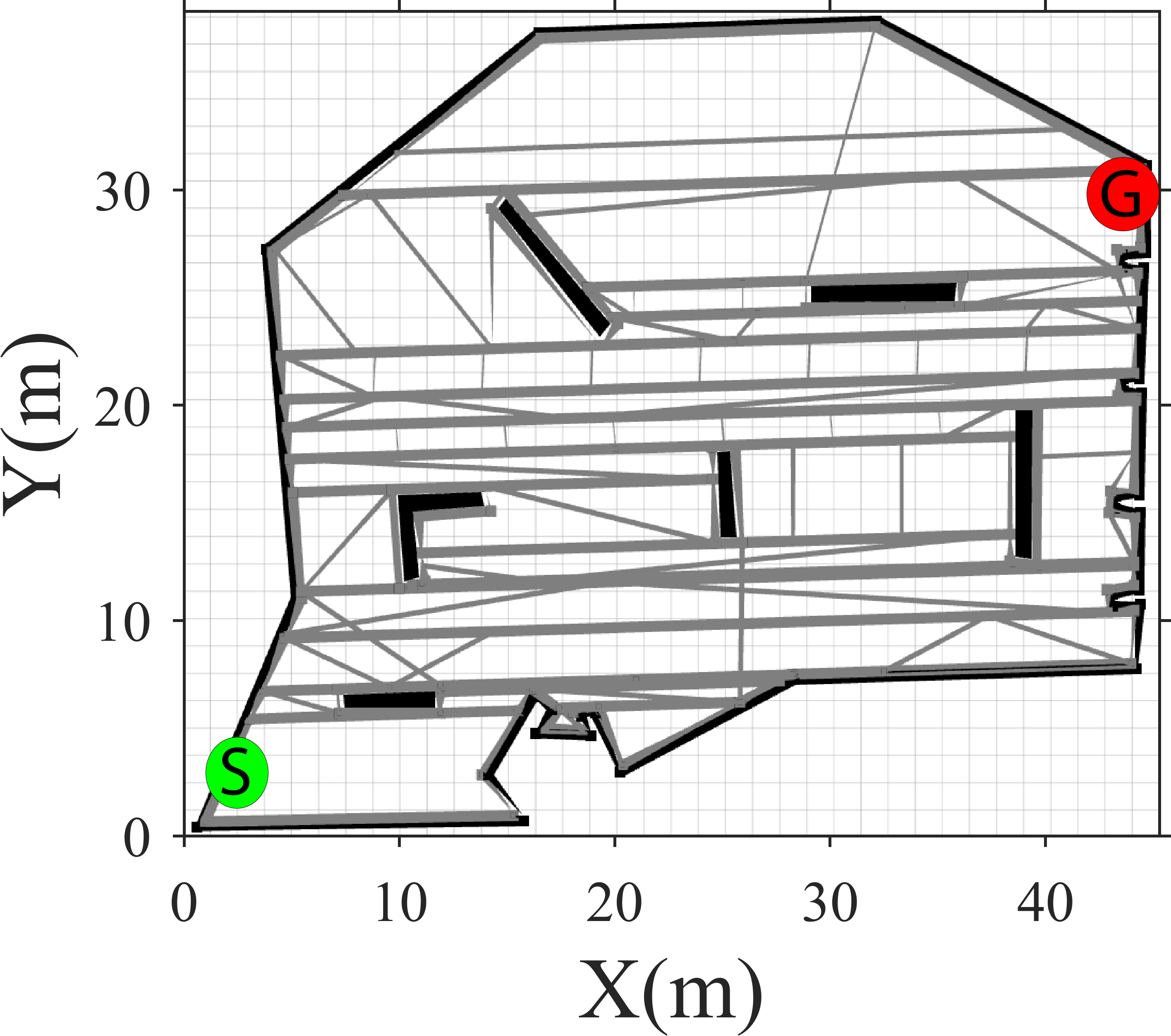} }}%  
     \hspace{0.05\textwidth}
   \subfloat[\centering ]{{\includegraphics[height=4.2cm]  {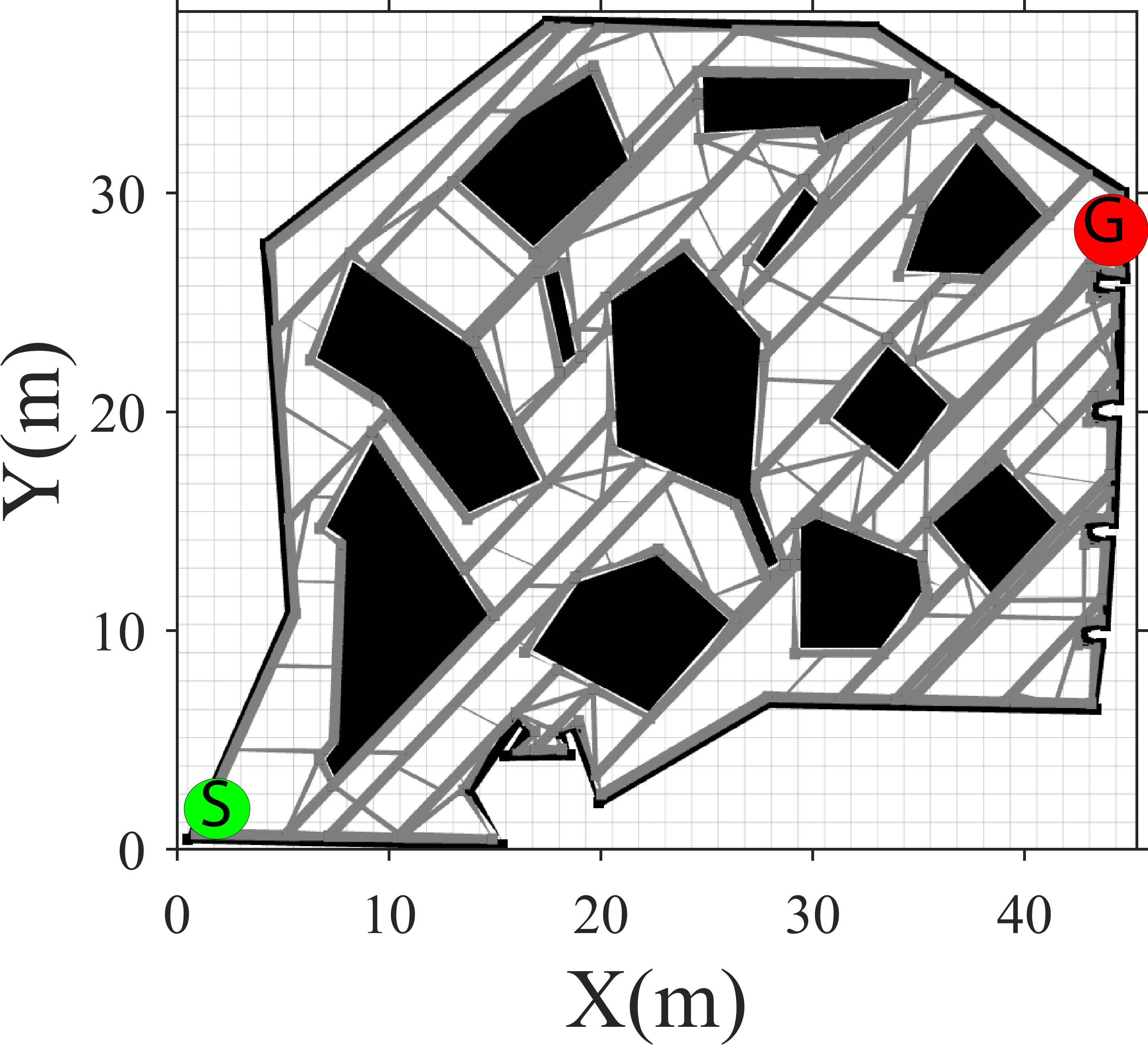} }}%
    \caption{Coverage of simulated environments using BCPP: (a)-(c) the cellular decomposition of the maps, and (d)-(f) the planned boustrophedon coverage path. S: Start position; G: Goal position.}
    \label{fig:3BCPP}
\end{figure}

For both the BCPP and CCPP applications, we set the robot's velocity to 0.22 m/s for all runs. To achieve coverage of all environments using BCPP, we set certain parameters (available in the configuration folder of the application) at constant values. These parameters are: (1) lateral overlap, which is set to 0 to maximize sweep distance and obtain the best possible coverage times; (2) maximum acceleration, set to 0.3 m/s$^2$ (the maximum acceleration of the TurtleBot3); and (3) wall distance, set to 0.2 m to account for the dimensions of the robot. The robot has a width of 0.173 m (its largest dimension). For the comparison, we varied the lateral footprint (LF) parameter for each environment. The lateral footprint parameter sets the sweep distance and is defined to be twice the sensing range ($SR$) with the robot's center at the sweep line. With this choice, there would be no coverage overlap between sweeps; it allows optimized progressive scanning of unvisited areas for BCPP application. This ensures fair comparison between the two CPP applications. The actual total coverage after the BCPP operation is not shown in the program output. To this end, the  lateral footprint is reduced as the map shape complexity and obstacle density increase, this is to ensure thorough coverage. The lateral footprint for Esquare, Ldense, and Hdense is 7.0 m, 4.0 m, and 3.0 m, respectively.  Fig. \ref{fig:3BCPP} shows the BCPP coverage for different environments. Figs.  \ref{fig:3BCPP}(a)-(c) present the cell decompositions and Figs.  \ref{fig:3BCPP}(d)-(f) showcase the robot's sweep across the environments. All the maps in Fig.  \ref{fig:3BCPP} were manually created in the application and are therefore a close representation of the actual maps (generated via SLAM). Some of the obstacles and the map boundaries have been slightly simplified to enable computation of polygons. The "S" and "G" icons shown in green and red are the start and stop positions of the sweeps.

\begin{figure} [htbp]
\centering   
     \subfloat[\centering ]{{\includegraphics[height=4.2cm]{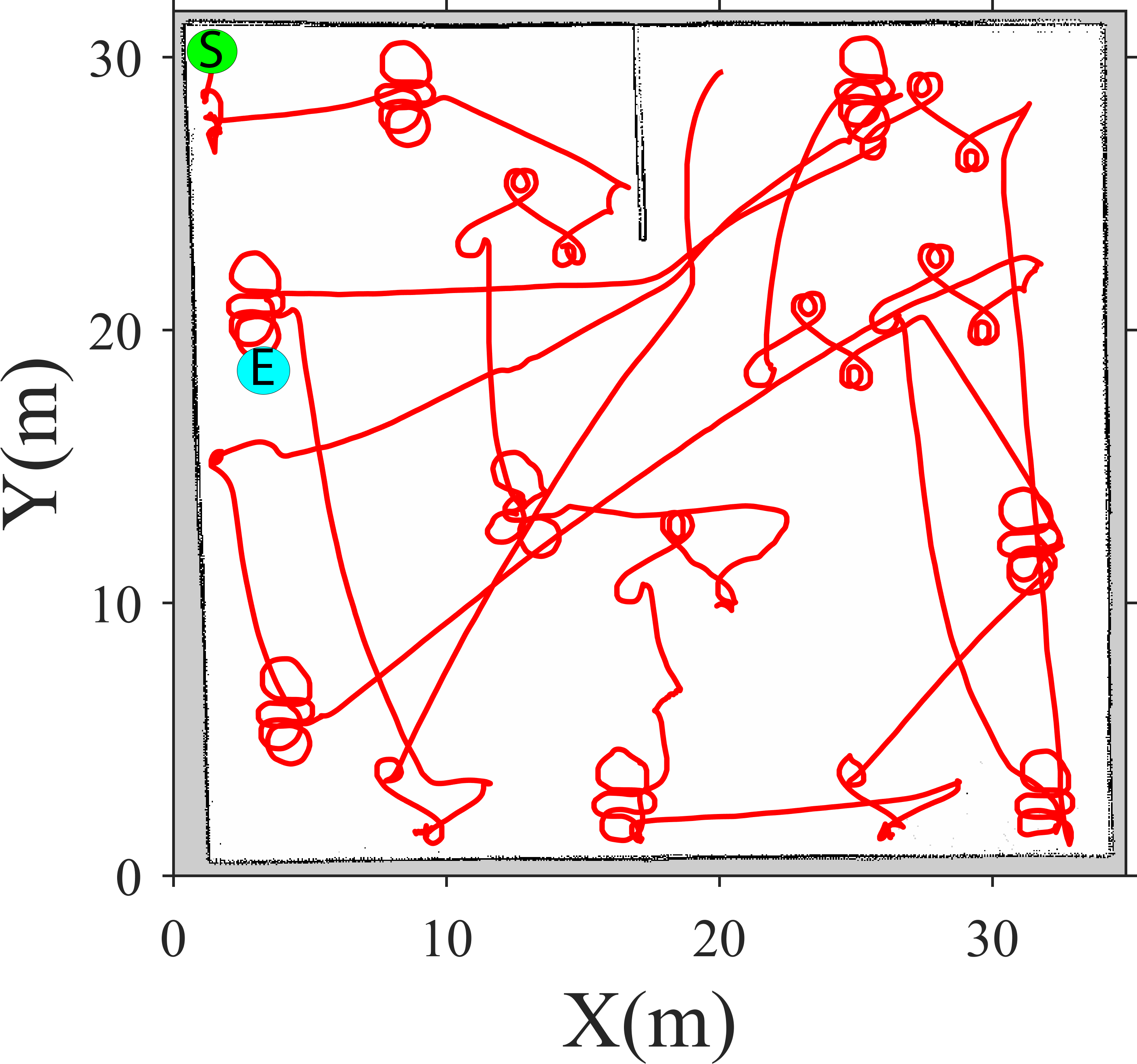} }}%  
     \hspace{0.05\textwidth}
   \subfloat[\centering ]{{\includegraphics[height=4.2cm] {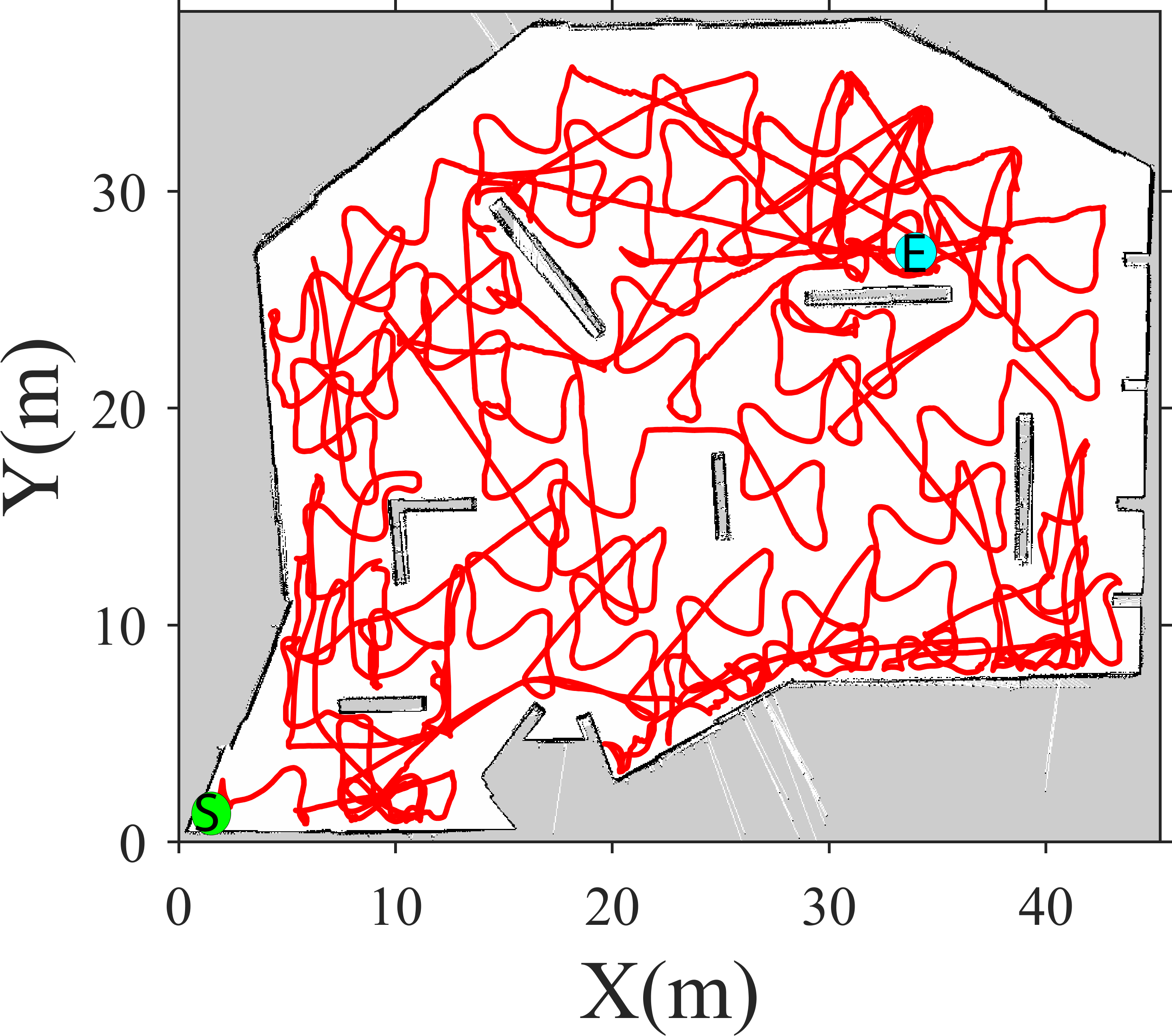} }}%
   \hspace{0.05\textwidth}
   \subfloat[\centering ]{{\includegraphics[height=4.2cm,width=4.6cm] {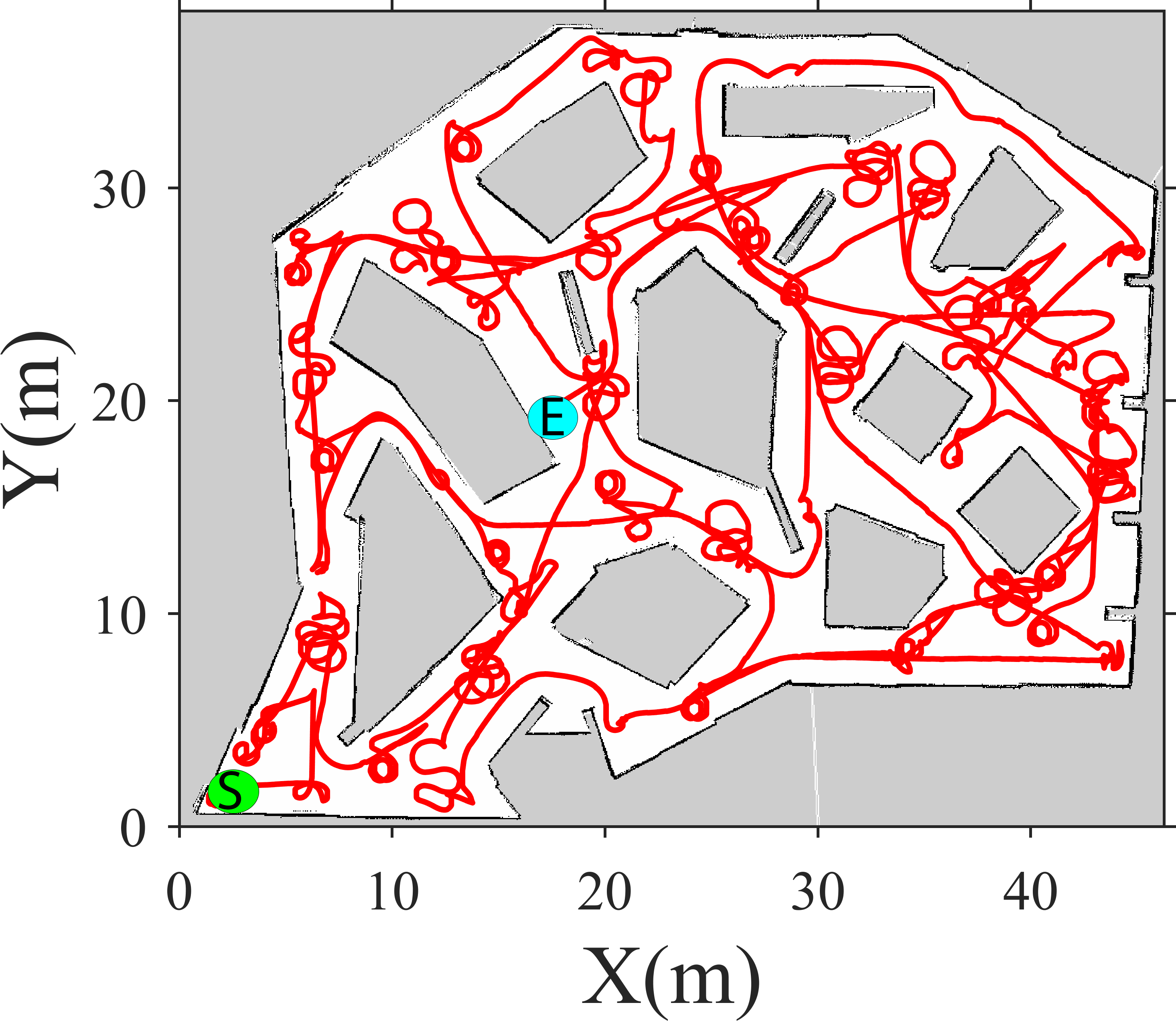} }}%  
      \\
   \subfloat[\centering ]{{\includegraphics[height=4.2cm]  {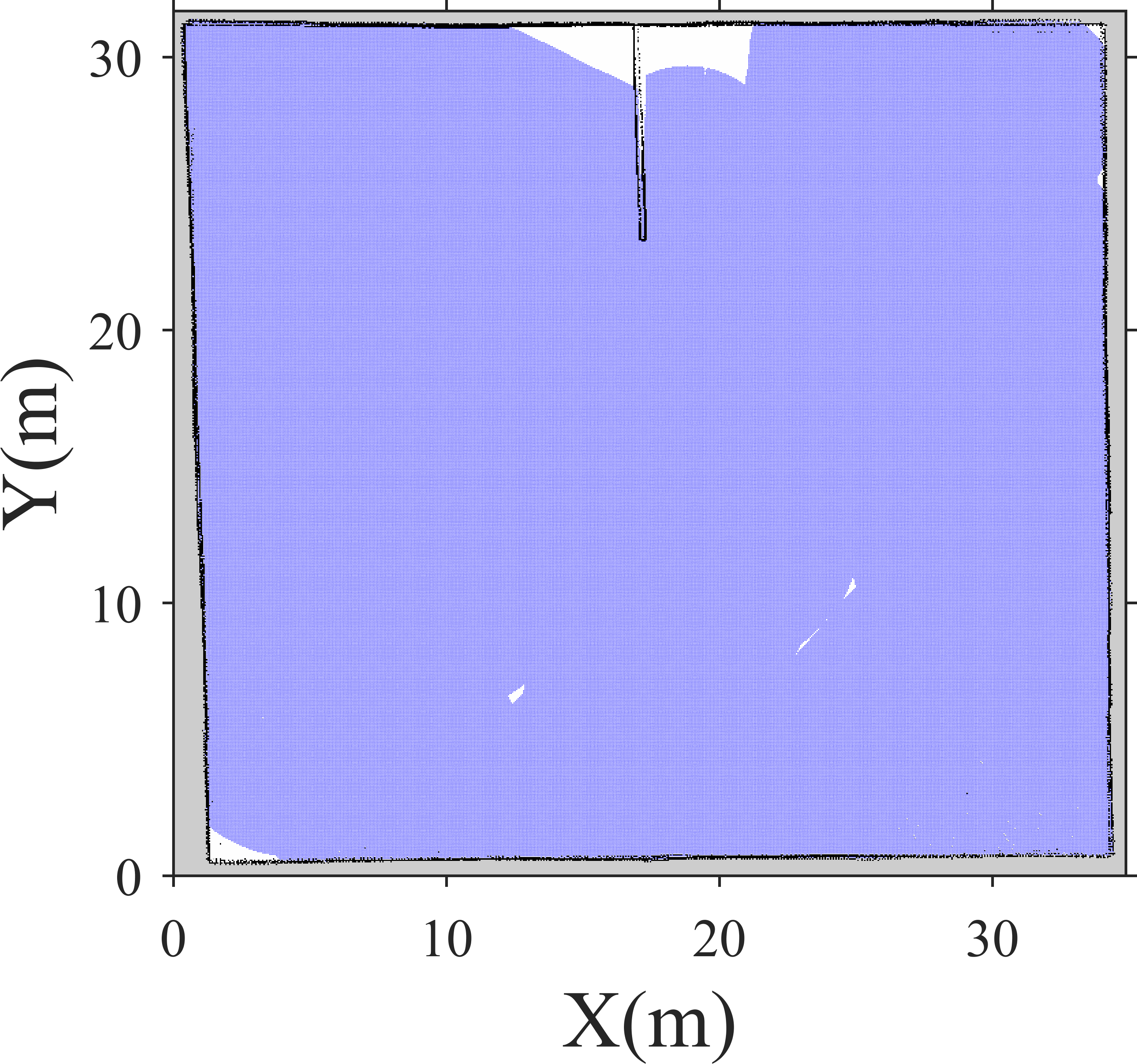} }}%
     \hspace{0.05\textwidth}
    \subfloat[\centering ]{{\includegraphics[height=4.2cm] {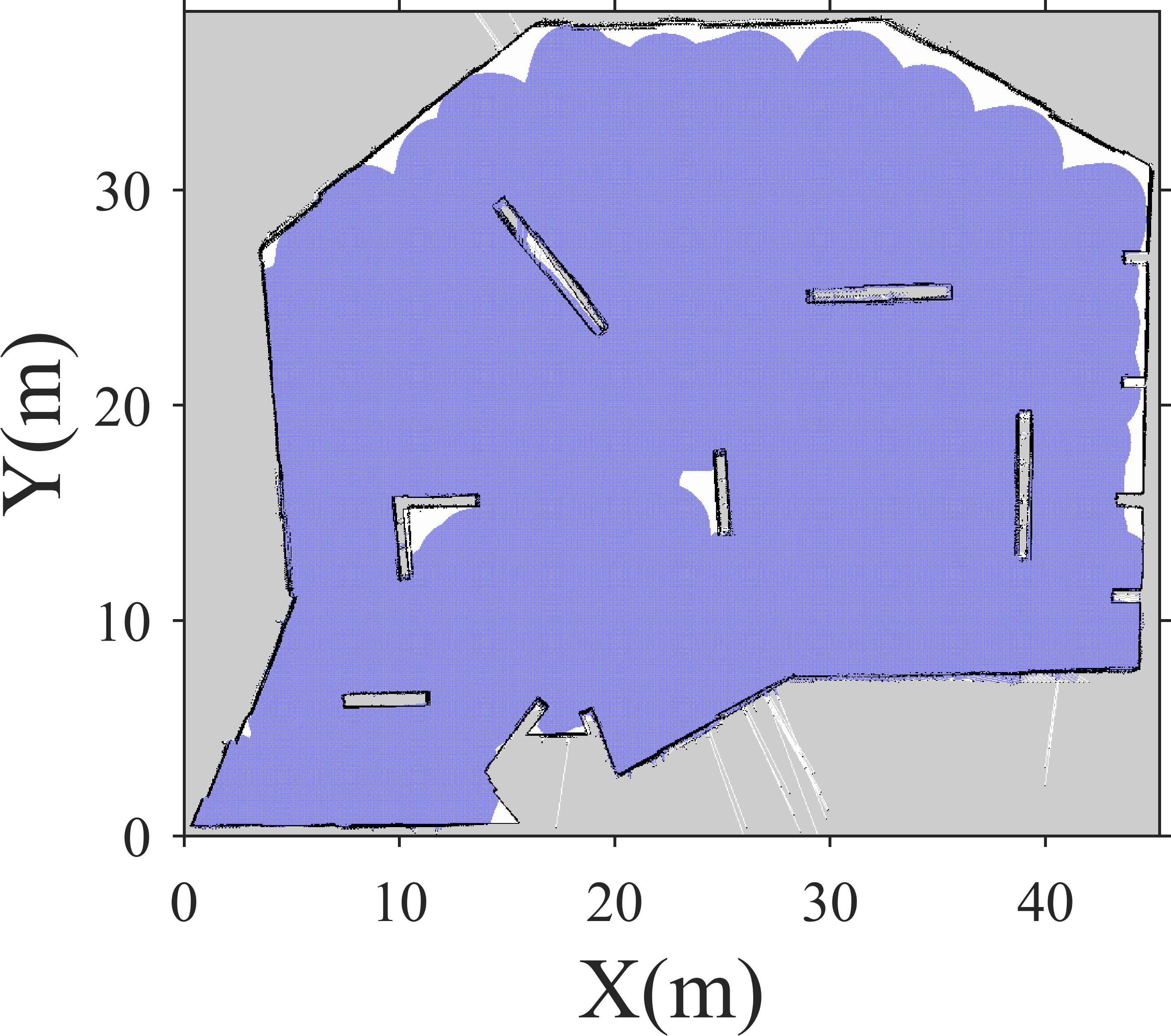} }}%  
     \hspace{0.05\textwidth}
   \subfloat[\centering ]{{\includegraphics[height=4.2cm,width=4.6cm]  {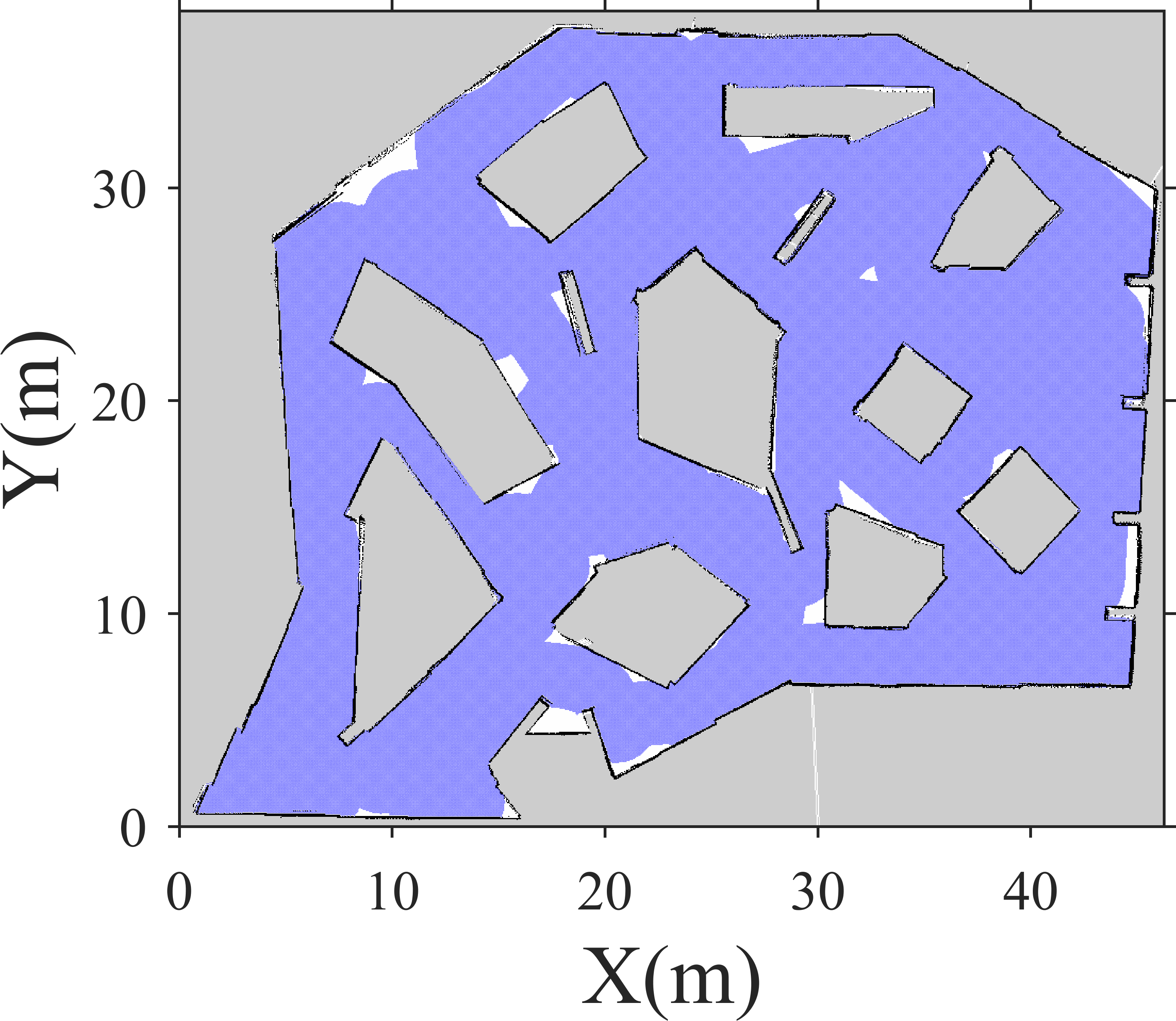} }}%
 \caption{Coverage of simulated environments using CCPP: (a)-(c) chaotic trajectories, and (d)-(f) covered cells. S: Start position; E: End position.}
    \label{fig:3CCPP}
\end{figure}

Fig. \ref{fig:3CCPP} shows the coverage across these same environments using CCPP. For CCPP, the start position of the robot in each environment is kept basically the same as those used in the BCPP application for accuracy.  The start position is shown by the green "S" icons in Figs. \ref{fig:3CCPP}(a)-(c), for convenience, the blue "E" icons mark where the robot stopped. These icons are used for all trajectory plots of the CCPP in this paper. As discussed, the SR used for each environment in CCPP application is half of the lateral footprint chosen for BCPP applications. The parameter values of zones, $n_{iter}$ and $n_s$, were chosen by trial and error. These parameters provided the smallest coverage times out of all the trial runs. Figs. \ref{fig:3CCPP}(a)-(c) and \ref{fig:3CCPP}(d)-(f) show the chaotic trajectories and covered cells for each environment, respectively. For the CCPP application, the desired coverage rate ($dc$) for every environment is set to approximately 97\% for the following reasons. (1) As discussed, the $tc$ of the BCPP operation is not outright provided as part of the output. It is difficult to visually discern the completeness of coverage. For more complex-shaped maps, $tc$ might not be 100\% as a result of the simplified polygons used to describe the obstacles. This aspect of performance drop off of BCPP is further discussed in section \ref{sec:Potential advantages of CCPP over BCPP}.
(2) By selecting a slightly lower $tc$ than 100\%, we aim to compensate for the previously discussed map inaccuracies during the SLAM that leads to account non-existent free space in coverage calculations in the CCPP algorithms.
And (3) the CCPP application shares the same difficulty in reaching a $tc$ of 100\% as chaotic trajectories cannot be fully controlled. 

\begin{table}[htbp]
 \centering
  \caption{Performance of BCCP and CCPP in various maps.}
  \begin{tabular}{m{1.5cm} m{2.0cm} m{0.6cm} m{0.6cm} m{0.6cm} m{0.6cm} m{2.0cm} m{1.0cm} m{1.0cm} m{1.0cm}}
    \hline
    Map &  Approx. free area ($m^2$) & $SR$ (m) & LF (m) & $n_{iter}$ & $ns$ & Number of zones & $CT_{BCPP}$ (min) & $CT_{CCPP}$ (min) & $ratio_{CT}$\\
    \hline
    Esquare & 1010.44 & 3.5 & 7.0 & 20 & 20 & 40 & 13.22 & 39.28 & 2.97  \\
    \hline
     Ldense & 1217.47 & 2.0 & 4.0 & 20 & 20 & 60 & 43.71 & 101.08 & 2.31  \\
    \hline
    Hdense & 895.18 & 1.5 & 3.0 & 20 & 20 & 90 & 47.52 & 92.37 & 1.94\\
    \hline
  \end{tabular}
  \label{tab:BCCP and CCPP}
\end{table}

Table \ref{tab:BCCP and CCPP} compares the coverage time ($CT$) for BCCP and CCPP in different maps. At the parameters chosen, our CCPP is approximately 2.97, 2.31, and 1.94 times slower than the BCPP application for Esquare, Ldense, and Hdense environments, respectively. The results indicate reasonably comparable performance of CCPP and BCPP, specially in moderately and highly cluttered environments. Nevertheless, it is important to consider the following factors when comparing the two methods.
In general the BCPP can exert more control over the trajectories for two reasons. (1) There is a prior knowledge of the dimensions and shapes of the boundaries and obstacles of the environments which enables effective path tracing within cells.
(2) The application uses a start and a goal position to generate an overall coverage path plan which minimizes repeated coverage as the robot journeys from cell to cell, therefore optimizing the coverage time ($CT$). 
These attributes of BCPP result in the creation of a more absolute predetermined coverage path plan with very little wasted robot motion. 
Additionally, the BCPP application has lateral footprint and lateral overlap parameters that enable optimization of progressive coverage. This along with the ability to plan the entire coverage path provides a significant control of amount of coverage overlap as the robot moves around. Our CCPP application does not have have such parameters to fully control the coverage overlap. Other than the techniques we used to disperse the chaotic trajectories and avoid the obstacles, this work does not employ any other forms of trajectory control. It must also be accounted that the BCPP application used in this paper provides a more theoretical coverage time. The $CT$ might be more or less than what was calculated if a robot (in Gazebo or real-life) traveled the generated paths. Lastly, the CCPP corresponds to longer coverage times as the method must balance between two goals (as opposed to just one in BCPP): maintaining unpredictable movement and providing desired coverage. It is also possible that a different set of parameters not tried here would have brought the coverage times closer to those of the BCPP application. 

\begin{figure} [htbp]
\centering
   \subfloat[\centering ]{{\includegraphics[height=4cm] {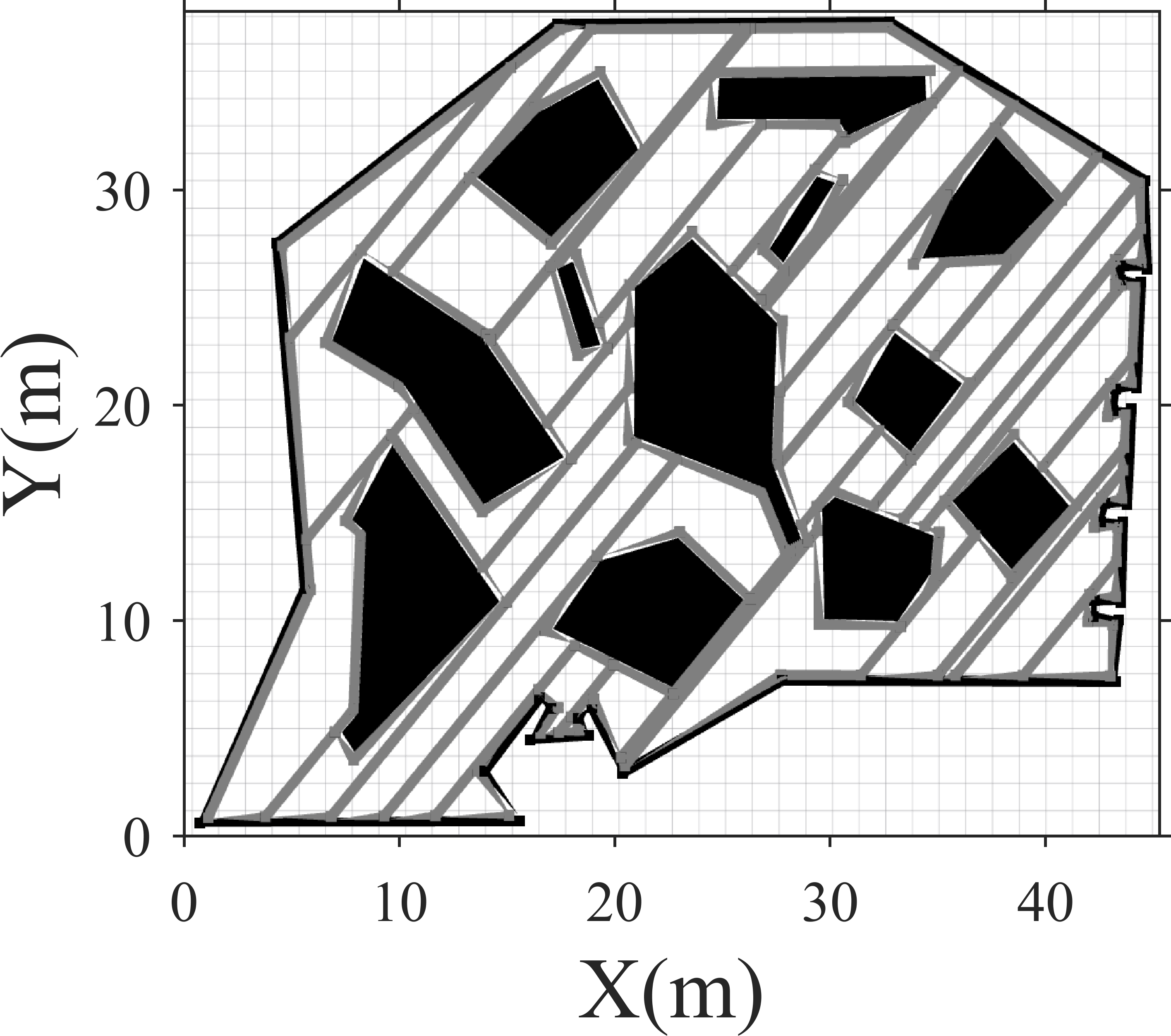} }}%
    \hspace{0.05\textwidth}
   \subfloat[\centering ]{{\includegraphics[height=4cm,width=4.6cm] {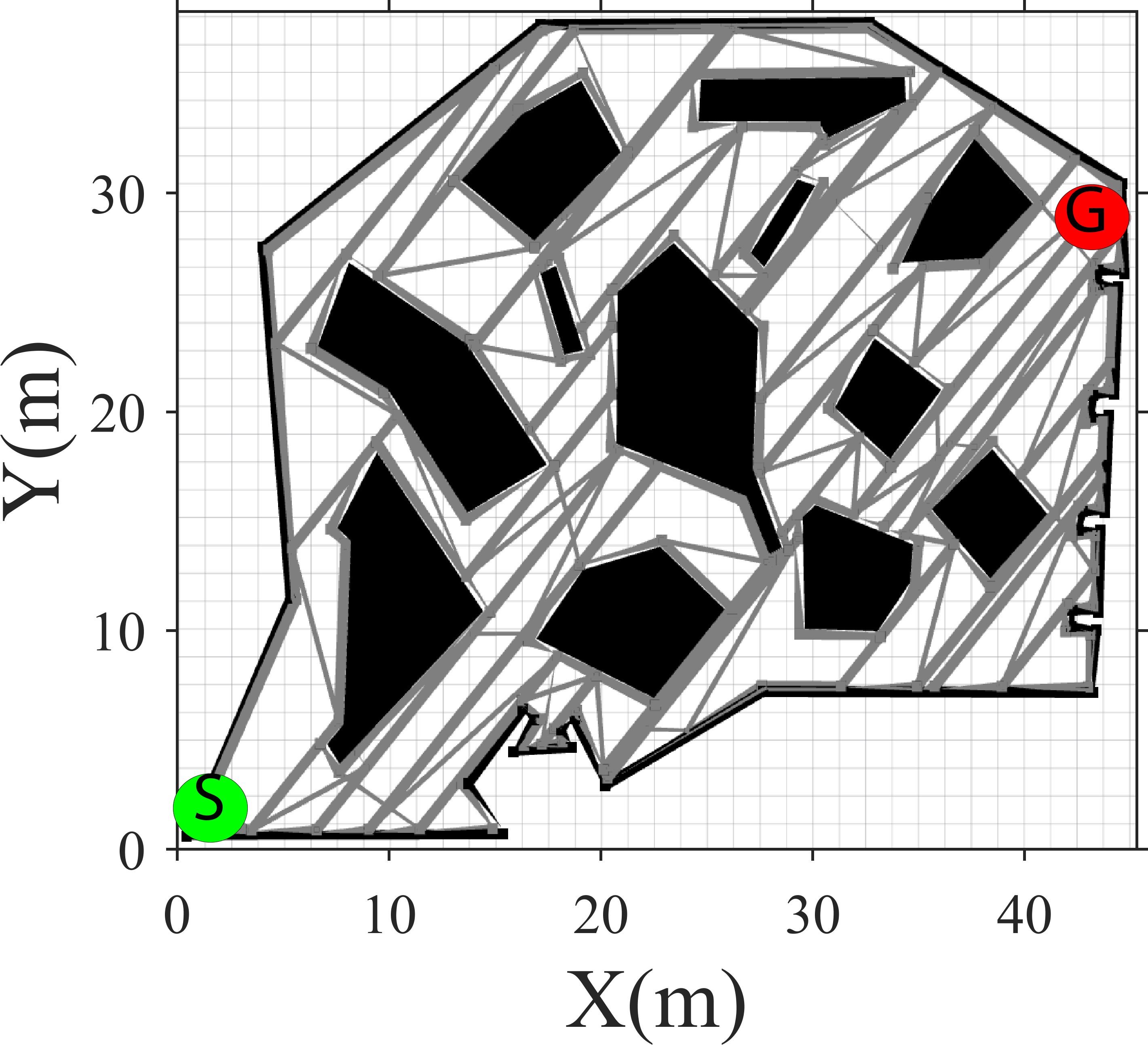} }}%
    \\
   \subfloat[\centering ]{{\includegraphics[height=4cm] {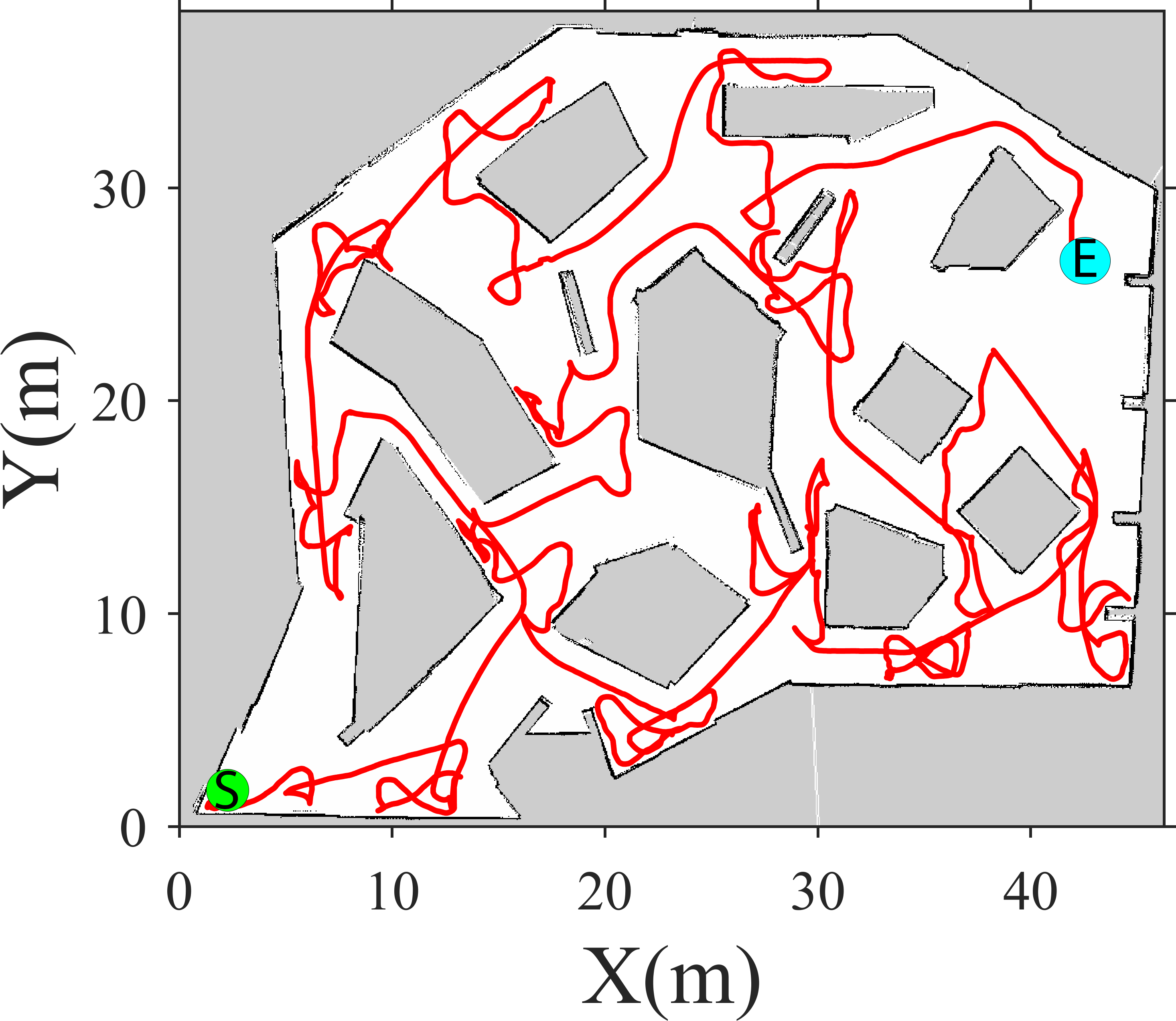} }}%
    \hspace{0.03\textwidth}
   \subfloat[\centering ]{{\includegraphics[height=4cm] {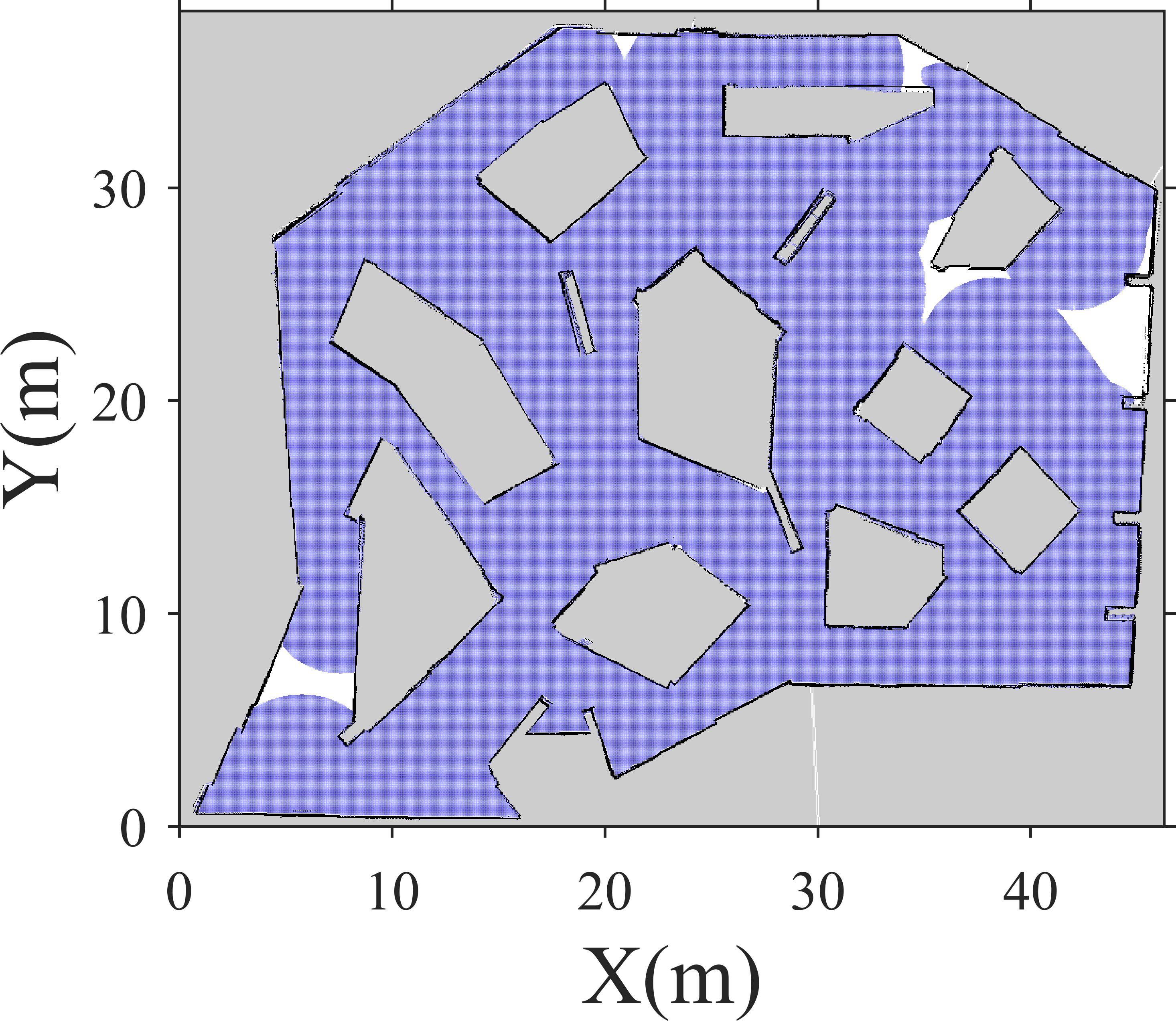} }}%
\caption{The BCPP and CCPP application used in Hdense: (a)-(b) shows cellular decomposition and planned boustrophedon coverage paths. (c)-(d) shows chaotic trajectories and covered cells. S: Start position; G: Goal position; E: End position.}
    \label{fig:3BCPP/CCPP-3.5}
\end{figure}

We ran as well a comparison using the  maximum $SR$ of the simulated TurtleBot3 Lidar in Hdense. Figs.\ \ref{fig:3BCPP/CCPP-3.5} (a) and (b) present the coverage results of the BCPP. In this experiment, the BCPP application covered the Hdense environment in 33.56 minutes with a lateral footprint of 7 m, indicating a 29.40\% drop in $CT$ compared to 3m lateral footprint (see Table \ref{tab:BCCP and CCPP} and Figs.\ \ref{fig:3BCPP}(c) and (f)). Figs.\ \ref{fig:3BCPP/CCPP-3.5} (c) and (d) showcase the performance of our CCPP in the Hdense environments at 97\% coverage. The $SR$ was 3.5 m and the $CT$ was 44.24 minutes at parameter values set in Table ~\ref{tab:BCCP and CCPP}. There is a 56.23\% decrease in $CT$ at this $SR$ from the CT using $SR$ of 1.5m (Figs.\ \ref{fig:3CCPP}(b) and (e)). Section \ref{sec:Potential advantages of CCPP over BCPP} will further discuss the potential advantages of our CCPP method and problems that arise when BCPP faces worst case scenario computational complexity issues as well as some additional problems thereof. 

%\newpage
\subsection{Potential Advantages of CCPP Over BCPP}
\label{sec:Potential advantages of CCPP over BCPP}
When it comes to sensor based coverage applications, our CCPP method might be more adaptable to various types of obstacles. The BCPP method requires the setting of obstacles as polygons. In environments featuring complex-shaped obstacles, the BCPP application encounters challenges in generating cells and, consequently, paths. One approach to mitigate this problem is to represent complex obstacles with simpler polygon shapes; however, this may lead to less comprehensive coverage. The use of simpler polygon shapes can result in cells that may not precisely capture the entire free space in the real physical environment. Considering this factor alongside the adoption of a rigid path planning strategy might lead to a notable decline in the performance of the coverage task. Given the BCPP application used for comparison, it is impossible to provide definite proof of this as $tc$ is not provided in output. On this same note, it is important to highlight that the performance of our CCPP method improves as the environment becomes more cluttered with obstacles, given the results in Table \ref{tab:BCCP and CCPP}. The $ratio_{CT}$ reduces from 2.97 to 1.94 as we change the environment from simple Esquare to more complex Hdense. This indicates the potential adaptability of our CCPP approach. \color{black}

This issue may get compoundingly worse in sensor-based surveillance task scenarios depending on the $FOV$ of the sensor used in the real-world or Gazebo simulation. In order to mitigate any theoretical performance drop off, a user of this application may have to set the lateral footprint parameter at values < $2 \times SR$.  However, reducing the lateral footprint prolongs the coverage time. Our CCPP application does not have the aforementioned drawbacks. It does not depend on cell decomposition for the path planning process and therefore bypasses any hurdles involving computational complexity in this area. As long as an accurate occupancy-grid map exists, it can handle a wide variety of obstacle types and environments and provide absolute certainty that the desired coverage of the physical environment was met upon completion. The real-life implementation discussed in Section \ref{sec:Coverage of a classroom with a TurtleBot2 with CCPP and BCPP applications} casts light on the BCPP challenges.

\subsubsection{Coverage of a classroom with a TurtleBot2 with CCPP and BCPP applications}
\label{sec:Coverage of a classroom with a TurtleBot2 with CCPP and BCPP applications}
\begin{figure} [h]
\centering   
    \includegraphics[height=4cm,width=7cm]{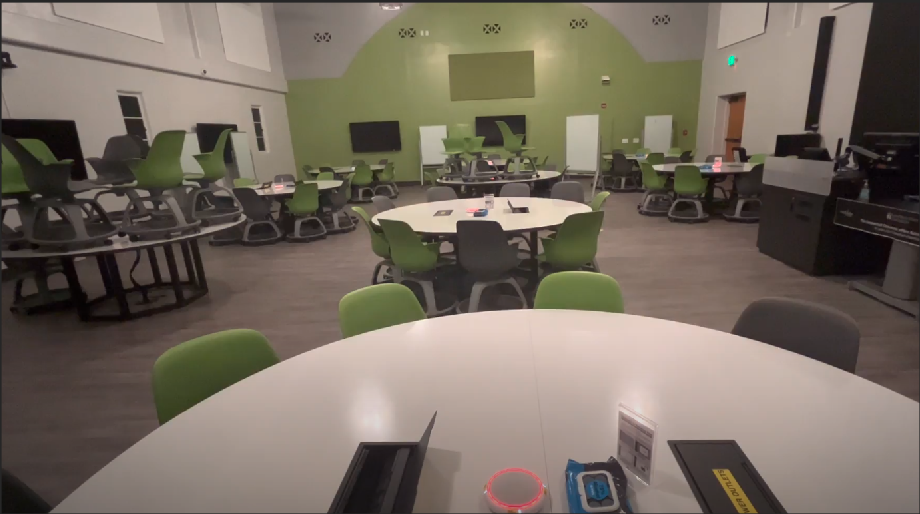}%
 \caption{The classroom used for live testing.}
    \label{fig:classroom}
\end{figure}

This section showcases the performance of our CCPP against BCPP in a real environment. The environment for this experiment was a classroom on campus (see Fig. \ref{fig:classroom}). Some chairs were placed on tables to provide environmental variety and improve ease of map creation via SLAM. As in section ~\ref{sec:Comparing coverage times}, the parameters were set as follows: $n_{iter}=20$, $ns=20$, the number of zones$=20$, $v=0.45$ m/s, $dc=90\%$, and $SR=2.5$ m. The camera sensor has a $FOV$ ranging from -1.57 to 1.57 radians. The coverage time at these parameters was 8.87 minutes. These parameters were arbitrarily chosen, and it is uncertain whether this coverage time is the shortest possible for this $dc$ and $SR$. 

\begin{figure} [htpb]
\centering
   \subfloat[\centering ]{{\includegraphics[height=4cm] {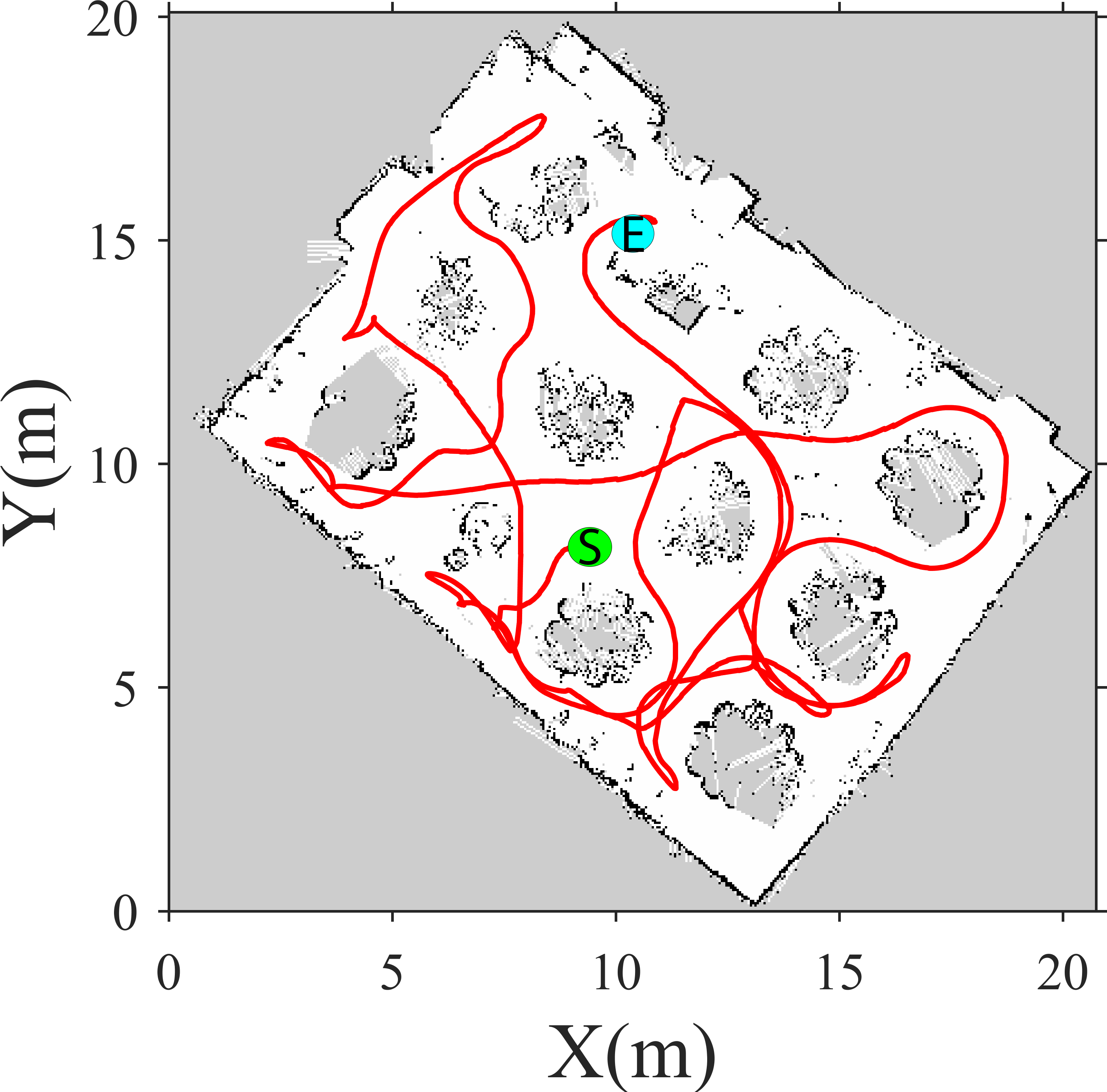} }}%  
    \hspace{0.05\textwidth}
   \subfloat[\centering ]{{\includegraphics[height=4cm] {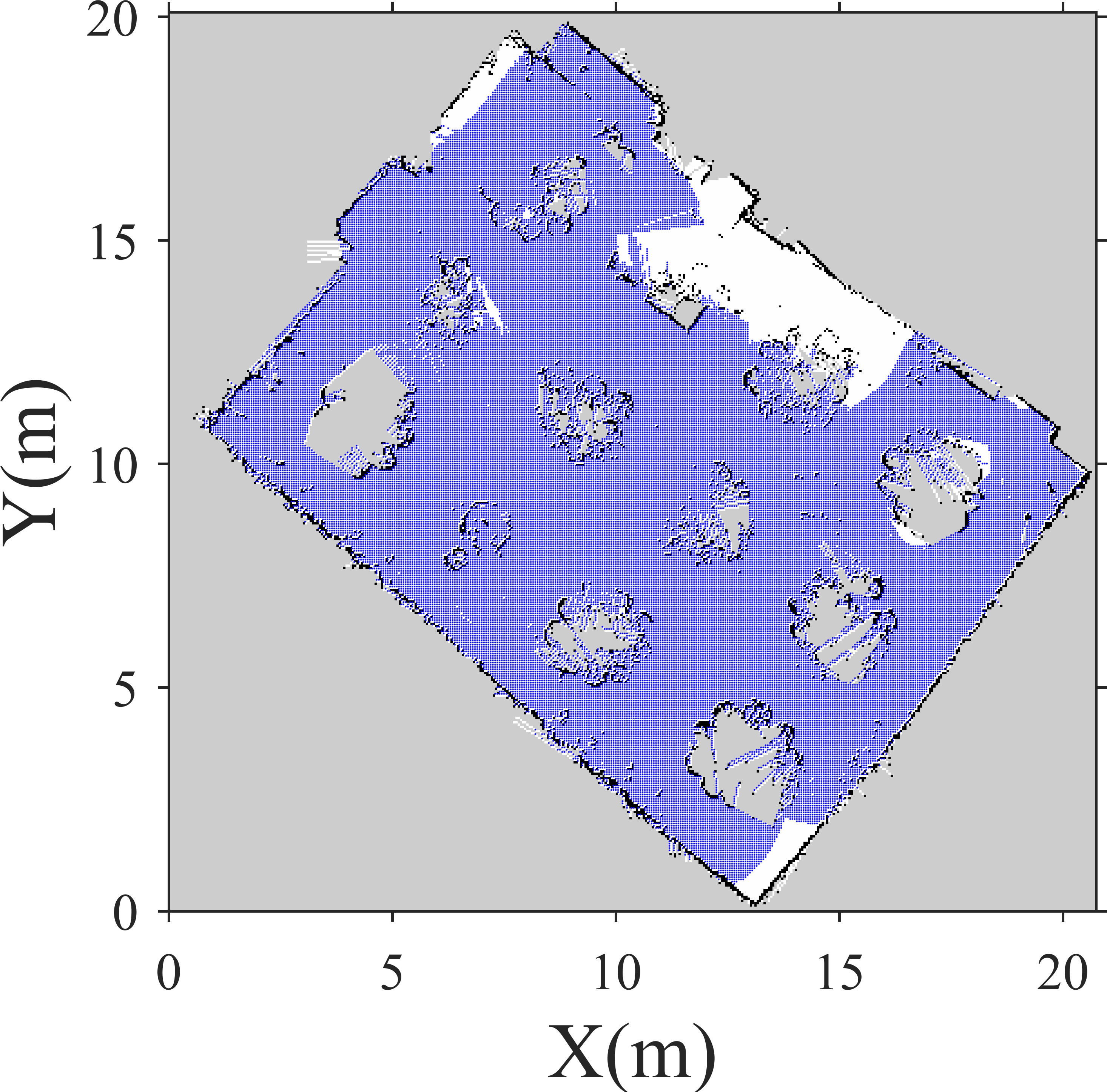} }}%
\caption{CCPP application ($SR$ = 2.5 m)  used in a classroom for 90\% coverage: (a) chaotic trajectories, and (b) covered cells. S: Start position; E: End position.}
    \label{fig:CScoverage}
\end{figure}

Fig. \ref{fig:CScoverage} represents the chaotic trajectories and coverage. The $dc$ is set at 90\% to compensate for the inaccuracies of the generated map. The SLAM process generated non-existent free space in certain areas of the map as a result of sensor problems. This $dc$ was a compromise to ensure the coverage task could be completed. The parameters of the BCPP application were set as follows: (1) the lateral footprint was set to 5.0 m, (2) the maximum acceleration was set to 0.1 m/$s^2$ (the maximum acceleration of the TurtleBot2), and (3) the wall distance was set to 0.36 m to accommodate the size of the TurtleBot2. The CCPP application provides the robot the ability to maneuver around the sets of tables and chairs, as well as the various whiteboards spread in the map.

\begin{figure} [htpb]
\centering
    \subfloat[\centering ]{{\includegraphics[height=4cm] {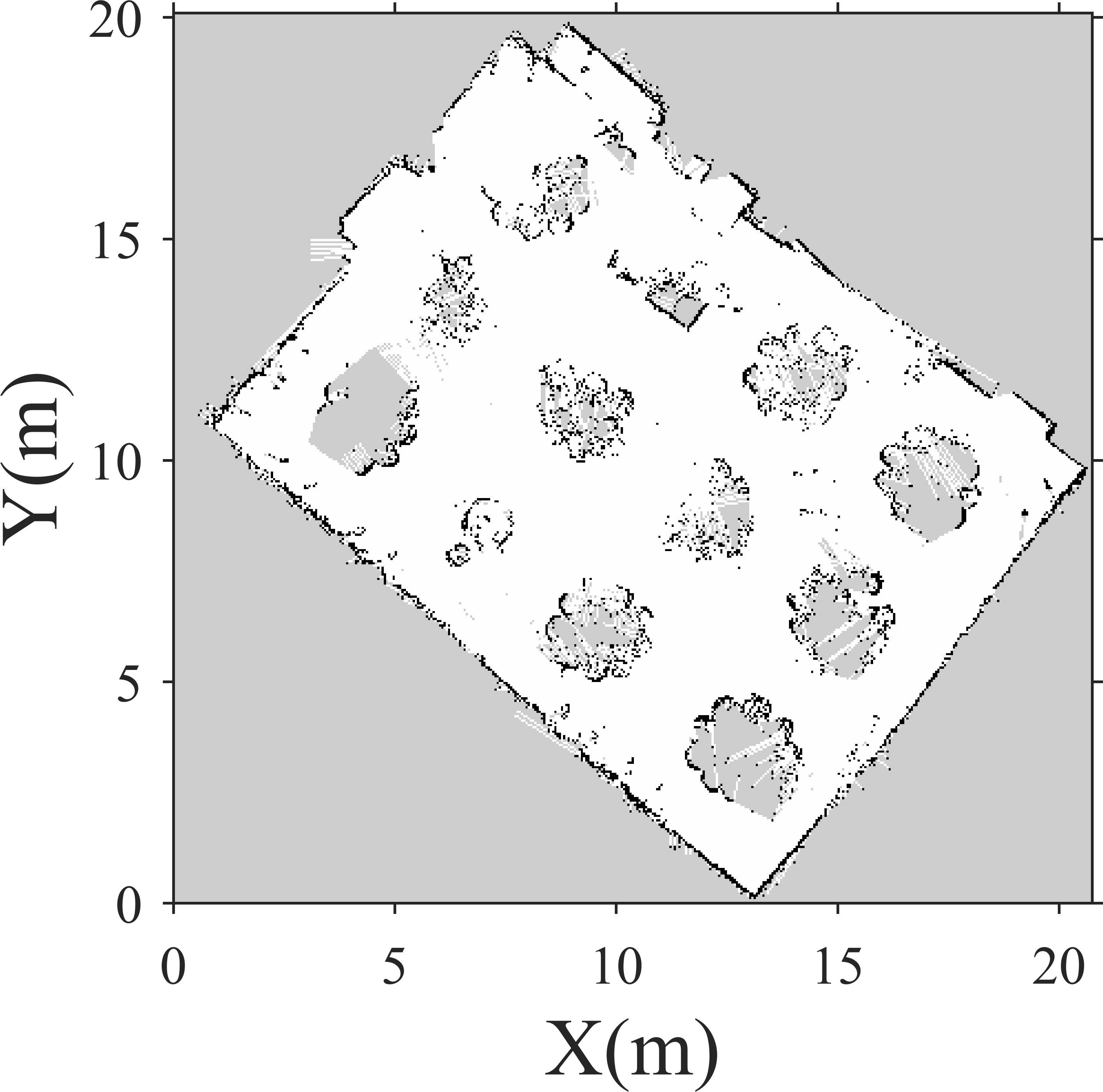} }}%
    \hspace{0.01\textwidth}
   \subfloat[\centering ]{{\includegraphics[height=4cm] {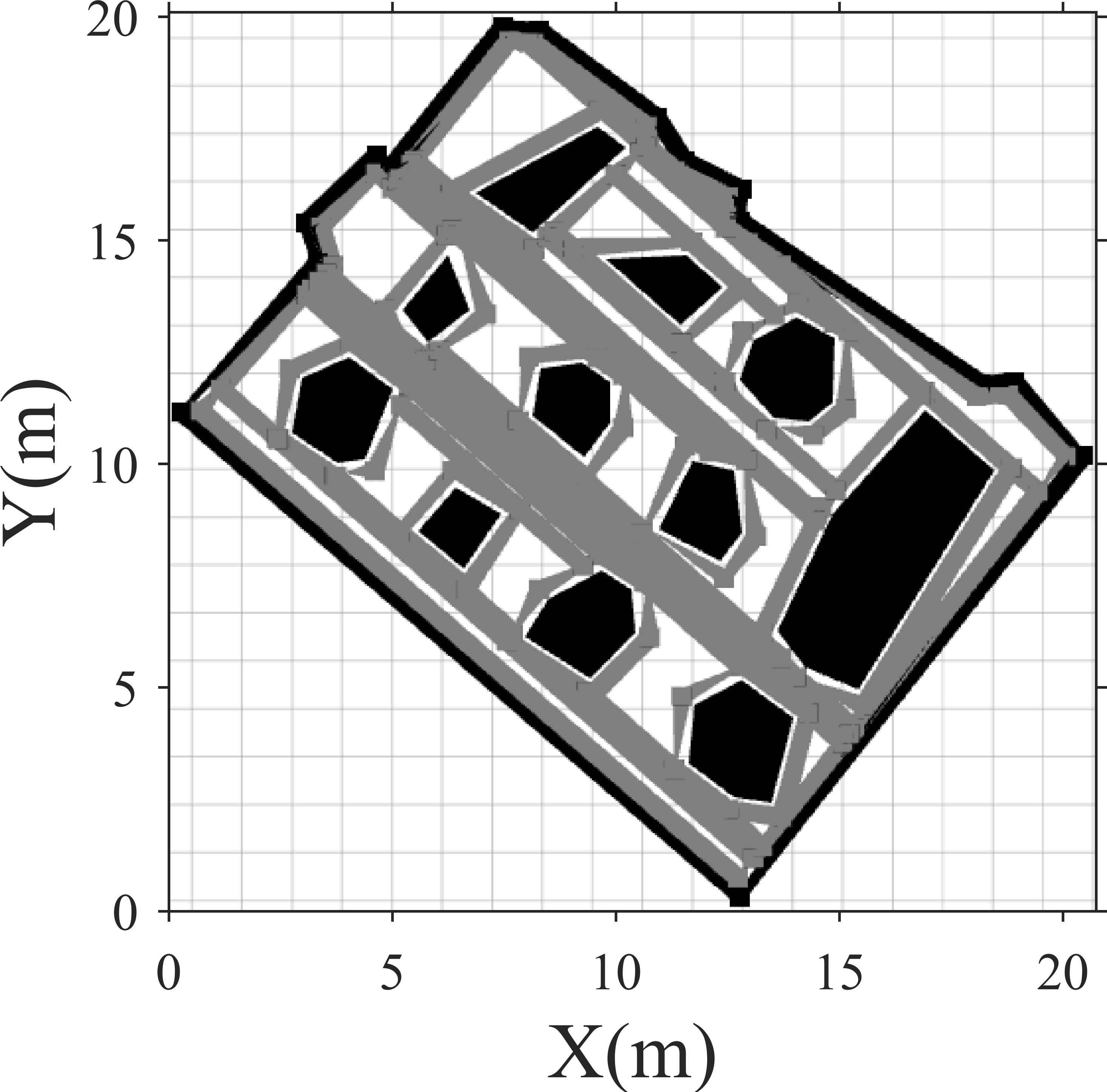} }}%
    \hspace{0.01\textwidth}
   \subfloat[\centering ]{{\includegraphics[height=4cm] {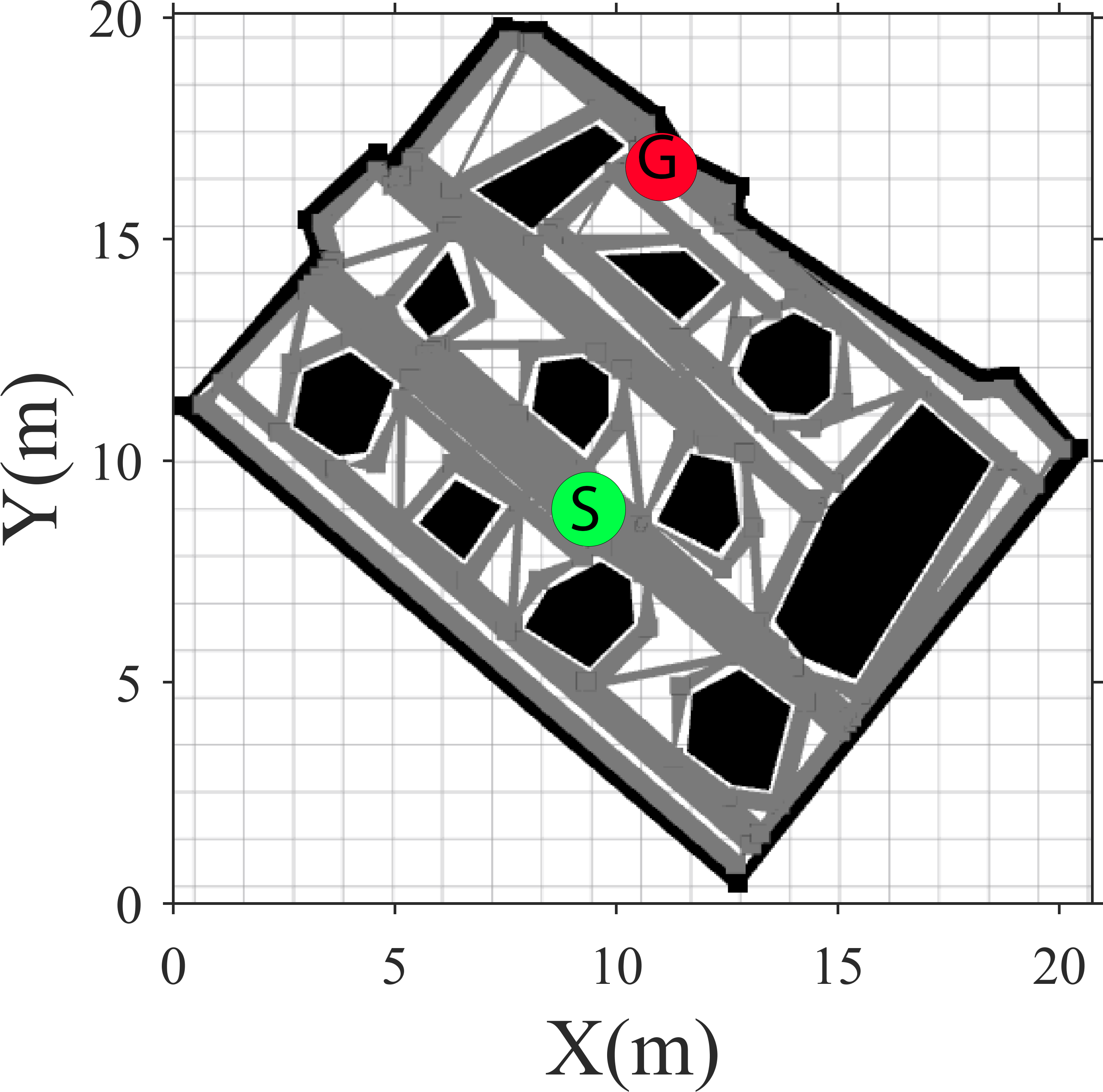} }}%
\caption{BCPP application ($LF$ = 5.0 m)  used to cover classroom: (a) the map (generated via SLAM), (b) cellular decomposition, and (c) planned boustrophedon coverage path. S: Start position; G: Goal position.}
    \label{fig:BCcoverage}
\end{figure}

Fig. \ref{fig:BCcoverage} showcases the cellular decomposition and coverage plan of the classroom with the BCPP application ($CT$ at 10.42 minutes). As seen in Fig.\ \ref{fig:BCcoverage}, some of the obstacles (two sets of tables and chairs, and a white board) did not only have to be simplified, but combined into one polygon shape in order to ensure the cellular decomposition was computationally possible for the BCPP application to handle. Parts of the environment boundaries had to be simplified to this effect as well. A better CPU might have handled the necessary computations. A side by side view of the map in Fig.\ ~\ref{fig:BCcoverage}(a) (generated via SLAM) and the polygon representation of the environment in Figs.\ ~\ref{fig:BCcoverage}(b)-(c) show the critical differences in environment representation. The combination of sensor characteristics and inaccurate polygon representation of obstacles could cause significant coverage performance drop off within the context of thoroughness of coverage. It is possible that our CCPP method has the same advantages across all methods which use cellular decomposition, such as the Morse-based cellular decomposition method \cite{galceran2013survey,galceran2012efficient}.

%\newpage
\subsection{Test of Computational Speed of Coverage Calculation Technique}
\label{sec:Test of computational speed of coverage calculation technique}
This experiment was developed to test the computational performance of our coverage calculation method. The setup for this experiment is as follows. (1) The maximum velocity at which the simulated TurtleBot3 can reach is set to 5m/s. (2) The robot's starting position is at one end of a runway (in simulation). The robot is driven in a straight line to the opposite end of the runway, which ends at a wall. (3) Upon the robot's maximum velocity reaching 5m/s, the algorithms are signaled to start coverage calculation until the robot stops at the wall. (4) The distance traveled between each call of Algorithm 5 is recorded throughout the journey. Table ~\ref{tab:2} contains the maximum, minimum, and average recorded  distance of each experiment. 

\begin{table}[h]
  \centering
  \caption{Performance of computational speed of coverage calculation.}
  \begin{tabular}{m{1.5cm} m{1.0cm} m{1.2cm} m{1.0cm} m{1.5cm} m{1.3cm} m{1.2cm}}
    \hline
    Runway & No. of Threads & $v$ (m/s) & $SR$ (m) & Average distance (m) & Max distance (m) & Min distance (m)\\
    \hline
    Runway 1 & 1 & 5.0 & 5.0 & 2.29 & 3.16 & 1.84\\
    \hline
    Runway 2 & 1 & 5.0 & 7.5 & 17.76 & 20.45 & 15.45\\
    \hline
    Runway 2 & 20 & 5.0 & 7.5 & 13.18 & 20.99 & 1.83\\
    \hline
    Runway 2 & 1 & 3.0 & 7.5 & 6.08 & 12.00 & 2.15\\
    \hline
  \end{tabular}
  \label{tab:2}
\end{table}

These recorded distances provide the basis for judging the computational efficiency of coverage calculation. If the average recorded distance is less than the $SR$, this means that coverage updates generally occur at a faster rate than the robot is able to completely move out of the area covered during the previous iteration of Algorithm 4. For this experiment, the sensor $FOV$ is set from 0 to 6.28 radians, and sensor $FOV$ mapping is configured so that the coverage status of every queried cell is updated. In this way, accurate coverage is neglected for the sake of maximizing the computational burden placed on our coverage calculation method. For this experiment, Algorithm 4 includes a multi-threading option that divides the cells $\in$ Query into subsets, which are then processed with separate threads of Algorithm 5 for possibly faster computation. Multi-threading is tested to check for any improvements in computational performance. It must be noted that: (1) the value of 1 in the "No. of threads" column of Table ~\ref{tab:2} represents no multi-threading, and (2) the maximum velocity value of 5m/s was chosen as this was the highest robot velocity that the simulation would allow without creating computational errors within the physics engine of Gazebo.

%\newpage
\begin{figure} [htpb]
\centering
  \subfloat[\centering ]{{\includegraphics[height=1.0cm,width=7.50cm] {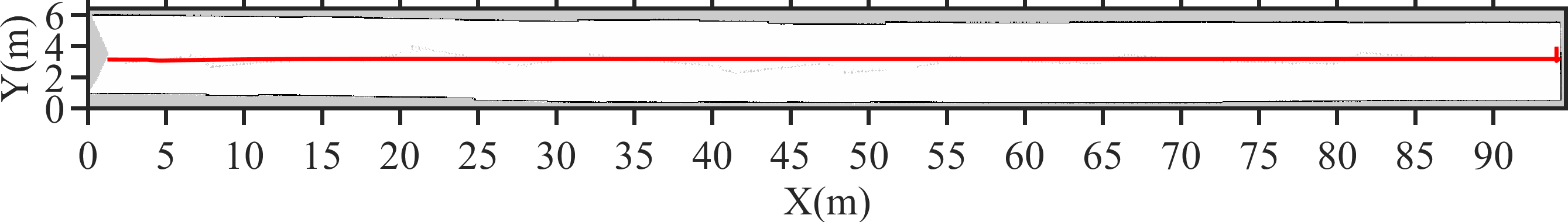} }}%  
   \hspace{0.3cm}
   \subfloat[\centering ]{{\includegraphics[height=1.0cm,width=7.50cm]   {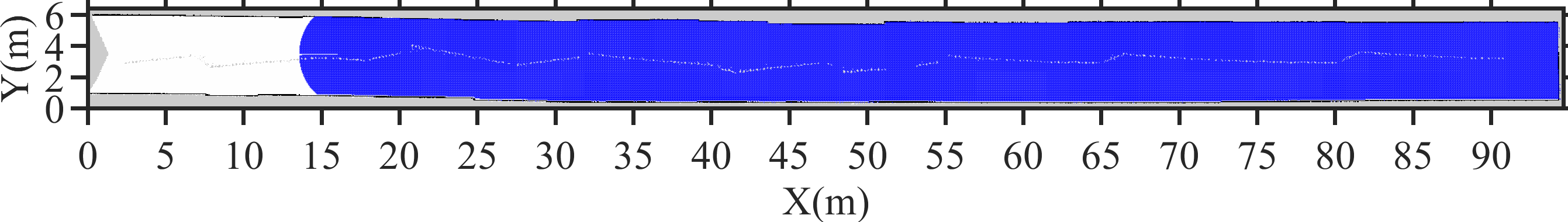} }}%
       \\
     \subfloat[\centering ]{{\includegraphics[height=1.0cm,width=7.50cm]  {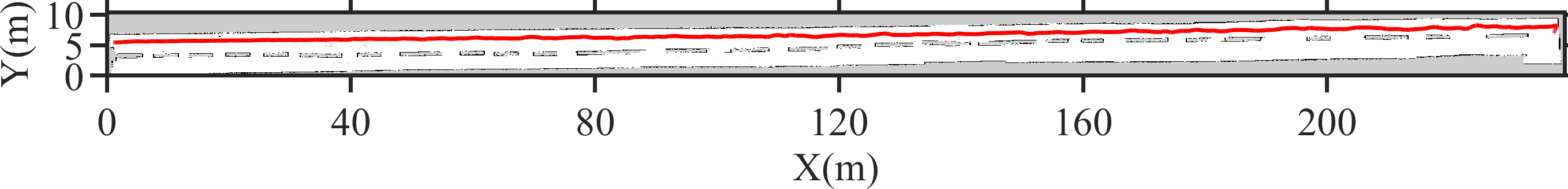} }}%  
    \hspace{0.3cm}
   \subfloat[\centering ]{{\includegraphics[height=1.0cm,width=7.50cm]  {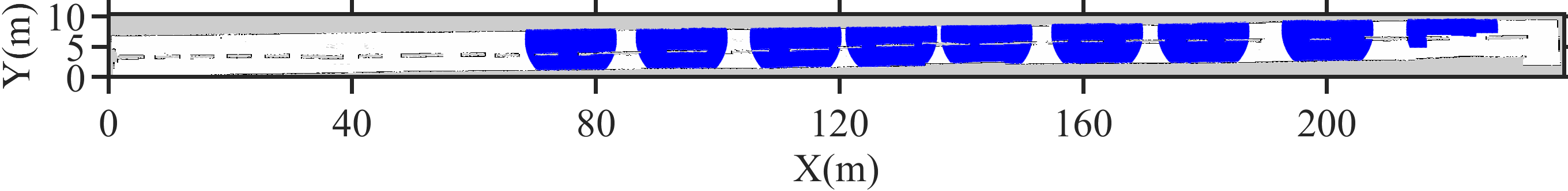} }}%
      \\
     \subfloat[\centering ]{{\includegraphics[height=1.0cm,width=7.50cm] {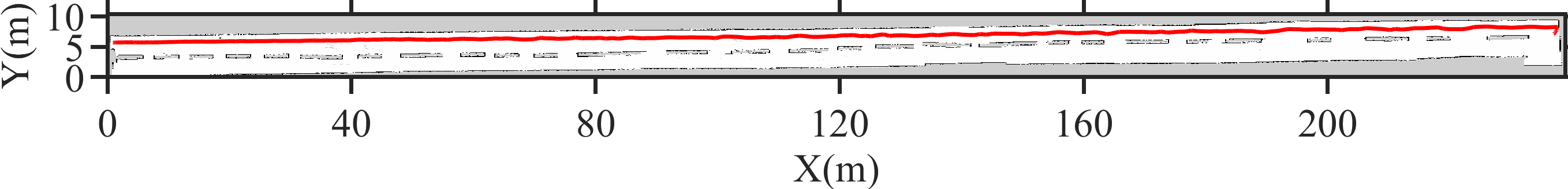} }}% 
    \hspace{0.3cm}
   \subfloat[\centering ]{{\includegraphics[height=1.0cm,width=7.50cm] {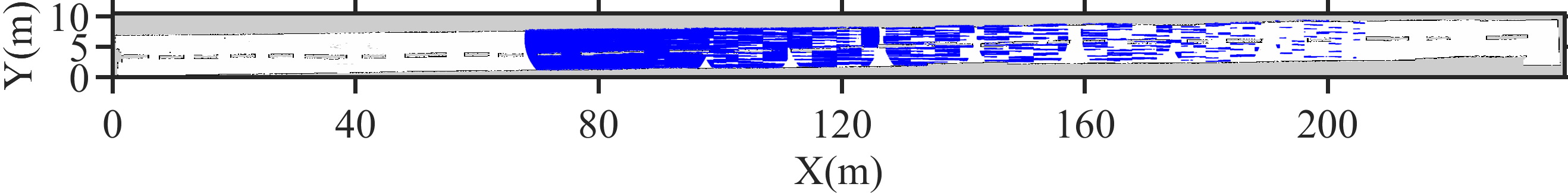} }}%
   \\
    \subfloat[\centering ]{{\includegraphics[height=1.0cm,width=7.50cm]  {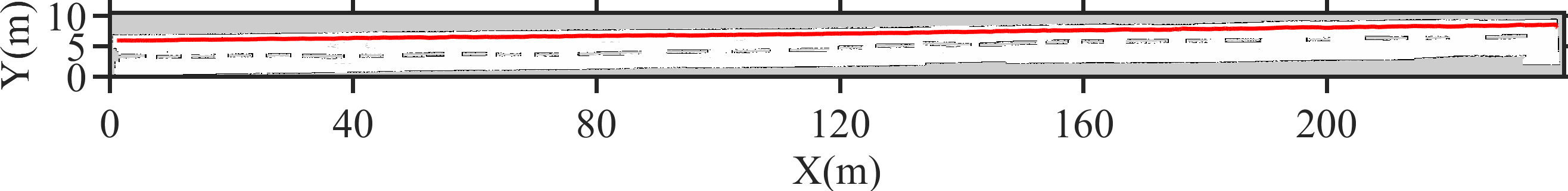} }}%  
    \hspace{0.3cm}
   \subfloat[\centering ]{{\includegraphics[height=1.0cm,width=7.50cm] {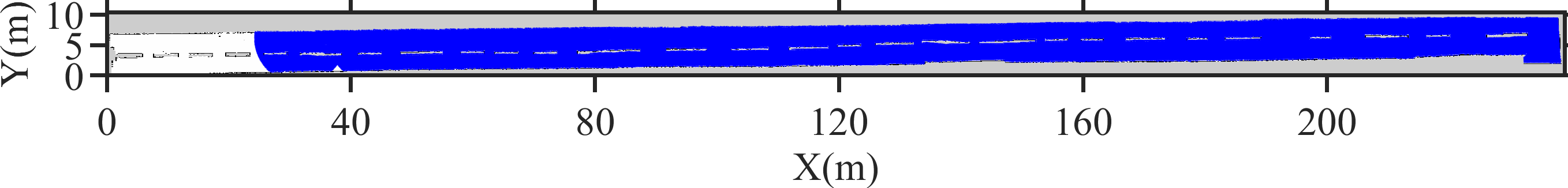} }}%

\caption{The visual representation of the performance of the coverage calculation method tested on 2 simulated runways. The figures on the left and right depict the robot's trajectory and coverage across the runway, respectively. (a, b) Runway 1, No. of Threads$=1$, $v=5$ m/s, $SR=5$ m; (c, d) Runway 2, No. of Threads$=1$, $v=5$ m/s, $SR=7.5$ m   ; (e, f) Runway 2, No. of Threads$=20$, $v=5$ m/s, $SR=7.5$ m ; (g, h) Runway 2, No. of Threads$=1$, $v=3$ m/s, $SR=7.5$ m.}
    \label{fig:runways}
\end{figure}

Figs.\ \ref{fig:runways}(a), (c), (e), and (g) depict the straight line trajectory of robot from one end of a runway to the other. Fig.\ \ref{fig:runways}(b), (d), (f), and (h) depict the coverage started from the point at which simulated TurtleBot3 reaches the maximum velocity on the aforementioned trajectory. The smallest runway (Figs. \ref{fig:runways}(a) and (b)) is covered using an $SR$ of 5.0 m. The average, maximum, and minimum recorded distances are 2.29 m, 3.16 m, and 1.84 m, respectively. This means that at no point is there a gap in coverage update. That is to say, from the first to the last call of Algorithm 5, the entire area within that time frame is covered (as seen in Fig. \ref{fig:runways}(b)). The biggest runway (Figs. \ref{fig:runways}(c) to (h)) is covered using an $SR$ of 7.5 m. As an aside, the obstacles placed in the middle of the runway provided unique features which enabled mapping the runway environment. With multi-threading disabled, the average, maximum, and minimum recorded distances are 17.76 m, 20.45 m, 15.45 m, respectively. These values indicate that there will be gaps in the coverage, and Fig. \ref{fig:runways}(d) visually shows this result. With multi-threading (20 threads), the average, maximum, and minimum recorded distances are 13.18 m, 20.99 m, 1.83 m, respectively. The average distance is incongruent with what is shown in Fig. \ref{fig:runways}(f). We speculate that the Python Global Interpreter Lock (GIL) and the Operating System Thread Scheduler delay the execution of threads across the iterations of Algorithm 4. Therefore, threads from previous iterations could still be in queue to start or finish their execution. This results in the peculiar breaks in coverage shown in Fig. \ref{fig:runways}(f). Developing the source code in programming languages (such as C++) that are not limited by the GIL could possibly solve this issue and lead to effective use of the multi-threading option for faster computations of the coverage.

As a last trial run, we reduced the speed of the maximum velocity of the TurtleBot3 to 3 m/s and ran the same experiment with multi-threading disabled at this velocity. The average, maximum, and minimum distances are 6.08 m, 12.00 m, and 2.15 m, respectively. The results (shown in Figs.\ \ref{fig:runways} (g) and (h)) indicate that all areas between the start and end of coverage calculations are completely covered. The computation speed of coverage calculation enables effective coverage rate update even at robot speeds of up to 3 m/s with $SR$ at 7.5 m.

\subsection{Parameters Influence on Coverage Time}
\label{Parameters influence and Future Directions}
\begin{figure} [htbp]
\centering
     \subfloat[\centering]{{\includegraphics[height=4cm]{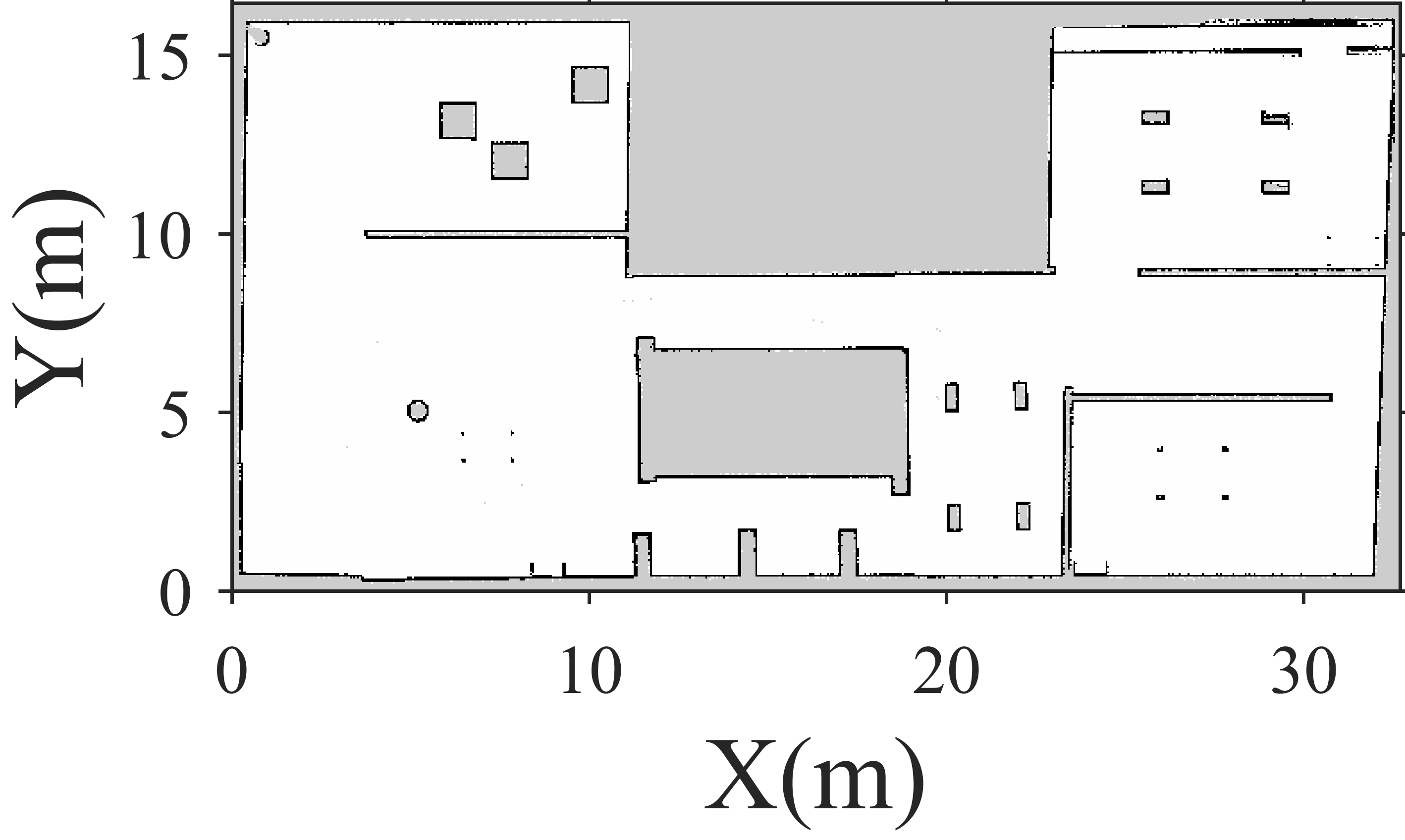} }}%
     \hspace{0.01\textwidth}
    \subfloat[\centering ]{{\includegraphics[height=5cm]{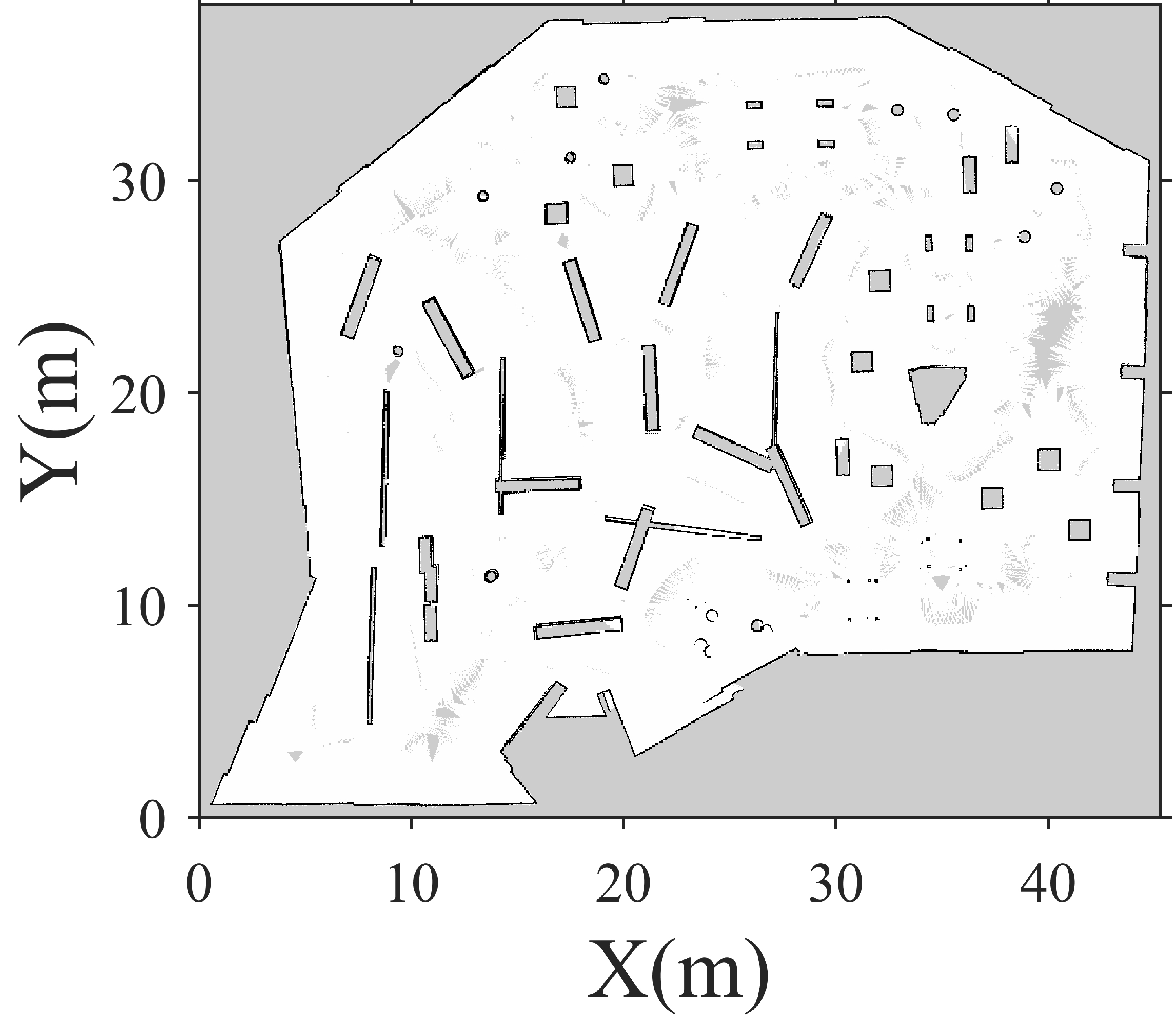} }}%
    \caption{Maps of simulated environments: (a) Bungalow, and (b) LDW1.}
    \label{fig:BandL}
\end{figure}

This section discusses the effect of parameters on the coverage time. Two environments have been covered multiple times using different sets of parameters, and the resulting coverage times have been recorded in Table ~\ref{tab:3}. Fig. \ref{fig:BandL} depicts the maps of these environments. The names of the environments shown in Figs. \ref{fig:BandL}(a)-(b) are: (1) Bungalow, and (2) LDW1 in that order. The $dc$ for every experiment was set to 90\%. The robot's starting position was kept the same for every test carried out in an environment, and the velocity ($v$) for every experiment was 0.22m/s. The simulated TurtleBot3 and 2D LiDAR were utilized in all experiments. The $FOV$ for all experiments is ranging from 0 to 6.28 radians. The maximum sensing area for all experiments is 38.48 $m^2$. From the results in Table \ref{tab:3}, we try to glean some general patterns in coverage performance within the context of $CT$. 
\begin{table}[htbp]
  \centering
  \caption{Performance of CCPP on two environments under various parameters.}
  \begin{tabular}{m{1.5cm} m{1.5cm} m{1.0cm} m{0.7cm} m{0.7cm} m{1.0cm} m{1.5cm} m{1.5cm} m{1.0cm} m{1.0cm}} % Adjust the column widths as needed

  \hline
  Map & Approx. free area ($m^2$) & Case \# & $n_{iter}$ & $ns$ & No. of zones & Maximum Sensing Area ($m^2$) & Average Zone Size ($m^2$) & $R_{SZ}$ & $CT$ (min) \\
  \hline

  \multirow{5}{*}{Bungalow} & \multirow{5}{*}{365.03} & 1 & 1000 & 20 & \multirow{2}{*}{15} & \multirow{5}{*}{38.48} & \multirow{2}{*}{24.34} & \multirow{2}{*}{1.58} & 38.12 \\
  & & 2 & 10 & 10 & & & & & 9.01\\
  & & 3 & 1000 & 20 & \multirow{2}{*}{50} &  & \multirow{2}{*}{7.30} & \multirow{2}{*}{5.27} & 65.72\\
  & & 4 & 10 & 10 & & & & & 13.50\\
  & & 5 & 10 & 10  & 25 &  & 14.60 & 2.64 & 12.06 \\

  \hline
  \multirow{4}{*}{LDW1} & \multirow{4}{*}{1146.34} & 6 & 20 & \multirow{4}{*}{20} & 25 & \multirow{4}{*}{38.48} & 45.85 & 0.84 & 30.43 \\
  & & 7 & 2000 & & 15 & & 76.42 & 0.50 & 92.14 \\
  & & 8 & 20 & & \multirow{2}{*}{50} & & \multirow{2}{*}{22.93} & \multirow{2}{*}{1.68} & 31.01 \\
  & & 9 & 2000 & & & & & & 179.68 \\
  
  \hline
  \multirow{4}{*}{Esquare} & \multirow{4}{*}{1010.44} & 10 & 20 & \multirow{4}{*}{20} & 21 & \multirow{4}{*}{38.48} & \multirow{2}{*}{48.12} & \multirow{2}{*}{0.80} & 24.88 \\
  & & 11 & 1000 & & 21 & & & & 57.57 \\
  & & 12 & 20 & & 45 & & \multirow{2}{*}{22.45} & \multirow{2}{*}{1.71} & 24.47 \\
  & & 13 & 1000 & & 45 & & & & 61.44 \\

  \hline
  \end{tabular}
  \label{tab:3}
\end{table}

Due to the fact that parameters must be set manually, we discovered the following. (1) Setting the number of zones such that the ratio of the maximum sensing area to the average zone size (\textbf{$R_{SZ}$}) is greater than or equal to 1 leads more favourable results, as $n_{iter}$ does not need to be set high to reach $dc$. In this way there is less burden in making some kind of educated guess as to how high to set $n_{iter}$ to even reach $dc$ much less perform the coverage task within a reasonable time frame. (2) In the same vain, setting the number of zones such that \textbf{$R_{SZ}$} is less than 1 is less favorable. As a wrong guess on the $n_{iter}$ value could at best, result in a $CT$ that deviates drastically from the optimal $CT$. Results not shown here suggest that, at worst, the wrong $n_{iter}$ results in a scenario where $dc$ is never reached. However, (3) it is not necessary for \textbf{$R_{SZ}$} to be greater or equal to 1. For cases 6 and 8 which have a \textbf{$R_{SZ}$} of 0.84 and 1.68, respectively, the $CT$ was approximately the same (about 30 minutes at $n_{iter}=$20). The $CT$ for these cases was as well the lowest for LDW1. It could be the case that a \textbf{$R_{SZ}$}$\approx$1 is good enough. (4) From cases 2, 4, and 5, we found that at low $n_{iter}$, a further increase in ratio beyond an approximate value of 2 may not result in better $CT$. For Bungalow, a \textbf{$R_{SZ}$} of 1.58, 5.27, and 2.64 result in a $CT$ of 9.01, 13.50, and 12.06 minutes respectively (at $n_{iter}=$10). (5) From cases 1 and 3, we found that at a high $n_{iter}$, a further increase in ratio beyond an approximate value of 2 may not result in better $CT$. For Bungalow, a \textbf{$R_{SZ}$} of 1.58 and 5.27 (at $n_{iter}=$1000) result in a $CT$ of 38.12 and 65.72 minutes, respectively. (6) From cases 7 and 9, we found that a \textbf{$R_{SZ}$} much less than 1 (0.5) with a high enough $n_{iter}$ could result in lower $CT$ than a run with a \textbf{$R_{SZ}$}$\geq$1.

Ultimately, it is uncertain as to whether the findings (3) to (6) are general patterns. There could be a great deal of interplay between the results and the environment properties such as size, shape, and particularly obstacle density. Table \ref{tab:3} additionally includes the results corresponding to a simple low-cluttered environment, Esquare, to further corroborate these findings. In Esquare environment, cases 10 and 12 support finding (3) and (4) (especially (3)) while Cases 11 and 13 support findings (5) and (6). However, it is important to reinstate that the properties of the environment (and the robot) could have a significant effect on the results.

%\newpage
\section{Conclusions}
\label{sec:Conclusions}
This study proposes new techniques to realize the practical application of CCPP in realistic environments. Three techniques were developed to achieve this goal. These contributions were integrated to develop a new CCPP application in the ROS framework that enables effective coverage of large, complex-shaped environments with varying obstacle densities. The developed coverage calculation technique was shown to be computationally efficient at speeds up to 5 m/s for a robot with sensing range of 5.0 m. 

A multi-threading option could possibly provide faster computation for coverage calculations if the source code is developed in a programming language like C++ which is not limited by a global interpreter lock. It must also be noted that a complete refactor of the algorithms might allow the use of multiprocessing for faster coverage calculations as well. The current clustering method in use is sub-optimal for our use case. 

For instance, some clusters might be split apart by obstacles (see Figure. \ref{fig:mapzoning}(b)); this can significantly reduce effectiveness of coverage. \color{black}For optimization, it is probable that a supervised clustering method must either be developed or fine-tuned for our use case. This method will have to be paired with some prior classification step  so as to jointly handle the following:
(1) Separate classification of any $Cell_F$ separated by obstacles (i.e. walls). So that clusters can be compacted into concentrated areas.
(2) Even distribution of cluster size whenever possible.

Due to use of occupancy-grid maps, our method is susceptible to exponential memory usage dependent on size of environment and resolution used for grid map representation. This could lead to memory errors.  Further work must be done to tackle this issue.

The performance of our CCPP application was tested against a more established CPP, the boustrophedon coverage path planner.  The BCPP application used for this comparison performed coverage at shorter coverage times, but the CCPP was shown to have the potential to match it in certain environments if provided with the right set of system parameters. At the same time, the CCPP provides the unpredictability in the motion necessary to avoid attacks in adversarial environments. Additionally, the performance of our CCPP method improved when used in more complex obstacle clustered environments. A result which indicate strong adaptability to cluttered environments\color{black}. These findings must however be verified with extensive testing using a BCPP application which provides a robot in realistic simulation environments such as Gazebo or real-life environments to allow recording realistic coverage times. Further research is required to develop a machine-learning technique that enables CCPP to autonomously determine a proper combination of parameter values that would provide optimized coverage based on the robot's properties and the known properties of the covered environment. The CCPP method was shown to have potential advantages over BCPP due to the computational complexity involved in cellular decomposition. It is therefore plausible that our CCPP method has these same advantages over all methods which use cellular decomposition, for instance, the Morse-based cellular decomposition method. Future study should involve the testing of the CCPP method against other types of CPP methods. Development of a local planner specifically for use in CCPP will help to improve the kinematic \color{black} 
 efficiency of the planner for much higher velocities. Lastly, regardless of the experimental results, the algorithms presented in this work have not as
of yet been theoretically justified. This is an important future task to ensure robustness.\color{black}
 \section{Acknowledgements}

This work was supported by 2022 San Diego State University Seed Grant Program.

%\appendix
%\section{Example Appendix Section}
%\label{app1}


%%       \cite{<label>} ==> [1]

%%
%Example citation, See \cite{lamport94}.




\bibliographystyle{elsarticle-num}
\bibliography{jfrExampleRefs}

\begin{thebibliography}{10}
\expandafter\ifx\csname url\endcsname\relax
  \def\url#1{\texttt{#1}}\fi
\expandafter\ifx\csname urlprefix\endcsname\relax\def\urlprefix{URL }\fi
\expandafter\ifx\csname href\endcsname\relax
  \def\href#1#2{#2} \def\path#1{#1}\fi

\bibitem{choi2017b}
S.~Choi, S.~Lee, H.~H. Viet, T.~Chung, B-theta*: an efficient online coverage algorithm for autonomous cleaning robots, Journal of Intelligent \& Robotic Systems 87~(2) (2017) 265--290.

\bibitem{hameed2014intelligent}
I.~A. Hameed, Intelligent coverage path planning for agricultural robots and autonomous machines on three-dimensional terrain, Journal of Intelligent \& Robotic Systems 74~(3) (2014) 965--983.

\bibitem{choi2020energy}
Y.~Choi, Y.~Choi, S.~Briceno, D.~N. Mavris, Energy-constrained multi-uav coverage path planning for an aerial imagery mission using column generation, Journal of Intelligent \& Robotic Systems 97~(1) (2020) 125--139.

\bibitem{di2016coverage}
C.~Di~Franco, G.~Buttazzo, Coverage path planning for uavs photogrammetry with energy and resolution constraints, Journal of Intelligent \& Robotic Systems 83~(3) (2016) 445--462.

\bibitem{faigl2011sensor}
J.~Faigl, M.~Kulich, L.~P{\v{r}}eu{\v{c}}il, A sensor placement algorithm for a mobile robot inspection planning, Journal of Intelligent \& Robotic Systems 62~(3) (2011) 329--353.

\bibitem{grotli2012path}
E.~I. Gr{\o}tli, T.~A. Johansen, Path planning for uavs under communication constraints using splat! and milp, Journal of Intelligent \& Robotic Systems 65~(1) (2012) 265--282.

\bibitem{hsu2014complete}
P.-M. Hsu, C.-L. Lin, M.-Y. Yang, On the complete coverage path planning for mobile robots, Journal of Intelligent \& Robotic Systems 74~(3) (2014) 945--963.

\bibitem{kapoutsis2017darp}
A.~C. Kapoutsis, S.~A. Chatzichristofis, E.~B. Kosmatopoulos, Darp: divide areas algorithm for optimal multi-robot coverage path planning, Journal of Intelligent \& Robotic Systems 86~(3) (2017) 663--680.

\bibitem{li2014k}
Y.~Li, D.~Li, C.~Maple, Y.~Yue, J.~Oyekan, K-order surrounding roadmaps path planner for robot path planning, Journal of Intelligent \& Robotic Systems 75~(3) (2014) 493--516.

\bibitem{zhu2019complete}
D.~Zhu, C.~Tian, B.~Sun, C.~Luo, Complete coverage path planning of autonomous underwater vehicle based on gbnn algorithm, Journal of Intelligent \& Robotic Systems 94~(1) (2019) 237--249.

\bibitem{nakamura2001chaotic}
Y.~Nakamura, A.~Sekiguchi, The chaotic mobile robot, IEEE Transactions on Robotics and Automation 17~(6) (2001) 898--904.

\bibitem{pimentel2017chaotic}
C.~Pimentel-Romero, J.~M. Mu{\~n}oz-Pacheco, O.~Felix-Beltran, L.~Gomez-Pavon, C.~K. Volos, Chaotic planning paths generators by using performance surfaces, in: Fractional Order Control and Synchronization of Chaotic Systems, Springer, 2017, pp. 805--832.

\bibitem{li2013improved}
C.~Li, F.~Wang, L.~Zhao, Y.~Li, Y.~Song, An improved chaotic motion path planner for autonomous mobile robots based on a logistic map, International Journal of Advanced Robotic Systems 10~(6) (2013) 273.

\bibitem{sridharan2020multi}
K.~Sridharan, Z.~N. Ahmadabadi, A multi-system chaotic path planner for fast and unpredictable online coverage of terrains, IEEE Robotics and Automation Letters 5~(4) (2020) 5268--5275.

\bibitem{sridharan2022online}
K.~Sridharan, P.~McNamee, Z.~Nili~Ahmadabadi, J.~Hudack, Online search of unknown terrains using a dynamical system-based path planning approach, Journal of Intelligent \& Robotic Systems 106~(1) (2022) 1--19.

\bibitem{volos2013experimental}
C.~K. Volos, I.~M. Kyprianidis, I.~N. Stouboulos, Experimental investigation on coverage performance of a chaotic autonomous mobile robot, Robotics and Autonomous Systems 61~(12) (2013) 1314--1322.

\bibitem{majeed2020mobile}
A.~I. Majeed, Mobile robot motion control based on chaotic trajectory generation, Journal of Engineering and Sustainable Development 24~(4) (2020) 48--55.

\bibitem{tlelo2014application}
E.~Tlelo-Cuautle, H.~C. Ramos-L{\'o}pez, M.~S{\'a}nchez-S{\'a}nchez, A.~D. Pano-Azucena, L.~A. S{\'a}nchez-Gaspariano, J.~C. N{\'u}{\~n}ez-P{\'e}rez, J.~L. Camas-Anzueto, Application of a chaotic oscillator in an autonomous mobile robot, Journal of Electrical Engineering 65~(3) (2014) 157.

\bibitem{sooraska2010no}
P.~Sooraska, K.~Klomkarn, " no-cpu" chaotic robots: from classroom to commerce, IEEE circuits and systems magazine 10~(1) (2010) 46--53.

\bibitem{ahuraka2023chaotic}
F.~Ahuraka, P.~Mcnamee, Q.~Wang, Z.~N. Ahmadabadi, J.~Hudack, Chaotic motion planning for mobile robots: Progress, challenges, and opportunities, IEEE Access 11 (2023) 134917--134939.

\bibitem{samet1988overview}
H.~Samet, An overview of quadtrees, octrees, and related hierarchical data structures, Theoretical Foundations of Computer Graphics and CAD (1988) 51--68.

\bibitem{huang2020multi}
X.~Huang, M.~Sun, H.~Zhou, S.~Liu, A multi-robot coverage path planning algorithm for the environment with multiple land cover types, IEEE Access 8 (2020) 198101--198117.

\bibitem{jan2019complete}
G.~E. Jan, C.~Luo, H.-T. Lin, K.~Fung, Complete area coverage path-planning with arbitrary shape obstacles, Journal of Automation and Control Engineering Vol 7~(2) (2019).

\bibitem{volos2012implementation}
C.~K. Volos, N.~Bardis, I.~M. Kyprianidis, I.~N. Stouboulos, Implementation of mobile robot by using double-scroll chaotic attractors, Recent Researches in Applications of Electrical and Computer Engineering (2012) 119--124.

\bibitem{agiza2001synchronization}
H.~Agiza, M.~Yassen, Synchronization of rossler and chen chaotic dynamical systems using active control, Physics Letters A 278~(4) (2001) 191--197.

\bibitem{li2016bounded}
C.~Li, Y.~Song, F.~Wang, Z.~Wang, Y.~Li, A bounded strategy of the mobile robot coverage path planning based on lorenz chaotic system, International journal of advanced robotic systems 13~(3) (2016) 107.

\bibitem{lu2004new}
J.~L{\"u}, G.~Chen, D.~Cheng, A new chaotic system and beyond: the generalized lorenz-like system, International Journal of Bifurcation and Chaos 14~(05) (2004) 1507--1537.

\bibitem{petavratzis2022experimental}
E.~Petavratzis, C.~Volos, L.~Moysis, H.~Nistazakis, A.~Giakoumis, I.~Stouboulos, Experimental coverage performance of a chaotic autonomous mobile robot, in: 2022 11th International Conference on Modern Circuits and Systems Technologies (MOCAST), IEEE, 2022, pp. 1--4.

\bibitem{moysis2021chaotic}
L.~Moysis, K.~Rajagopal, A.~V. Tutueva, C.~Volos, B.~Teka, D.~N. Butusov, Chaotic path planning for 3d area coverage using a pseudo-random bit generator from a 1d chaotic map, Mathematics 9~(15) (2021) 1821.

\bibitem{petavratzis20212d}
E.~Petavratzis, C.~Volos, A.~Ouannas, H.~Nistazakis, K.~Valavanis, I.~Stouboulos, A 2d discrete chaotic memristive map and its application in robot’s path planning, in: 2021 10th International Conference on Modern Circuits and Systems Technologies (MOCAST), IEEE, 2021, pp. 1--4.

\bibitem{petavratzis2019coverage}
E.~K. Petavratzis, C.~K. Volos, I.~N. Stouboulos, H.~E. Nistazakis, K.~G. Kyritsi, K.~P. Valavanis, Coverage performance of a chaotic mobile robot using an inverse pheromone model, in: 2019 8th International Conference on Modern Circuits and Systems Technologies (MOCAST), IEEE, 2019, pp. 1--4.

\bibitem{arrowsmith1993bogdanov}
D.~K. Arrowsmith, J.~H. Cartwright, A.~N. Lansbury, C.~M. Place, The bogdanov map: Bifurcations, mode locking, and chaos in a dissipative system, International Journal of Bifurcation and Chaos 3~(04) (1993) 803--842.

\bibitem{curiac20142d}
D.-I. Curiac, C.~Volosencu, A 2d chaotic path planning for mobile robots accomplishing boundary surveillance missions in adversarial conditions, Communications in Nonlinear Science and Numerical Simulation 19~(10) (2014) 3617--3627.

\bibitem{li2017chaotic}
C.-h. Li, Y.~Song, F.-y. Wang, Z.-q. Wang, Y.-b. Li, A chaotic coverage path planner for the mobile robot based on the chebyshev map for special missions, Frontiers of Information Technology \& Electronic Engineering 18~(9) (2017) 1305--1319.

\bibitem{li2015chaotic}
C.~Li, Y.~Song, F.~Wang, Z.~Liang, B.~Zhu, Chaotic path planner of autonomous mobile robots based on the standard map for surveillance missions, Mathematical Problems in Engineering 2015 (2015).

\bibitem{dwa2020}
Dwa local planner, available from: \url{http://wiki.ros.org/dwa_local_planner} [Accessed: 13 May, 2024]. (2020).

\bibitem{teb2020}
Teb local planner, available from: \url{http://wiki.ros.org/teb_local_planner} [Accessed: 13 May, 2024]. (2020).

\bibitem{Quadtrees}
Christian, {{Q}}uadtrees \#2: {{I}}mplementation in {{P}}ython, available from: \url{https://scipython.com/blog/quadtrees-2-implementation-in-python} [Accessed November 12, 2023]. (2020).

\bibitem{navmsgs}
nav\_msgs/{{P}}ath {{M}}essage, \url{http://docs.ros.org/en/api/nav_msgs/html/msg/Path.html} [Accessed September 24, 2022]. (2022).

\bibitem{sensormsgs}
sensor\_msgs/{{L}}aserscan {{M}}essage, available from: \url{http://docs.ros.org/en/melodic/api/sensor_msgs/html/msg/LaserScan.html} [Accessed September 14, 2022]. (2022).

\bibitem{gmapping}
B.~Gerkey, {{ROS}} gmapping package, available from: \url{http://wiki.ros.org/gmapping} [Accessed October 03, 2023].

\bibitem{choset2000coverage}
H.~Choset, Coverage of known spaces: The boustrophedon cellular decomposition, Autonomous Robots 9 (2000) 247--253.

\bibitem{choset1998coverage}
H.~Choset, P.~Pignon, Coverage path planning: The boustrophedon cellular decomposition, in: Field and service robotics, Springer, 1998, pp. 203--209.

\bibitem{ntawumenyikizaba2012online}
A.~Ntawumenyikizaba, H.~H. Viet, T.~Chung, An online complete coverage algorithm for cleaning robots based on boustrophedon motions and a* search, in: 2012 8th International Conference on Information Science and Digital Content Technology (ICIDT2012), Vol.~2, IEEE, 2012, pp. 401--405.

\bibitem{coombes2019flight}
M.~Coombes, W.-H. Chen, C.~Liu, Flight testing boustrophedon coverage path planning for fixed wing uavs in wind, in: 2019 International Conference on Robotics and Automation (ICRA), IEEE, 2019, pp. 711--717.

\bibitem{bahnemann2021revisiting}
R.~B{\"a}hnemann, N.~Lawrance, J.~J. Chung, M.~Pantic, R.~Siegwart, J.~Nieto, Revisiting boustrophedon coverage path planning as a generalized traveling salesman problem, in: Field and Service Robotics: Results of the 12th International Conference, Springer, 2021, pp. 277--290.

\bibitem{rekleitis2008efficient}
I.~Rekleitis, A.~P. New, E.~S. Rankin, H.~Choset, Efficient boustrophedon multi-robot coverage: an algorithmic approach, Annals of Mathematics and Artificial Intelligence 52 (2008) 109--142.

\bibitem{ethz-asl}
R.~Girod, L.~Liu, D.~K. et~al., polygon\_coverage\_planning, available from: \url{https://github.com/ethz-asl/polygon_coverage_planning} [Accessed September 07, 2023]. (2018).

\bibitem{rjj}
{{C}}overage {{P}}lanning, available from: \url{https://github.com/RJJxp/CoveragePlanning} [Accessed October 04, 2023]. (2019).

\bibitem{gonzalez2005bsa}
E.~Gonzalez, O.~Alvarez, Y.~Diaz, C.~Parra, C.~Bustacara, Bsa: A complete coverage algorithm, in: proceedings of the 2005 IEEE international conference on robotics and automation, IEEE, 2005, pp. 2040--2044.

\bibitem{bormann2018indoor}
R.~Bormann, F.~Jordan, J.~Hampp, M.~H{\"a}gele, Indoor coverage path planning: Survey, implementation, analysis, in: 2018 IEEE International Conference on Robotics and Automation (ICRA), IEEE, 2018, pp. 1718--1725.

\bibitem{gomez2017optimal}
J.~I.~V. Gomez, M.~M. Melchor, J.~C.~H. Lozada, Optimal coverage path planning based on the rotating calipers algorithm, in: 2017 International Conference on Mechatronics, Electronics and Automotive Engineering (ICMEAE), IEEE, 2017, pp. 140--144.

\bibitem{green}
C.~B. Quinn, S.~Leonis, O.~So, S.~Macenski, boustrophedon\_planner, available from: \url{https://github.com/Greenzie/boustrophedon_planner} [Accessed September 23, 2023]. (2019).

\bibitem{Ipiano}
A.~Stelter, coverage-planning, available from: \url{https://github.com/Ipiano/coverage-planning} [Accessed September 14, 2023]. (2019).

\bibitem{galceran2013survey}
E.~Galceran, M.~Carreras, A survey on coverage path planning for robotics, Robotics and Autonomous systems 61~(12) (2013) 1258--1276.

\bibitem{galceran2012efficient}
E.~Galceran, M.~Carreras, Efficient seabed coverage path planning for asvs and auvs, in: 2012 IEEE/RSJ International Conference on Intelligent Robots and Systems, IEEE, 2012, pp. 88--93.

\end{thebibliography}
\end{document}